\documentclass[pdflatex,sn-mathphys-num]{sn-jnl}

\usepackage{graphicx}%
\usepackage{caption}
\usepackage{subcaption}
\usepackage{multirow}%
\usepackage{amsmath,amssymb,amsfonts}%
\usepackage{amsthm}%
\usepackage{mathrsfs}%
\usepackage[title]{appendix}%
\usepackage{xcolor}%
\usepackage{textcomp}%
\usepackage{manyfoot}%
\usepackage{booktabs}%
\usepackage{algorithm}%
\usepackage{algorithmicx}%
\usepackage{algpseudocode}%
\usepackage{listings}%
\usepackage{longtable}

\usepackage{tikz}
\usetikzlibrary{decorations.pathmorphing}

\DeclareMathOperator*{\argmin}{arg\,min}\usepackage{graphicx}%

\usetikzlibrary{plotmarks}
\newcommand{\markercircle}{\tikz{\pgfuseplotmark{o}}}
\newcommand{\markersquare}{\tikz{\pgfuseplotmark{square}}}

\theoremstyle{thmstyleone}%
\theoremstyle{thmstyletwo}%

\theoremstyle{thmstylethree}%
\newtheorem{definition}{Definition}%

\begin{document}

\title[Dynamic Design of Machine Learning Pipelines via Metalearning]{Dynamic Design of Machine Learning Pipelines via Metalearning
}


\author*[1]{\fnm{Edesio} \sur{Alcoba\c{c}a}}\email{e.alcobaca@gmail.com}

\author[1]{\fnm{André C. P. L. F.} \sur{de Carvalho}}\email{andre@icmc.usp.br}

\affil*[1]{\orgdiv{Institute of Mathematics and Computer Sciences}, \orgname{University of S\~ao Paulo}, \orgaddress{\street{Av. Trabalhador S\~ao Carlense}, \city{S\~ao Carlos}, \postcode{13566-590}, \state{S\~ao Paulo}, \country{Brazil}}}


\abstract{
Automated machine learning (AutoML) has democratized the design of
machine learning based systems, by automating model selection, hyperparameter tuning and feature engineering.
However, the high computational cost associated with traditional search and optimization strategies, such as Random Search, Particle Swarm Optimization and Bayesian Optimization, remains a significant challenge.
Moreover, AutoML systems typically explore a large search space, which can lead to overfitting.
This paper introduces a metalearning method for dynamically designing search spaces for AutoML system.
The proposed method uses historical metaknowledge to select promising regions of the search space, accelerating the optimization process.
According to experiments conducted for this study, the proposed method can reduce runtime by 89\% in Random Search and search space by (1.8/13 preprocessor and 4.3/16 classifier), without compromising significant predictive performance.
Moreover, the proposed method showed competitive performance when adapted to Auto-Sklearn, reducing its search space.
Furthermore, this study encompasses insights into meta-feature selection, meta-model explainability, and the trade-offs inherent in search space reduction strategies.
}




\maketitle

\section{Introduction}

Automated Machine Learning (AutoML) has become an essential tool for democratizing machine learning (ML) by automating key aspects of model selection, hyperparameter tuning, and feature engineering \cite{he-2021, barbudo-2023}.

However, the efficiency of AutoML frameworks remains a significant challenge, as the search for optimal configurations is often computationally expensive \cite{thornton-2013, olson-2016a, fabris-2019}. Traditional search strategies, such as Random Search (RS) and Bayesian Optimization (BO), indiscriminately explore large search spaces, resulting in high resource consumption \cite{thornton-2013, olson-2016b, de-sa-2017}.

 To address this challenge, we propose a metalearning approach that dynamically designs search spaces for an AutoML solution, reducing computational costs while maintaining competitive predictive performance.
 The proposed method leverages historical metaknowledge to identify and prioritize promising regions of the search space, enabling more efficient optimization.
 By predicting the performance of preprocessor-classifier combinations, a meta-model,
 induced using metalearning, can provide a warm-start advantage, accelerating the AutoML search process.


This study evaluates the effectiveness of the proposed 
approach through an extensive set of experiments, analyzing both computational efficiency and predictive performance. According to the experimental results,
the dynamically generated search spaces significantly reduce runtime, while maintaining high-quality solutions. In particular, the RS-mtl-95 configuration achieved an 89\% reduction in runtime without compromising predictive performance.
Additionally, the findings highlight key insights into meta-feature selection, classifier preferences, and the trade-offs present in search space reduction strategies.


The main contributions from
this paper are:
(i) empirical analysis of the computational cost associated with extracting meta-features using the \textit{pymfe} package~\cite{alcobacca-2020}; (ii) proposal of a new set of meta-features derived from historical pipeline performance data; (iii) analysis of meta-feature selection and meta-model interpretability in the context of pipeline design; (iv) study of the trade-offs between search space reduction and predictive accuracy for pipeline design; (v) development of a novel metalearning framework for the dynamical construction of pipeline search spaces with an experimental validation showing significant runtime reductions with minimal impact on predictive performance.

The remainder of this paper is structured as follows: Section~\ref{sec:background-dp} discusses background and related work in AutoML and metalearning. Section~\ref{sec:design-machine-learning-pipelines} presents the proposed method, detailing the metalearning framework and dynamic search space generation process. Section~\ref{sec:methodology-dp} describes the experimental setup, including datasets, evaluation metrics, and baselines. Section~\ref{sec:results-and-discussion} analyzes the results, and 
Section~\ref{sec:final-remarks} provides key insights and limitations of this study.
Finally,
Section~\ref{sec:conclusion-dp-dp} presents the main conclusions and points out future research directions.

\section{Background}
\label{sec:background-dp}

AutoML comprehends a set of methods developed to automatically design ML
solutions, learning to learn the learning process with minimal human interaction. Its main goals include democratizing ML access, assisting data scientists in laborious and repetitive tasks involved in designing ML solutions, identifying the best possible solution given resource limitations and constraints, and accumulating meta-knowledge to enhance efficiency at each step of the learning process \cite{olson-2016, de-sa-2017, feurer-2022}. 

Various methods have been developed to automatically design ML solutions~\cite{he-2021, barbudo-2023}. Metalearning, e.g., uses previous learning experiences for algorithm recommendation \cite{rivolli-2022}. Hyperparameter optimization is employed to fine-tune hyperparameters of ML algorithms to improve model performance \cite{morales-2023}. Neural architecture search (NAS) looks for the best neural network architectures, usually for deep learning tasks \cite{zheng-2023}. Automated pipeline design streamlines the creation of end-to-end ML pipelines by selecting and tuning both preprocessing steps and modeling algorithms \cite{barbudo-2023}. In the following subsection, these methods are covered in the context of pipeline design.

\subsection{Metalearning}

Metalearning is a broad subfield of ML that focuses on understanding and improving the learning process itself by leveraging accumulated knowledge to enhance new learning tasks \cite{brazdil-2009}. The most common application of metalearning is to recommend ML algorithms \cite{brazdil-2003, lemke-2010, cruz-2017}. However, its scope extends beyond algorithm selection, encompassing various tasks such as hyperparameter recommendation \cite{soares-2004}, preprocessing algorithm selection \cite{filchenkov-2015, garcia-2016, parmezan-2017}, identifying the need for hyperparameter tuning \cite{mantovani-2019}, and warm-start optimization processes \cite{gomes-2012, feurer-2015b}. According to \citeauthor{brazdil-2009}, metalearning refers to the systematic use of meta-knowledge to improve the learning processes by recommending ML algorithms for new tasks~\cite[p.~10]{brazdil-2009}.

\begin{figure}[!htb]
 \centering
 \includegraphics[width=1.0\textwidth]{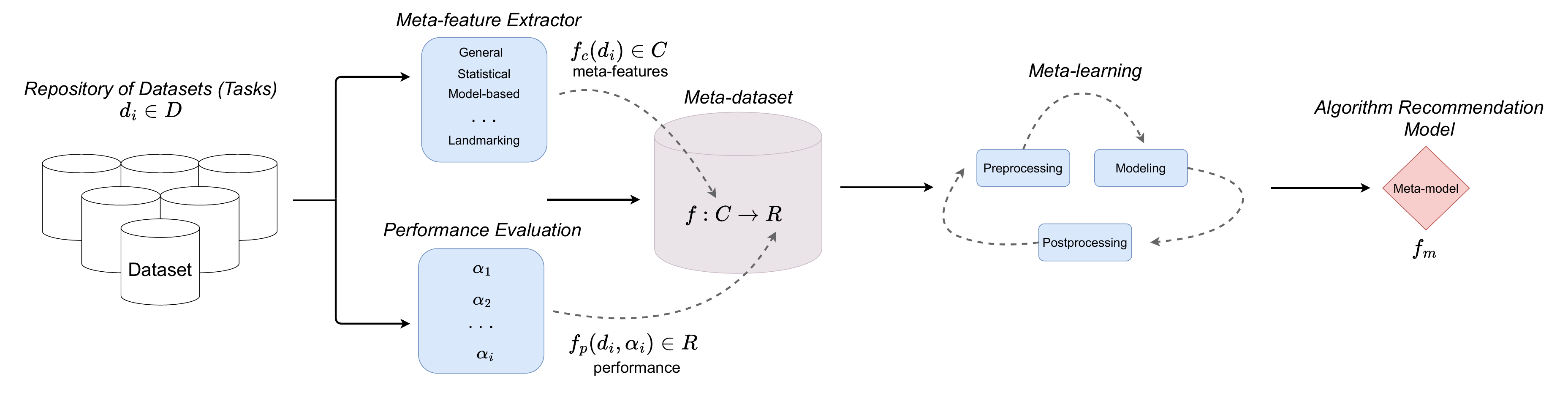}
\caption{Metalearning system.}
\label{dynamic-seach-space-meta-learning}
\end{figure}

Figure \ref{dynamic-seach-space-meta-learning} illustrates a typical metalearning recommendation system for selecting 
ML algorithms. In this system, meta-features are extracted from datasets along with the performance metrics of various algorithms. Meta-features capture the key characteristics of a dataset and its relationship to the corresponding learning task. The performance outcomes, referred to as the meta-target, are combined with the meta-features to create a meta-dataset. The meta-learner, which can be a conventional ML
algorithm, uses this meta-dataset to build a recommendation model, known as the meta-model. When a new task, i.e., a new dataset, is introduced, its meta-features are extracted and provided as input to the meta-model. The meta-model then recommends the most suitable algorithm specifically designed to the unseen dataset~\cite{brazdil-2009}.

One of the most challenging tasks in metalearning is generating meta-features. This process requires domain expertise to engineer meaningful features for ML 
tasks (e.g., classification, regression, clustering). To address this challenge, \citeauthor{pinto-2016} proposes a framework for the systematic extraction of meta-features, which demonstrates improved performance compared to traditional hand-made methods. \citeauthor{rivolli-2018} extended this framework by formalizing meta-features and developing procedures to guide the extraction process. This formalization enhances the reproducibility of metalearning experiments. Additionally, the \textit{pymfe} package is a tool designed to systematically extract meta-features, providing a robust way of getting meta-features~\cite{alcobacca-2020}.

The following categories of meta-features are commonly used in literature: (i) \textit{general}~\cite{brazdil-2009}; (ii) \textit{statistical}~\cite{reif-2014}; (iii). \textit{information-theoretic}~\cite{segrera-2008}; (iv) \textit{model-based}~\cite{peng-2002}; (v) \textit{landmarking} ~\cite{pfahringer-2000} (vi) \textit{subsampling landmarking}~\cite{furnkranz-2001}; (vii) \textit{relative landmarking}~\cite{soares-2001, vanschoren-2010}.
    
The literature has demonstrated that metalearning is a promising approach for designing recommendation systems for ML 
algorithms~\cite{soares-2004, reif-2014, mantovani-2019}. One of its main strengths is that all the effort required for recommendation occurs offline, meaning that the computational cost for recommending algorithms and hyperparameters during the online step is very low \cite{vanschoren-2018}. Additionally, the analysis of the generated meta-knowledge can contribute to the development of ML 
algorithms \cite{smith-miles-2009}.

A major drawback is the need to collect a representative set of tasks, which can be costly and time-consuming \cite{hutter-2019}. Furthermore, the success of metalearning heavily depends on problem characterization, particularly in meta-feature engineering, which requires careful design to ensure meaningful and relevant results~\cite{smith-miles-2009}.

\subsection{Optimization}

\citeauthor{boyd-2004} defines an optimization problem (or mathematical optimization problem) as presented in Definition~\ref{def-optimization-problem-dp}.

\quad
\begin{definition}[Optimization Problem]\label{def-optimization-problem-dp}
Let the vector $\textbf{x} = (x_1,\dotsc, x_n)$ be the \textit{optimization variable} of the problem, let the function $f : \mathbb{R}^n \to \mathbb{R}$ be the \textit{objective function}, let the functions $g_i : \mathbb{R}^n \to \mathbb{R},~i=1,\dotsc,m$ be the \textit{constraint function} (inequality constraint) and the constants $b_1,\dotsc, b_m$ be the limits (or bounds) for the constraints. An optimization problem can be defined as follows:

\begin{subequations}
\begin{alignat}{2}
&\!\text{\textit{minimize}}        &\qquad& f(\textbf{x})\\
&\text{\textit{subject to}} &      & g_i(\textbf{x}) \leq b_i, \qquad i=1,\dotsc, m.
\end{alignat}
\end{subequations}
\end{definition}
\quad

The optimization problem can be named
linear programming if the objective and constraint functions $f, g_1,\dots,g_m$ are linear, otherwise it is called nonlinear programming. Nonlinear problems typically assume that the objective function $f$ is convex or at least mathematically defined. However, there are problems where the objective function is computationally expensive or even impossible to compute, and the derivatives and convexity properties are unknown~\cite{brochu-2010}. 

AutoML is an example of this kind of optimization problem where, although it is possible to compute the objective function, it is unknown (so-called black-box function), with no guarantee of convexity nor derivatives, and may have multiple local minima and maxima. Moreover, the objective function may return noisy evaluations. Unfortunately, there is no effective method for solving these complex nonlinear problems \cite{boyd-2004}. According to \citeauthor{boyd-2004}, these problems, even the simplest, can be extremely challenging, expend much computational time, or even be intractable.

Optimization can also be local or global. In global optimization, the goal is to find the $\textbf{x}^*$, i.e., the optimal point that satisfies the constraints, for all feasible points \cite{boyd-2004}. For local optimization, finding the  $\textbf{x}^*$ is also the goal, but only for a neighborhood of feasible points, which do not guarantee having the lowest objective value among all other feasible points. Most of the optimization methods proposed for AutoML are global optimization problems. Nonetheless, it is important to note that local approaches can bring progress and refine the best solution found \cite{hutter-2019}.

There are many ways to model AutoML as an optimization problem. The most usual is the Combined Algorithm Selection and Hyperparameter optimization (CASH)~\cite{thornton-2013}, described in Definition \ref{def-cash-dp}.

\quad
\begin{definition}[CASH]
\label{def-cash-dp}
Let $A = \{  \alpha_1, \cdots, \alpha_n \}$ be a set of algorithms and $\{ \Lambda_1, \cdots, \Lambda_n\}$ the respective associated algorithm hyperspaces. Given a loss function $\mathcal{L}(\alpha_j(d^t_i, \lambda),  d^v_i)$, where $\lambda \in \Lambda_j$, the $k$-fold cross-validation inner training set $d^\text{t}_i \in D_\text{train}$ and validation set as $d^\text{v}_i \in D_\text{train}$ with $d^\text{t}_i \cap d^\text{v}_i = \emptyset$, $d^\text{t}_i \cup d^\text{v}_i = D_{\text{train}}$ and $i \leq k \mid i, j, k \in \mathbb{N}^*$, we can define CASH as:

\begin{equation}
    \alpha^{*}\lambda ^{*}  \in  \argmin_{\alpha_j \in A, \lambda \in \Lambda_j} \frac{1}{k}\sum_{i=1}^{k} \mathcal{L}(\alpha_j(d^t_i, \lambda),  d^v_i).
\end{equation}
\end{definition}
\quad

Note that the function $f$ in CASH is approximated via cross-validation, though alternatives, such as
holdout or leave-one-out, may also be used. Moreover, several methods have been proposed for hyperparameter optimization (HPO), such as Random Search, Grid Search, Evolutionary Algorithms, and Bayesian Optimization~\cite{bergstra-2011, pappa-2014, feurer-2019, yang-2020}.

\subsection{Automated Pipeline Design}

\begin{figure}[!htb]
 \centering
 \includegraphics[width=0.70\textwidth]{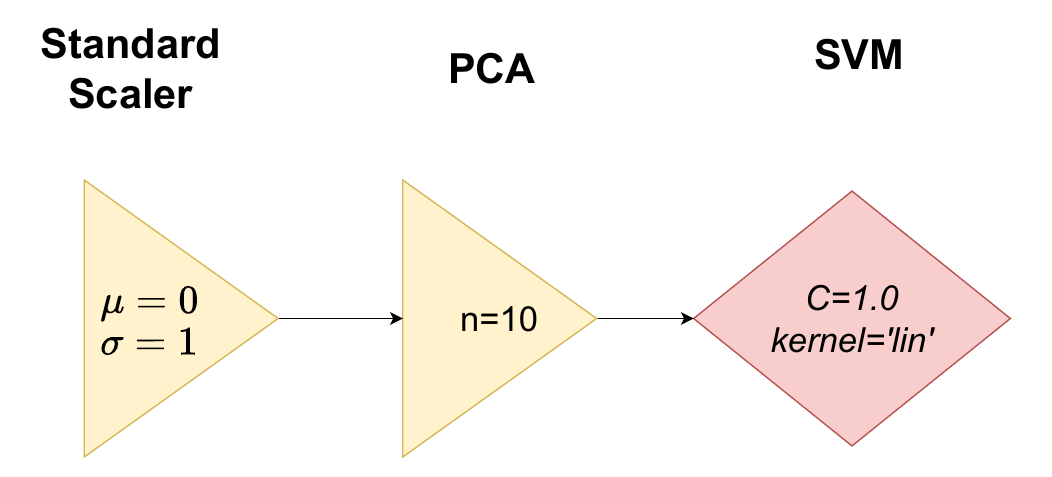}
  \caption{A machine learning pipeline.}
 \label{fig-automl-pipeline}
\end{figure}

A key concept in AutoML is its
pipeline,
a sequence of operations over the data \cite{olson-2016b}. Figure \ref{fig-automl-pipeline} shows an example of a pipeline, where a sequence of configured algorithms is used to solve an illustrative problem. In this scenario, the pipeline has three operations, two for preprocessing 
(Standard Scaler and 
Principal component analysis), and one for data modeling (a support Vector machine model). A pipeline typically consists of three main stages: (i) the preprocessing, for dealing with data balancing, noise reduction, normalization, and dimensionality reduction; 
(ii) 
modeling;
(iii) postprocessing.

Several AutoML systems have been developed to automatically design ML 
pipelines. Auto-Weka, one of the first, is
based on Bayesian Optimization, which leverages methods, e.g.,
SMAC and TPE to explore a hierarchical hyperparameter space over the Weka library~\cite{thornton-2013}. Auto-Sklearn also employs Bayesian Optimization but introduces key innovations, such as metalearning for warm-starting the optimization and ensemble construction as a post-processing step~\cite{feurer-2015a}. In contrast, TPOT uses Genetic Programming to evolve pipelines with flexible structures, allowing multiple preprocessing and modeling stages, and employs Pareto optimization to balance performance and pipeline complexity~\cite{olson-2016a}. RECIPE also relies on Genetic Programming but adopts a grammar-based approach to ensure the validity of generated pipelines and includes mechanisms to mitigate overfitting, such as periodically altering training and validation splits~\cite{de-sa-2017}.

Recent advances 
in AutoML research have focused on improving pipeline synthesis, search efficiency, and configurability. One notable contribution is the incremental construction of the search space guided by meta-features, as proposed by~\cite{zoller-2021}. This approach mimics human-like behavior by incrementally expanding the pipeline structure and leveraging the meta-features of intermediate datasets to prune unpromising configurations early. This enables the construction of flexible, dataset-specific pipelines and has shown competitive performance across standard AutoML benchmarks. Another important advancement is Auto-Sklearn 2.0, which introduces PoSH Auto-Sklearn, a method designed for efficiency on large datasets and under tight time constraints~\cite{feurer-2022}. It removes the reliance on traditional meta-features by employing a meta-feature-free metalearning strategy and uses a bandit-based budget allocation method. Additionally, it explores the broader AutoML design space to create a more user-friendly and hands-free AutoML system. Finally,~\cite{kedziora-2024} investigated the use of opportunistic and systematic meta-knowledge to reduce the AutoML configuration space. By leveraging historical performance data from previous AutoML runs, this method identifies promising subsets of algorithms and hyperparameters tailored to the characteristics of new datasets. Together, these studies highlight a trend toward more intelligent and adaptive AutoML systems.

\section{Dynamic Design of Machine Learning Pipelines}
\label{sec:design-machine-learning-pipelines}

Despite recent advances in AutoML pipeline design, several challenges remain open. This work focuses on two key aspects: computational cost and large hyperparameter search spaces \cite{thornton-2013, olson-2016b, de-sa-2017}. The first results in computationally expensive systems that require a significant amount of time for model training. The second arises due to the natural combinatorial complexity of the CASH problem, often leading to an excessive number of algorithms in the search space. This increased complexity can, in turn, lead to overfitting~\cite{fabris-2019}.

\begin{figure}[!htb]
    \centering
    \includegraphics[width=1.0\textwidth]{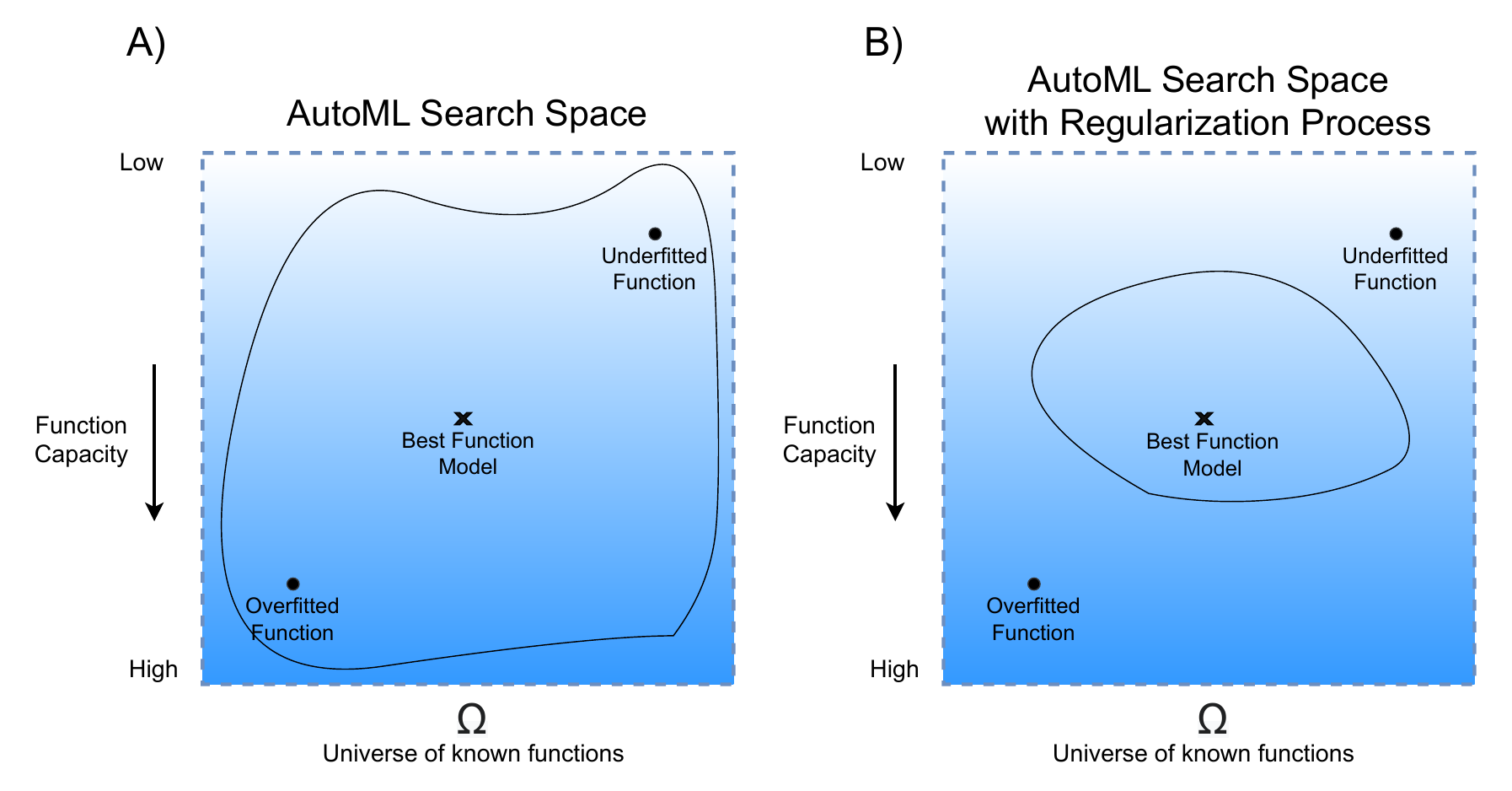}
    \caption{Illustration of the theoretical overfitting problem that large search spaces can cause. An AutoML system with an unrestricted algorithm search space is presented in part A (left side). A hypothetical regularization technique is introduced to reduce the search space, preventing the selection of excessively complex models and mitigating overfitting, which is presented in part B (right side).}

    \label{figure:dynamic-seach-space-overfitting-problem}
\end{figure}

Figure \ref{figure:dynamic-seach-space-overfitting-problem} illustrates the theoretical overfitting problem in AutoML due to a large search space. In part A, an AutoML system with an unrestricted set of algorithms can generate an overly complex function that, while it fits the training data perfectly, fails to generalize. Given sufficient time, this system will present overfitting. Conversely, part B of the figure presents a hypothetical regularization technique designed to reduce the search space, thereby preventing the selection of excessively complex models and mitigating overfitting. In this work, we aim to explore an approach that dynamically reduces the search space, which could act as a regularization mechanism for AutoML.

Unlike previous works, 
the proposed
approach dynamically reduces the search space~\cite{thornton-2013, olson-2016, de-sa-2017, feurer-2022}. We propose a metalearning framework that generates pipeline search spaces based on prior meta-knowledge. Initially, we learn from past experiences to determine which combinations of preprocessor-classifier perform best for specific ML 
tasks. Using this information, we build a tailored search space for each task, thereby constraining AutoML's capacity while maintaining performance.

\begin{figure}[!htb]
    \centering
    \includegraphics[width=1.0\textwidth]{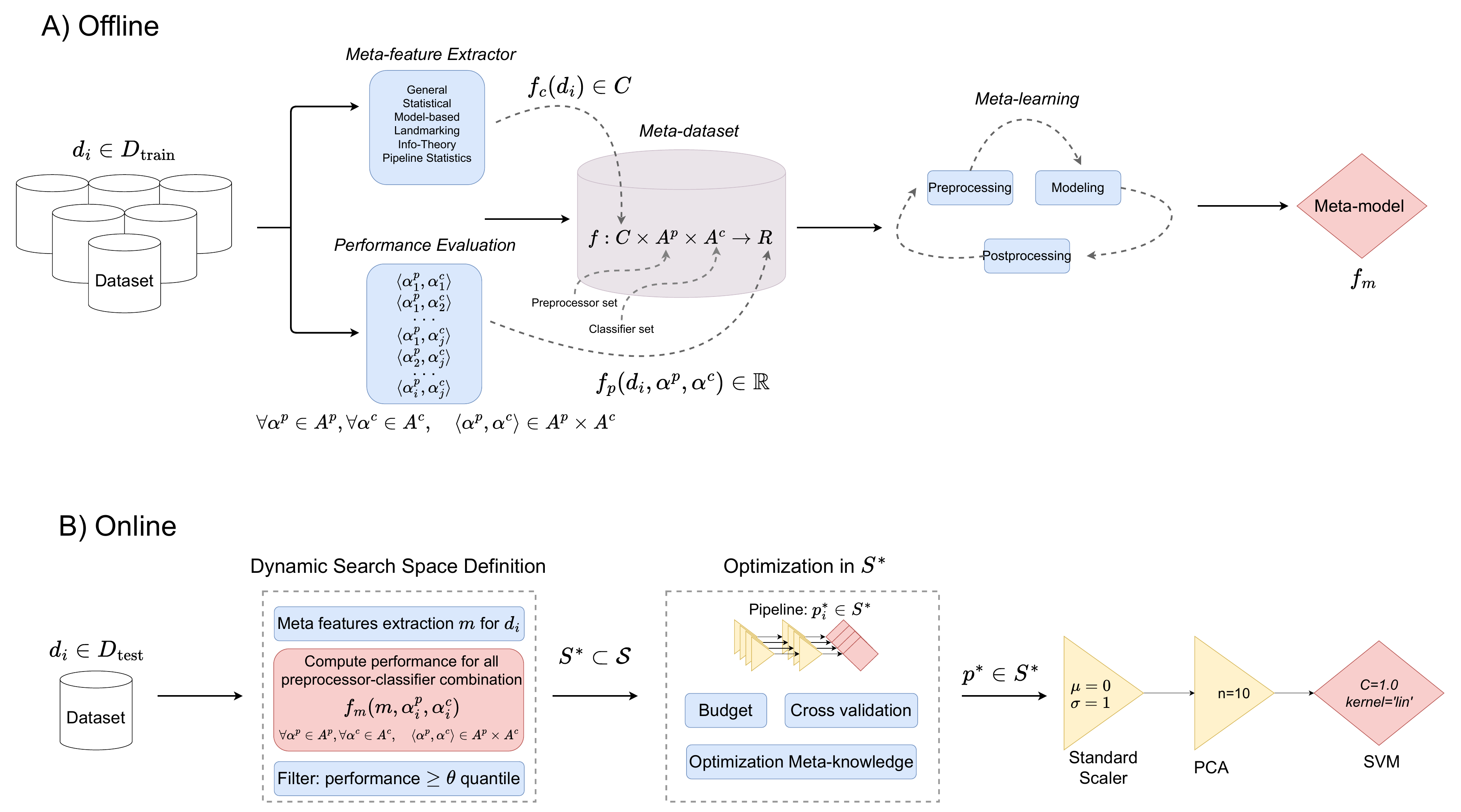}
    \caption{Overview of the dynamic design of ML pipelines. (A) The offline phase includes generating meta-knowledge and training a meta-model using a predefined set of tasks. (B) In the online phase, when a new task is presented, the system dynamically creates a tailored search space. Next, an optimization process is carried out to search for the best pipelines, and the best found is returned.}
    \label{figure:dynamic-seach-space-automl}
\end{figure}

Figure \ref{figure:dynamic-seach-space-automl} outlines 
the
proposed approach. In part A, the offline phase involves generating meta-knowledge using a predefined set of tasks $d_i \in D_\text{train}$. This process entails running ML 
pipelines and extracting meta-features $ f_c(D) \in C $, which are then used to train a meta-model $f_m$. This meta-model predicts the expected performance of different preprocessor $\alpha^p \in A^p$ and classifier $\alpha^c \in A^c$ combinations for a given dataset. In part B, the online phase begins when a new task is presented  $d_j \in D_\text{test}$, where $D_\text{train} \cap D_\text{test} = \emptyset$. The system computes meta-features for the dataset and uses the meta-model $f_m$ to estimate the performance of each possible preprocessor-classifier combination, i.e., $ \forall \alpha^p \in A^p, \forall \alpha^c \in A^c, \quad \langle \alpha^p, \alpha^c \rangle \in A^p \times A^c $. Preprocessor–classifier combinations with performance rankings larger 
than or equal to a threshold \(\theta\) are selected to design a reduced search space \(S^* \in \mathcal{S}\), which is then optimized to identify the best pipeline configuration \(p^* \in S^*\). The threshold \(\theta\) corresponds to a performance quantile; for example, \(\theta = 0.95\) indicates that only pipeline combinations within the top 5\% of predicted performance are included in \(S^*\).

\begin{figure}[!htb]
    \centering
    \includegraphics[width=.8\textwidth]{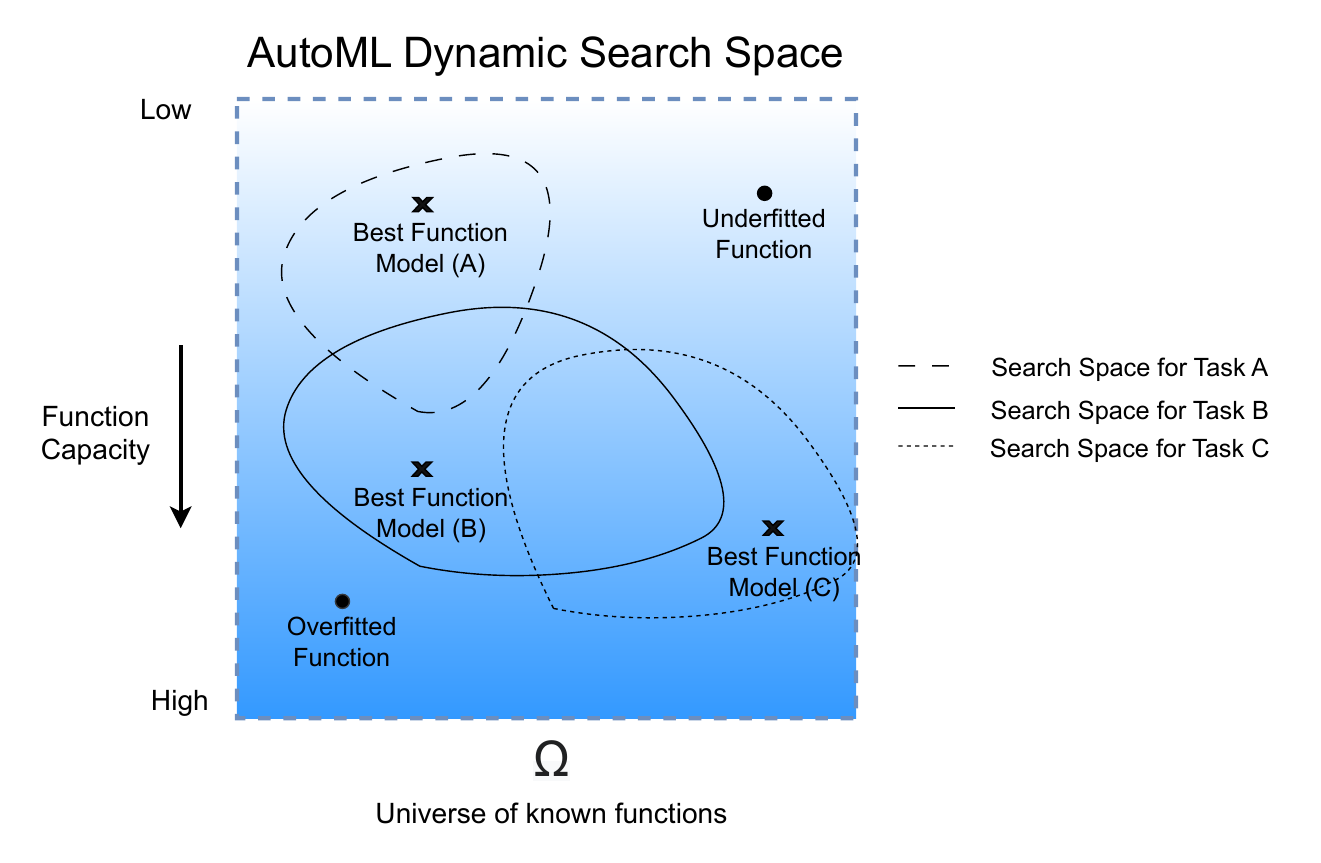}
    \caption{Theoretical representation of dynamic search space reduction for three different tasks (A, B, and C). }

    \label{figure:dynamic-seach-space-overfitting-problem-our-approach}
\end{figure}

Algorithm \ref{alg:dynamic_search_space} systematically describes how an AutoML system integrates dynamic search space generation. First, the system computes the predicted performance of all possible preprocessor-classifier combinations for a given dataset. Next, these combinations are ranked based on their predicted performance. The top-performing combinations, determined by applying a threshold $\theta$, define the tailored search space. Finally, an optimization method selects and fine-tunes the most promising pipeline candidate. Figure \ref{figure:dynamic-seach-space-overfitting-problem-our-approach} depicts how our proposed solution theoretically operates in response to the problem presented in Figure \ref{figure:dynamic-seach-space-overfitting-problem} by dynamically reducing the search space and potentially acting as a regularization mechanism. 

\begin{algorithm}[htbp]
\caption{Dynamic Search Space for AutoML}
\label{alg:dynamic_search_space}
\begin{algorithmic}[1]
\State \textbf{Input:} Dataset $d$, meta-model $f_{\mathcal{M}}$, threshold $\theta$, classifier algorithms $A^c$, preprocessor algorithms $A^p$
\State \textbf{Output:} Optimized pipeline $p^*$
    \State
    \State $m$ $\gets$  meta-features-extraction($d$)
    \State ht $\gets$  hash-table()
    \For{ $\{ \langle \alpha^p , \alpha^c \rangle : \alpha^p \in A^p , \alpha^c \in A^c \}$ }
        \State ht[$\alpha^p_i$, $\alpha^c_i$] $\gets$ $f_{\mathcal{M}}(m, \alpha^p_i, \alpha^c_i)$
    \EndFor
    \State $r_\theta$ $\gets$  quantile-filter(rank(ht), $\theta$)
    \State $S^*$ $\gets$  tailored-search-space($r_\theta$)
    \State $p^*$ $\gets$  optimize-pipelines($S^*$, $d$)
    \State \textbf{return} $p^*$
\end{algorithmic}
\end{algorithm}

Additionally, we found
a gap in existing meta-features regarding the interactions between 
classification algorithms 
and preprocessing techniques. To address this gap, we propose the use of Pipeline Statistics meta-features, which leverage historical performance data from preprocessor-classifier combinations. Equation \ref{eq:mfpc} formalizes
the extraction of these meta-features. For a given set of past ML tasks, represented by a dataset collection $D_{\text{train}}$, we estimate
the performance $\mathcal{L}$ of a specific preprocessor $\alpha^p \in A^p$ and classifier $\alpha^c \in A^c$ across multiple hyperparameters configurations. This results in a performance vector, which is
summarized using statistical functions $\sigma$, such as mean, median, standard deviation, minimum, and maximum.

\begin{gather}
    \label{eq:mfpc}
    f_{M}\big(d_j\big) = \sigma\big(\mathcal{L}(D_{\text{train}}, \alpha^p, \alpha^c))
\end{gather}

For computing the Pipeline Statistic meta-features, one can follow this procedure: (i) get historical meta-knowledge from some source, or generate meta-knowled from pipelines runs; (ii) aggregate performance scores for each dataset based at least on three component types: classifier, preprocessor, and classifier-preprocessor pairs; (iii) for each dataset-component combination, apply summaries such as mean, and standard deviation. Since this meta-knowledge is generated offline from a curated dataset collection, no additional computation is needed during the online phase, which is an advantage. However, this approach can be sensitive to the optimization budget used in creating the statistics. When combined with traditional meta-features, as demonstrated in Section \ref{result:meta_model_design}, these new meta-features provide promising results.

The next sections show that the dynamic design of search spaces offers several key advantages: (i) it enables automatic pipeline composition through metalearning, leading to more efficient search spaces; (ii) it acts as a regularization mechanism by limiting AutoML’s capacity, reducing the risk of overfitting, but still having a competitive performance; and (iii) it decreases computational costs by narrowing the search space.

\section{Methodology}
\label{sec:methodology-dp}

This section details the methodological framework employed in the study, including the dataset selection and split process, the design of the ML pipeline configuration space, meta-feature extraction and the general experimental setup.

\subsection{Datasets}
For the experiments reported in this study, 197 datasets from a wide range of domains to capture the diversity of classification problems were collected. These datasets were sourced from OpenML~\cite{vanschoren-2013} and selected based on curated collections used in prior AutoML research~\cite{feurer-2022, gijsbers-2024}. Each dataset was randomly partitioned 
into a training subset, $D_{\text{train}}$, with 75\% of the original dataset, and a test subset, $D_{\text{test}}$, with the remaining dataset.
The OpenML IDs and associated metadata for these datasets are provided in Table \ref{tab:sup-dataset-md-train} (train) and Table \ref{tab:sup-dataset-md-test} (test) in the Appendix \ref{sec:appx-datasets-split-train-test}.

\subsection{Meta-feature Extraction}
We systematically extracted meta-features using \textit{pymfe} package~\cite{alcobacca-2020}. We used the general, information-theoretic, model-based, statistical, and landmarking groups.
For the summary part of the systematic extraction, mean and standard deviation were employed.

We extracted the Pipeline Statistic meta-features using the following procedure: (i) we utilized historical meta-knowledge from~\cite{alcobaca-2025a}; (ii) We aggregated F1 scores for each dataset based on three component types: classifier, feature preprocessor, and classifier-preprocessor pairs; (iii) For each dataset-component combination, we extracted statistical summaries, including the minimum, maximum, mean, median, and standard deviation of the F1 performance. It is important to notice that these meta-features were only extracted using the $D_{\text{train}}$ to avoid data leakage.

\subsection{Configuration Space}
The configuration space defines the spectrum of possible pipeline architectures, encompassing components that can be optimized during the search process. We used the same search space proposed in ~\cite{alcobaca-2025a}, which follows the hierarchical tree-based structure by~\cite{feurer-2015a}. Thus, the configuration space consists of three primary components: model selection, feature preprocessing, and data preprocessing.

Data preprocessing was applied as needed to ensure compatibility with scikit-learn implementations. This included one-hot encoding for categorical variables, imputation of missing values, and rescaling of features through standardization or normalization. Therefore, this step is fixed and is applied when required.

For feature preprocessing, 13 techniques reflecting a wide range of strategies were incorporated. These include dimensionality reduction, feature selection, feature generation, embedding-based methods, model-based selection, as well as a \textit{no preprocessing} option. The modeling component includes 16 classification algorithms spanning various methodological families: (i) distance-based, linear models, neural networks, kernel-based, probabilistic, tree-based, and ensemble methods. To simplify, we use the term ``preprocessing" as a synonym for ``feature preprocessing" in this paper. Table \ref{fig:feature-data-class-used} in the Appendix \ref{app:feature-data-class-used}, shows a detailed list of algorithms used.

\subsection{Experimental Setup}
To assess the base-level performance, we utilized the pipeline runs provided by \citeauthor{alcobaca-2025a} benchmark, simulating a Random Search optimization with $500 \times 10$ pipeline samples by dataset. Each dataset was split into training (75\%) and test (25\%), being validation performance assessed by 10-fold cross-validation on the training part. Performance was assessed using F1-weighted score, which can take into account imbalanced datasets.

To asses the meta-level, we tested five meta-learners: Random Forest (RF), Support Vector Machine (SVM), k-Nearest Neighbors (kNN), Decision Tree (DT), and Multi-Layer Perceptron (MLP). These algorithms were selected as they represent diverse inductive biases in ML.
The evaluation metrics used were relative root mean squared error (RRMSE), root mean squared error (RMSE), and coefficient of determination \( R^2 \). RMSE provides a direct measure of the average prediction error magnitude, while RRMSE complements it by adding the mean baseline. Finally, $R^2$ quantifies the proportion of variance in the observed data that is explained by the model.

We used the Friedman-Nemenyi test~\cite{demsar-2006} to assess the statistical significance of differences between methods evaluated in this study. Moreover, we conducted 10 experimental repetitions to account for stochasticity.

All experiments were conducted using scikit-learn algorithms~\cite{pedregosa-2011}. Each pipeline was restricted to a maximum runtime of 600 seconds and each dataset was allotted up to 24 hours. Experiments were run on a Debian Linux system with Intel Xeon E5-2680v2 CPUs, 128~GB of DDR3 RAM, and a limit of 10~GB memory and a single core per pipeline execution.

To promote reproducibility and transparency, all experimental analysis, tools and code are available on GitHub\footnote{GitHub Repository: \href{https://github.com/ealcobaca/dynamic-design-machine-learning-pipelines}{https://github.com/ealcobaca/dynamic-design-machine-learning-pipelines}}. Additionally, the dataset containing all pipeline configurations, execution times, and performance results provided by \cite{alcobaca-2025a} is available online\footnote{Dataset available at \href{https://figshare.com/articles/dataset/Meta-datasets/28696262}{figshare}.}.

\section{Results and Discussion}
\label{sec:results-and-discussion}
In this section, we present the experimental results and analysis. The discussion is structured into four main sub-sections: dataset characteristics in Section \ref{result:dataset-characteristics}, meta-level analysis in Section \ref{result:meta-level-analysis}, dynamic pipeline design in Section \ref{result:pipeline-design}, and dynamic search space on Auto-sklearn in Section \ref{result:dynamic-search-space}.

We begin by analyzing the dataset characteristics for both \( D_{\text{train}} \), used to train the meta-model, and \( D_{\text{test}} \), used for evaluating the dynamic pipeline design in AutoML systems. The analysis covers the number of features, the number of instances, the number of classes, and the percentage of the minority class.

Next, we examine the meta-level, focusing on the computational cost of meta-feature extraction and the development of a meta-model for pipeline design. We also analyze the explainability of the meta-model to understand what it has learned.

Finally, we evaluate the meta-model within an AutoML system, where it is used to dynamically generate search spaces for unseen datasets. A Random Search strategy is applied for hyperparameter tuning. The proposed approach is compared against multiple baselines, assessing both performance and computational efficiency. Additionally, we analyze the inclusion of dynamic search spaces in Auto-Sklearn AutoML, which uses Bayesian Optimization with metalearning, warm start, and ensemble steps.

We structure our discussion around the following research questions, which will be addressed in the subsequent subsections:

\begin{itemize}
    \item \textbf{Q1}: What are the characteristics of the datasets (\(D_{\text{train}}\)) used as meta-knowledge for training the meta-model and the datasets (\(D_{\text{test}}\)) used to evaluate the AutoML approach?
    
    \item \textbf{Q2.1}: What is the computational cost of the extracted group of meta-features?
    
    \item \textbf{Q2.2}: Can we design a meta-model capable of predicting the performance of preprocessor-classifier combinations?
    
    \item \textbf{Q2.3}: What information does the meta-model learn from the meta-features?
    
    \item \textbf{Q3.1}: Can metalearning dynamically create search spaces for ML 
    pipeline design?
    
    \item \textbf{Q3.2}: Can dynamically generated search spaces effectively reduce optimization time?
    
    \item \textbf{Q3.3}: What are the most common search space recommendations made by the meta-model?
    \item \textbf{Q4}: Can we use dynamic search spaces in Auto-Sklearn to efficiently reduce the search space and overfitting?
\end{itemize}

\subsection{Dataset Characteristics}
\label{result:dataset-characteristics}

\begin{figure}[!htb]
    \centering
    \includegraphics[width=0.75\textwidth]{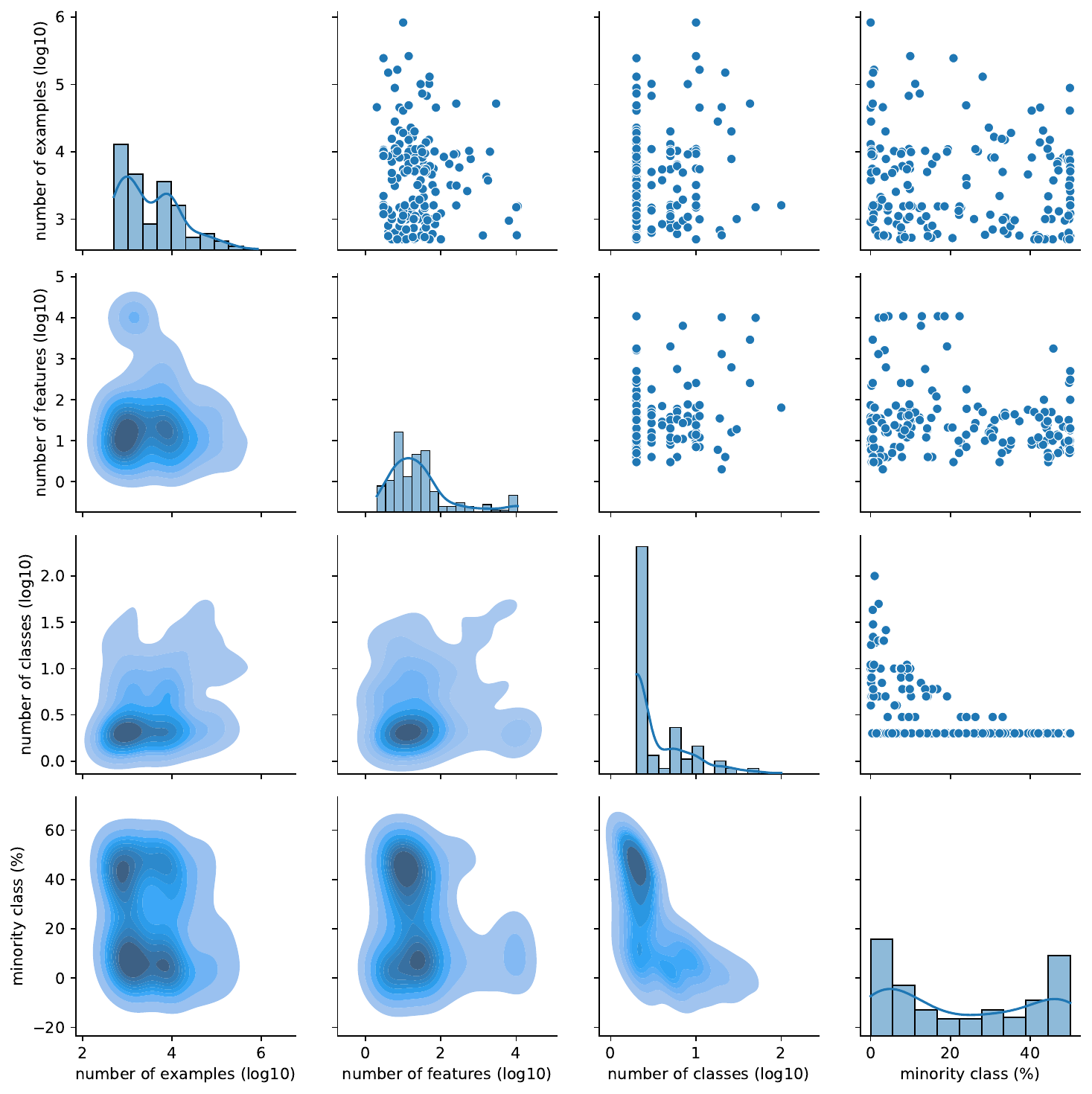}
    \caption{Pairwise plots summarizing dataset characteristics in Dataset set  \(D = D_{\text{train}} \cup D_{\text{test}}\). The diagonal plots show the distribution of each characteristic: the number of examples, number of features, number of classes (all on a \(\log_{10}\) scale), and the percentage of the minority class. The upper triangle contains scatter plots, where each point represents a dataset, while the lower triangle displays kernel density estimation plots.}
    \label{datasets_pair_grid_all}
\end{figure}

To address research question \textbf{Q1}, we analyzed the characteristics of both \( D_{\text{train}} \), used for training the meta-model, and \( D_{\text{test}} \), used for evaluating the AutoML system. Figure \ref{datasets_pair_grid_all} presents pairwise plots summarizing the characteristics of Dataset set \(D = D_{\text{train}} \cup D_{\text{test}}\). The analyzed characteristics include the number of features, number of examples, and number of classes, all on a \(\log_{10}\) scale, as well as the percentage of the minority class. The diagonal plots show the distributions of each characteristic. The upper triangle contains scatter plots, where each point represents a dataset, while the lower triangle displays kernel density estimation plots for pairwise relationships between characteristics.

The diagonal distributions show that most datasets have between \(10^3\) and \(10^4\) examples, with fewer exceeding \(10^4\). Most have around \(10^1\) features, while datasets with \(10^3\) or \(10^4\) features are uncommon. Most datasets are binary classification problems, though some contain between \(10^1\) and \(10^2\) classes. The percentage of minority class varies widely, with a high concentration near 0\% and 50\%.

The pairwise comparisons reveal that the number of examples and the minority class percentage, as well as the number of features and the minority class percentage, are widely spread. There are few datasets with both a high number of classes and a high minority class percentage. The number of features and examples paired with the number of classes showed a high concentration on problems with a low number of classes. Additionally, datasets with both a high number of examples and a high number of features are underrepresented.

These observations highlight gaps in datasets commonly used in AutoML and metalearning, particularly a scarcity of datasets with a high number of classes and those with both high numbers of classes and features.

\begin{figure}[!htb]
    \centering
    \includegraphics[width=0.75\textwidth]{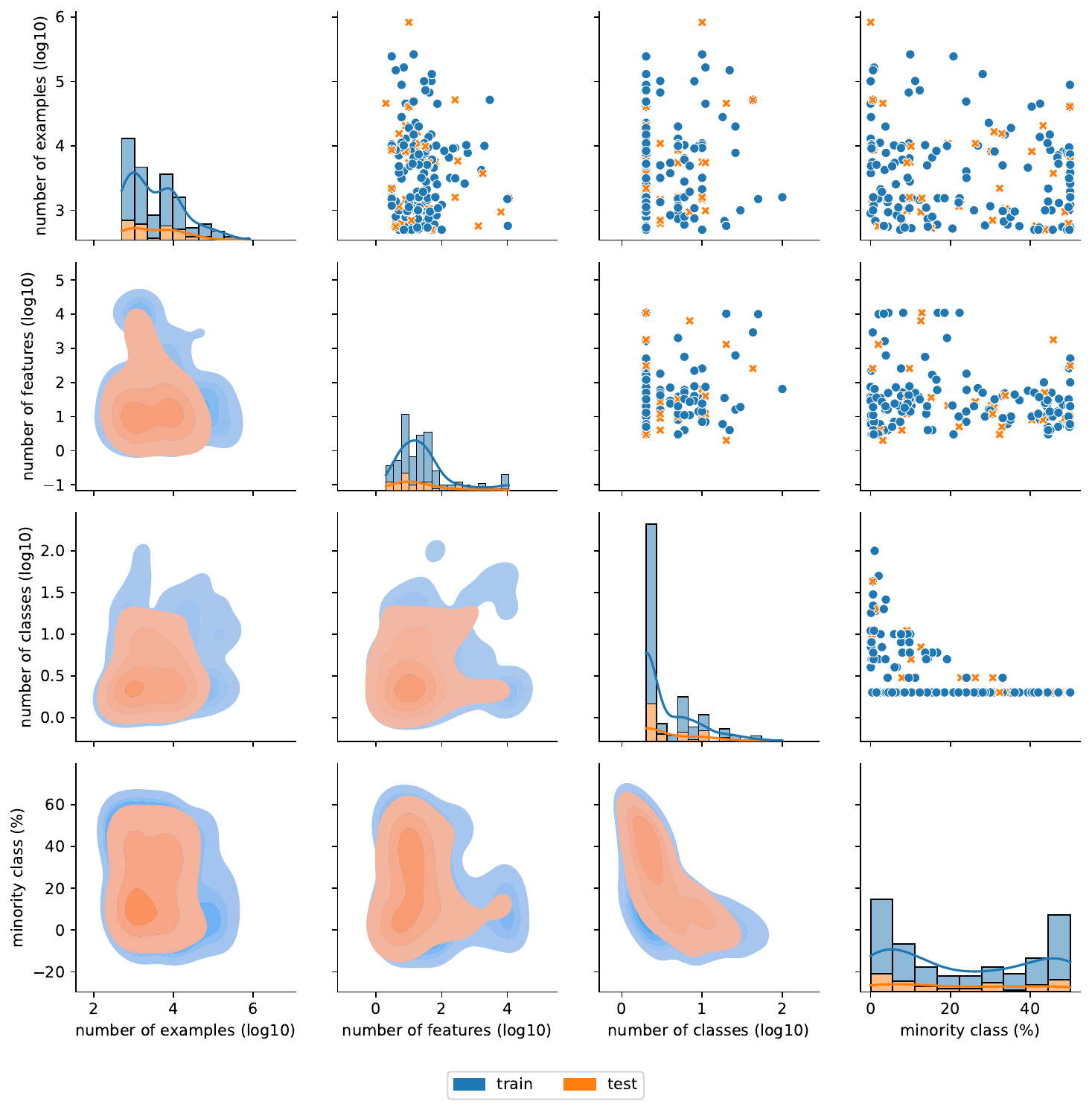}
    \caption{Pairwise plots summarizing dataset characteristics in \(D_{\text{train}}\) (blue) and  \(D_{\text{test}}\) (orange). The diagonal plots show the distribution of each characteristic: the number of examples, number of features, number of classes (all on a \(\log_{10}\) scale), and the percentage of the minority class. The upper triangle contains scatter plots, where each point represents a dataset, while the lower triangle displays kernel density estimation plots.}
    \label{datasets_pair_grid_group}
\end{figure}

Similarly to Figure \ref{datasets_pair_grid_all}, Figure \ref{datasets_pair_grid_group} presents pairwise plots summarizing dataset characteristics, with \(D_{\text{train}}\) in blue and \(D_{\text{test}}\) in orange. The datasets were split using a shuffled train/test split, allocating 80\% (157 datasets) to metalearning training and 20\% (40 datasets) to testing from the total of 197 datasets.

The diagonal plots show that the distributions of dataset characteristics are visually similar in \(D_{\text{train}}\) and \(D_{\text{test}}\). The pairwise kernel density estimation plots indicate that the orange surface (test set) largely overlaps the blue surface (train set), with few uncovered points. This suggests that the train-test split provides a representative distribution of the dataset characteristics, minimizing bias in the partitioning. Similar plots but with contour curves can be seen in Appendix \ref{app:pairwise-plots-contour}.

\subsection{Meta-level Analysis}
\label{result:meta-level-analysis}
In this section, we present the meta-level results. First, we analyze the computation time of meta-features, a key aspect of our approach, in Section~\ref{result:meta_feature_time_spent}. Second, we describe the meta-model design and the variations tested in Section~\ref{result:meta_model_design}. Finally, we conduct an explainability study on the impact of meta-features on the meta-model’s predictions in Section~\ref{result:meta_model_explainability}.

\subsubsection{Meta-features time spent}
\label{result:meta_feature_time_spent}

To address research question \textbf{Q2.1}, we analyzed the computation time of different meta-feature groups used in the meta-models, as search space design time is a key component of our approach. Pipeline statistics meta-features were excluded from this analysis, as they incur no computational cost during the online phase of the metalearning system, i.e., when making recommendations for unseen tasks.

\begin{figure}[!htb]
    \centering
    
    \begin{subfigure}[t]{0.49\textwidth}
        \subcaption{}
        \includegraphics[width=\linewidth]{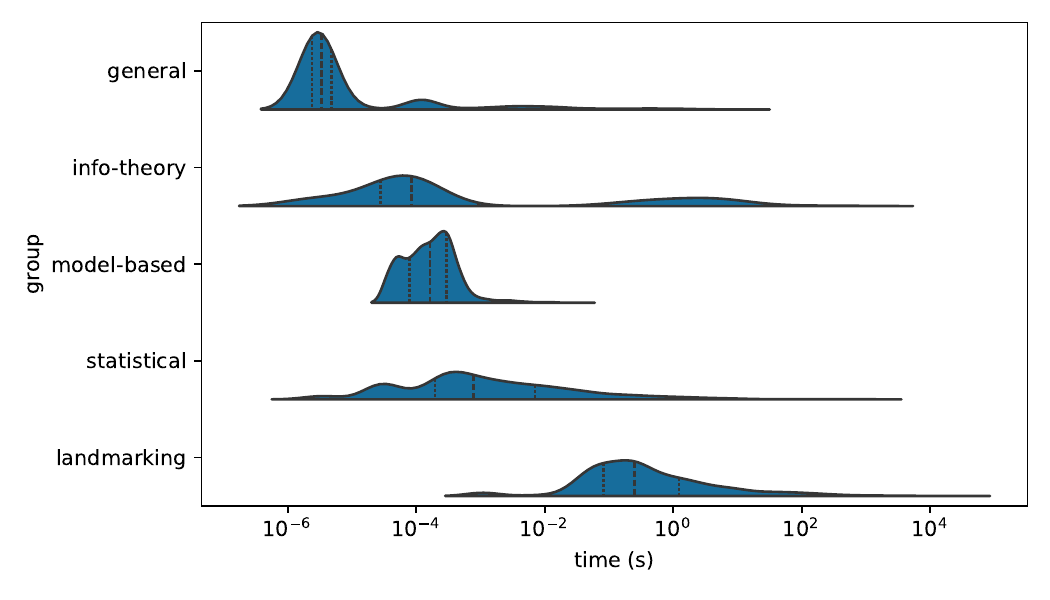}
    \end{subfigure}
    \hspace{2pt}
    \begin{subfigure}[t]{0.49\textwidth}
        \subcaption{}
        \includegraphics[width=\linewidth]{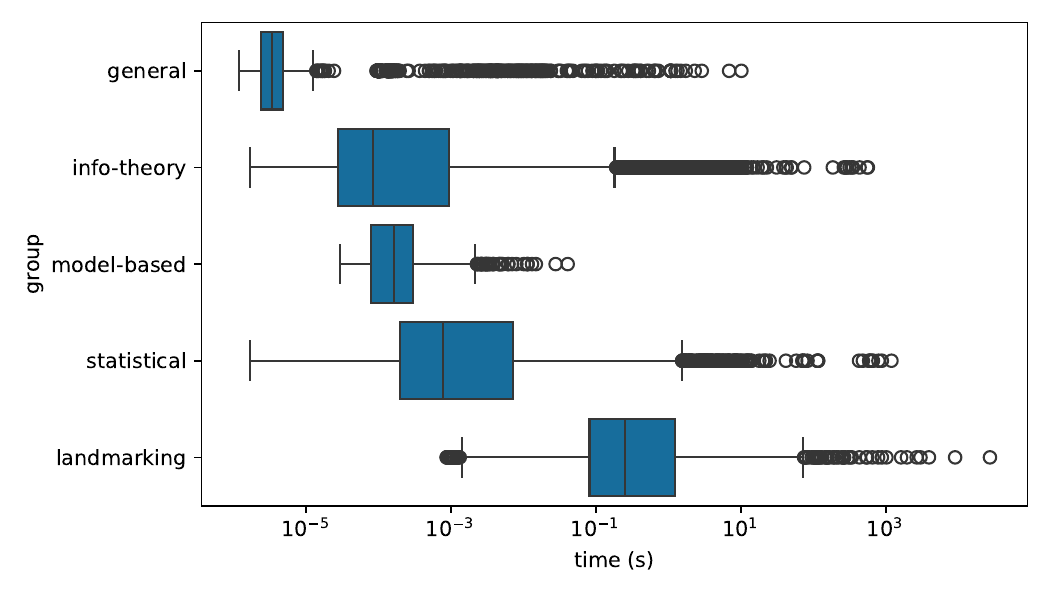}
    \end{subfigure}

    \vspace{2pt} 

    \begin{subfigure}[t]{0.49\textwidth}
        \subcaption{}
        \includegraphics[width=\linewidth]{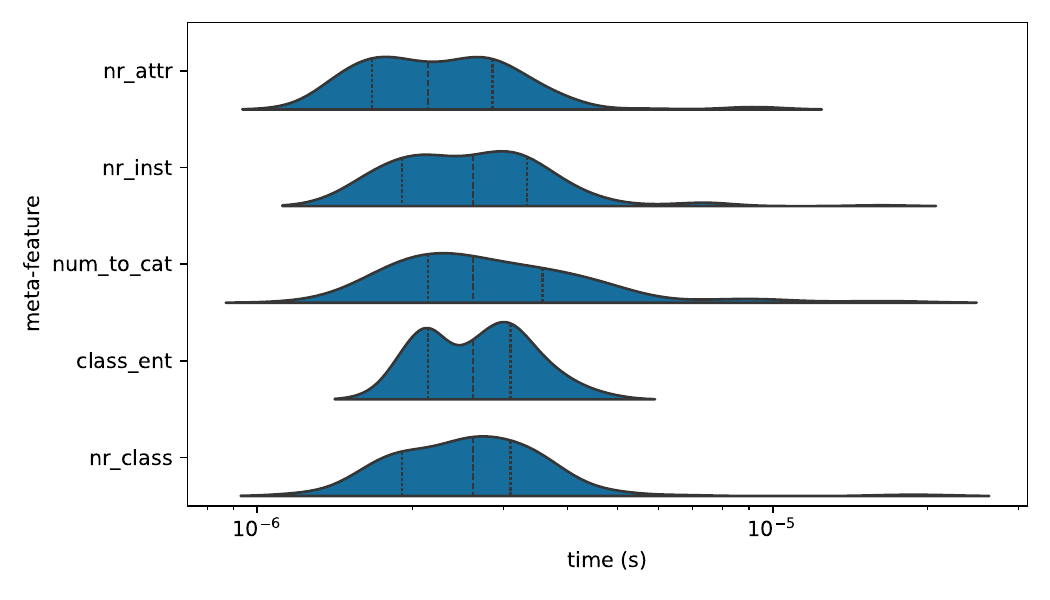}
    \end{subfigure}
    \hspace{2pt}
    \begin{subfigure}[t]{0.49\textwidth}
        \subcaption{}
        \includegraphics[width=\linewidth]{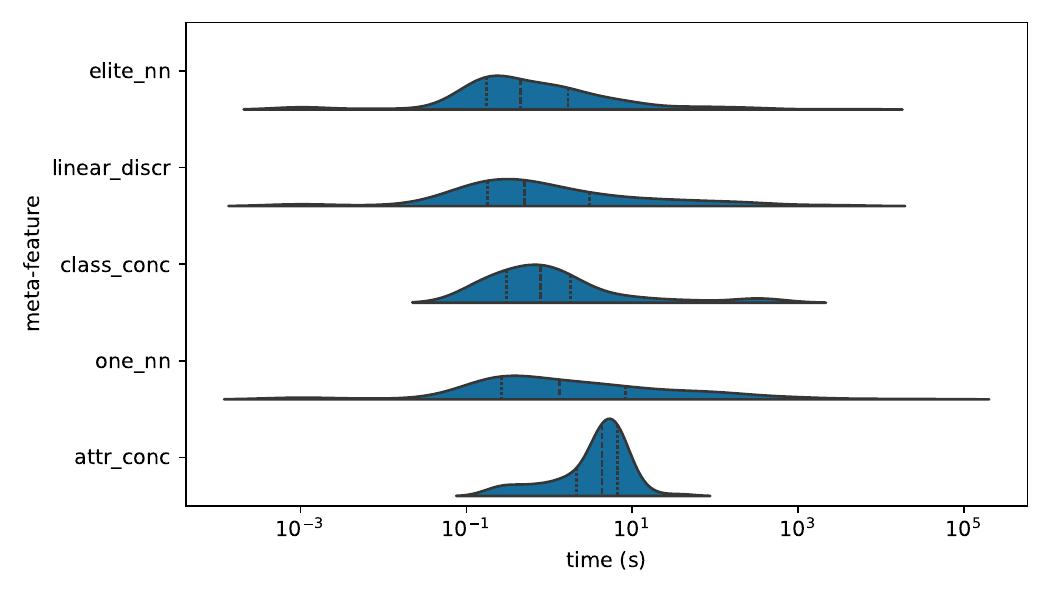}
        \label{asd}
    \end{subfigure}

    \caption{Computation time distribution of meta-feature groups across datasets. (A) Distribution of each meta-feature group, with dashed lines representing the first quartile, median, and third quartile. (B) Box plot of computation times, ordered by median. (C) and (D) Show the five fastest and slowest meta-features, respectively.}
    \label{fig:meta-feature-time-spent}
\end{figure}

Figure \ref{fig:meta-feature-time-spent}a presents the distribution of computation time for General, Info-Theory, Model-Based, Statistical, and Landmarking meta-features across datasets on a logarithmic scale (log10).  Figure \ref{fig:meta-feature-time-spent}b presents a box plot of computational time ordered by median values. The General meta-features are the fastest, with computation costs below \(10^{-5}\) seconds. Info-Theory and Model-Based meta-features have a median around \(10^{-4}\) seconds, though Info-Theory exhibits high dispersion, while Model-Based has the lowest. Statistical meta-features have a median close to \(10^{-3}\) seconds and the highest dispersion. Landmarking meta-features are the most time-consuming, with a median close to \(10^{-1}\) seconds.

Figures  \ref{fig:meta-feature-time-spent}c and  \ref{fig:meta-feature-time-spent}d show the five fastest and slowest meta-features, respectively. The top 5 fastest include: (i) number of features (\texttt{nr\_attr}), (ii) number of examples (\texttt{nr\_inst}), (iii) ratio of numerical to categorical features (\texttt{num\_to\_cat}), (iv) Shannon’s entropy of class distribution (\texttt{class\_ent}), and (v) number of classes (\texttt{nr\_class}). Except for \texttt{class\_ent}, which belongs to the Info-Theory group, all are from the General group.

The top 5 slowest meta-features include: (i) concentration coefficient between pairs of attributes (\texttt{attr\_conc}), (ii) performance of the 1-Nearest Neighbor classifier (\texttt{one\_nn}), (iii) concentration coefficient between attributes and class (\texttt{class\_conc}), (iv) performance of the Linear Discriminant classifier (\texttt{linear\_discr}), and (v) performance of the Elite Nearest Neighbor classifier (\texttt{elite\_nn}). While \texttt{one\_nn}, \texttt{elite\_nn}, and \texttt{linear\_discr} belong to the Landmarking group, \texttt{class\_conc} and \texttt{attr\_conc} are from the Info-Theory group.

To assess whether specific dataset characteristics influence meta-feature computation time, we calculated the Pearson correlation between dataset characteristics (as analyzed in Section~\ref{result:meta_feature_time_spent}) and meta-feature groups.

\begin{figure}[!htb]
    \centering
     \makebox[\textwidth]{\includegraphics[scale=0.50]{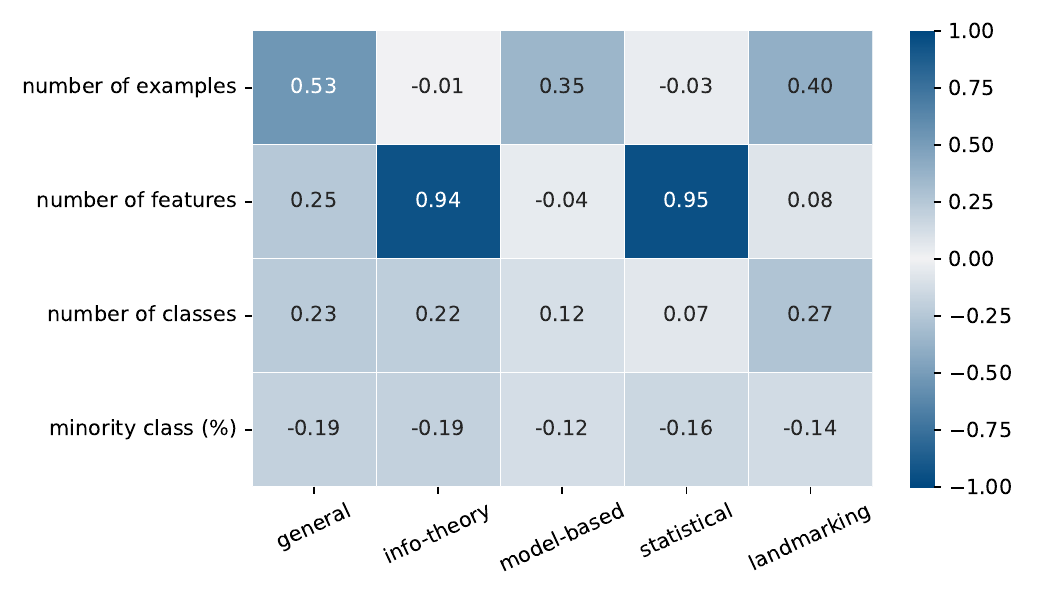}}
    \caption{Steps of the machine learning pipeline considered. It is shown the first levels of search space with algorithms available.}
    \label{fig:correlation-meta-feature-dataset}
\end{figure}

Figure \ref{fig:correlation-meta-feature-dataset} presents a correlation heatmap between dataset characteristics and meta-feature groups. A strong correlation (\(\geq 0.94\)) is observed between the number of features and both Info-Theory and Statistical meta-features, indicating that an increase in the number of features significantly impacts their computation time. Additionally, a moderate correlation (\(\geq 0.40\)) is found between the number of examples and both General and Landmarking meta-features.

To ensure that the proposed meta-model can efficiently recommend suitable pipeline compositions, we excluded computationally expensive meta-features. Specifically, only meta-features with a median computation time below \(10^{-1}\) seconds were retained. The complete list of removed meta-features is provided in Appendix~\ref{app:time-spent-analysis-all-metafeatures}.

\subsubsection{Meta-model design}
\label{result:meta_model_design}

In this subsection, we evaluate the proposed meta-model from three perspectives, addressing research question \textbf{Q2.2}. First, we analyze the performance of different meta-learners. Second, we assess the impact of meta-feature groups. Finally, we investigate the minimum number of meta-features required to maintain performance.

The meta-model proposed in this study aims to predict the performance of a composition of a classifier and a preprocessor. The input features consist of five different meta-feature groups, along with pipeline statistics meta-features introduced in this work, and two numerical features representing the classifier and preprocessor. The target variable is the highest F1-weighted score achieved by a given combination of preprocessing and classification for each dataset after a Random Search run. Thus, each sample corresponds to a specific combination of dataset, preprocessor, and classifier. The problem is formulated as a regression task. Mathematically, the meta-model maps the function \( f: C \times A^p \times A^c \rightarrow R \), where \( C \) represents the meta-features set, \( A^p \) the preprocessor set, \( A^c \) the classifier set, and \( R \) the F1 performance of the preprocessing-classifier combination. Unlike traditional metalearning, which recommends the best algorithm, our approach builds a meta-model, such as a surrogate, capable of estimating the performance of different combinations.

\begin{figure}[!htb]
    \centering
    \begin{minipage}{0.48\textwidth}
        \subcaption{}
        \includegraphics[width=\textwidth]{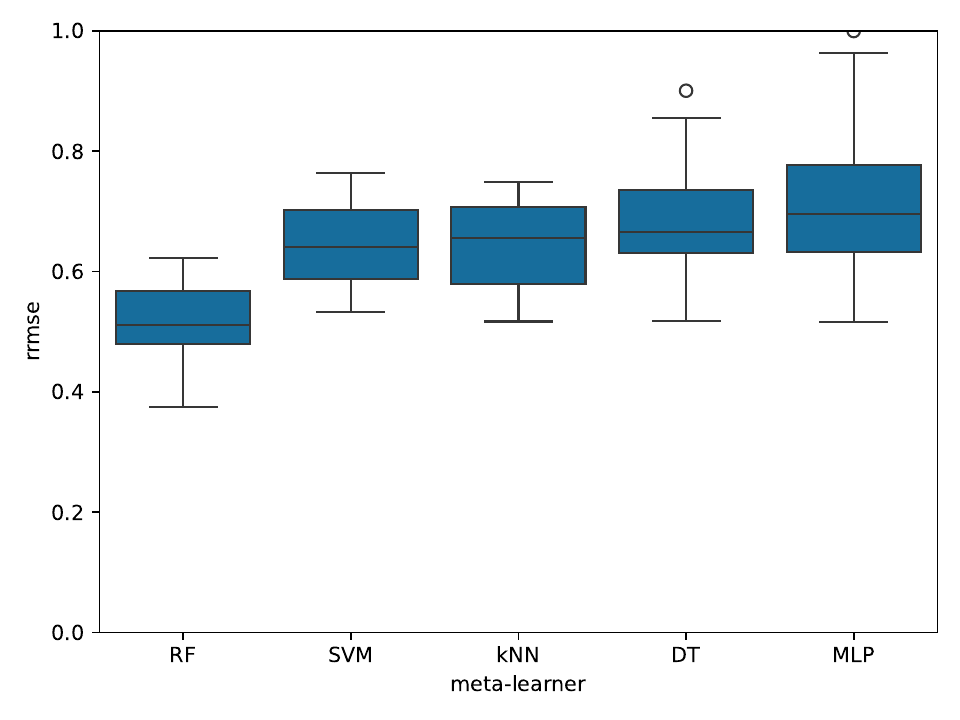}
    \end{minipage}
    \begin{minipage}{0.48\textwidth}
        \subcaption{}
        \includegraphics[width=\textwidth]{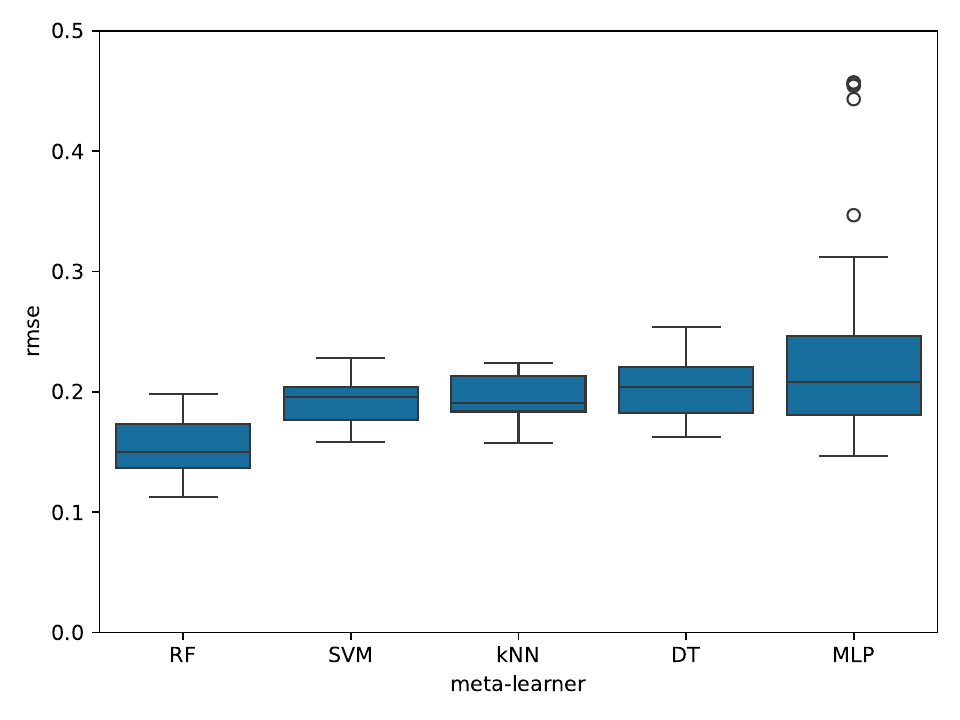}
    \end{minipage}
    \begin{minipage}{0.48\textwidth}
        \subcaption{}
        \includegraphics[width=\textwidth]{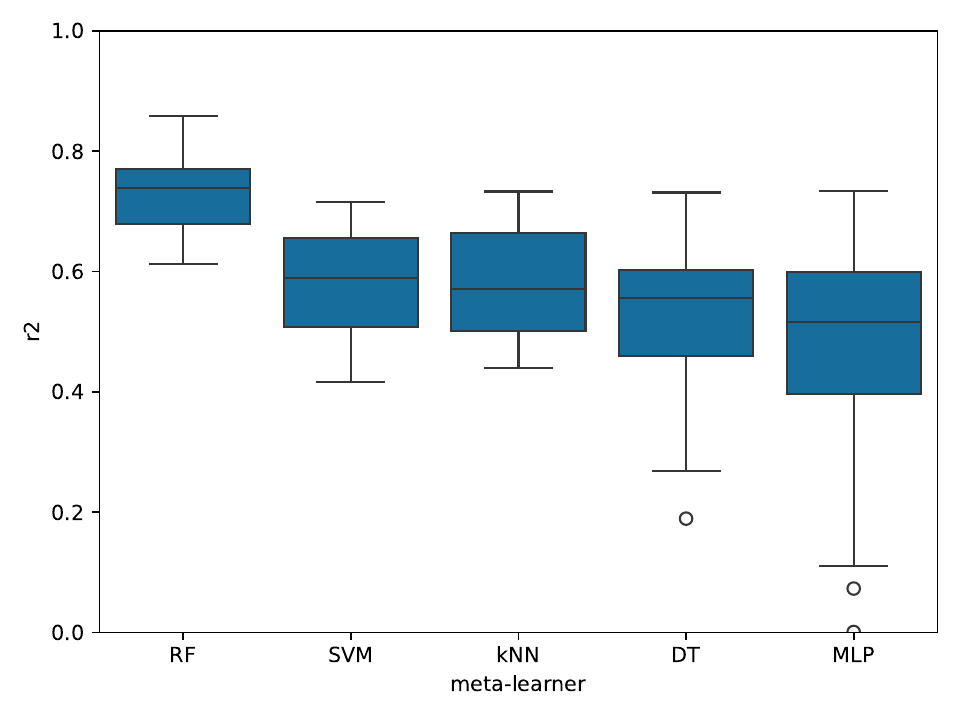}
    \end{minipage}
    \caption{Cross-validation performance of different meta-learners over 10 repetitions on \( D_{\text{train}} \). The evaluated meta-learners include Random Forest (RF), Support Vector Machine (SVM), k-Nearest Neighbors (kNN), Decision Tree (DT), and Multi-Layer Perceptron (MLP). Performance was assessed using RRMSE (a), RMSE (b), and \( R^2 \) (c). }
    \label{fig:meta-learners-performance}
\end{figure}

Figure \ref{fig:meta-learners-performance} presents the cross-validation performance of the proposed approach over 10 repetitions using different meta-learners on \( D_{\text{train}} \). We tested five meta-learners: Random Forest (RF), Support Vector Machine (SVM), k-Nearest Neighbors (kNN), Decision Tree (DT), and Multi-Layer Perceptron (MLP). These algorithms were selected as they represent diverse inductive biases in ML.
The evaluation metrics used were RRMSE (a), RMSE (b), \( R^2 \) (c). Lower values of RMSE and RRMSE indicate better performance, while higher \( R^2 \) values are preferable. The figure shows that RF achieved the highest performance across all three metrics. Additionally, since RRMSE incorporates the mean baseline—where predictions equal to 1 indicate performance equivalent to predicting the median target value—we can conclude that the meta-learners produced models that outperform the mean baseline. The respective mean and standard deviation for RF in RRMSE, RMSE and R2 \( R^2 \) are \( 0.51 \pm 0.07 \), \( 0.15 \pm 0.02 \) and \( 0.73 \pm 0.07 \).

\begin{figure}[!htb]
    \centering
    \begin{minipage}{0.48\textwidth}
        \subcaption{}
        \includegraphics[width=\textwidth]{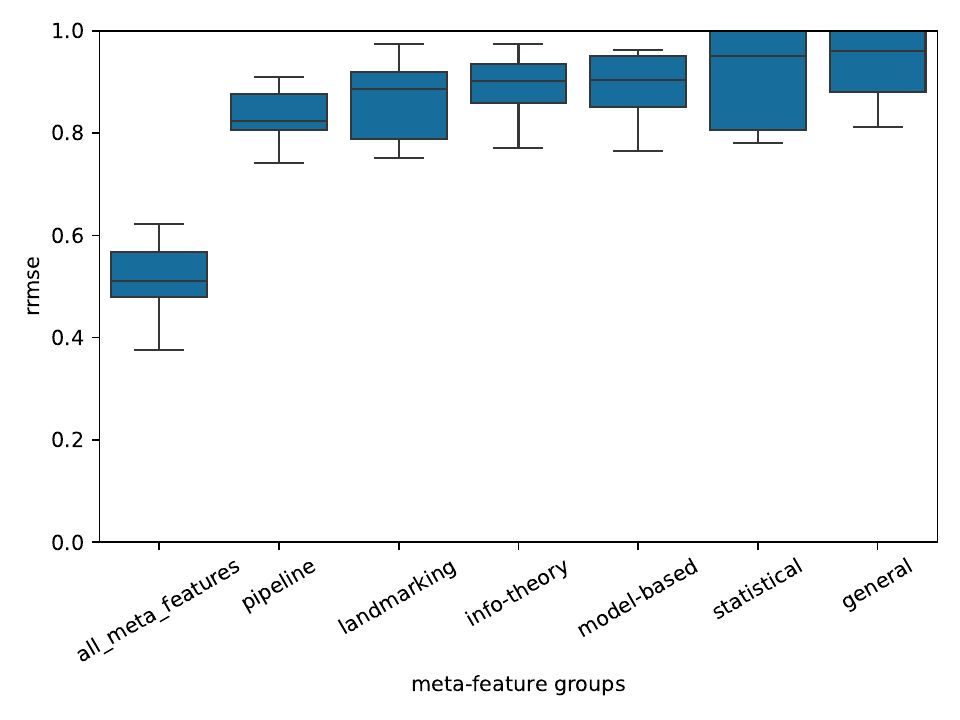}
    \end{minipage}
    \begin{minipage}{0.48\textwidth}
        \subcaption{}
        \includegraphics[width=\textwidth]{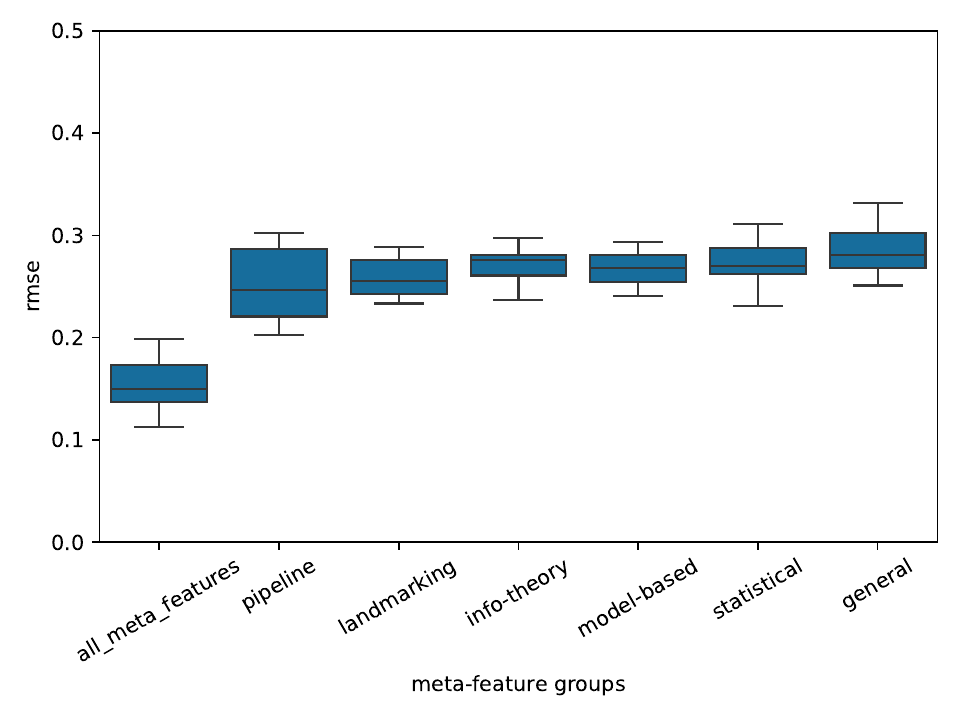}
    \end{minipage}
    \begin{minipage}{0.48\textwidth}
        \subcaption{}
        \includegraphics[width=\textwidth]{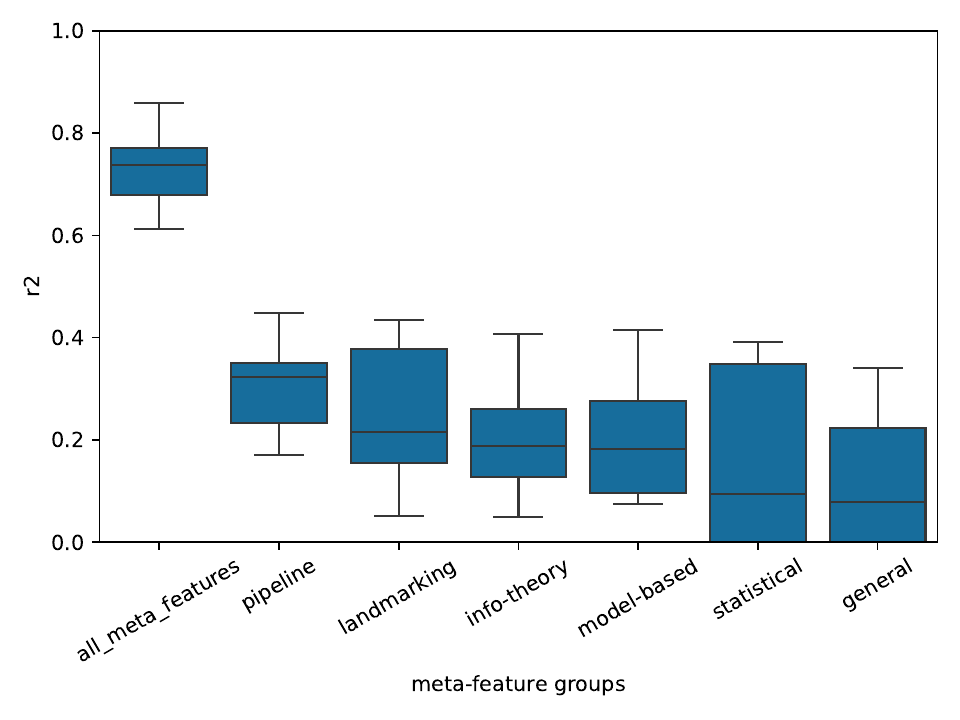}
    \end{minipage}
    \caption{Cross-validation performance of different meta-feature groups over 10 repetitions on \( D_{\text{train}} \), using RF as the fixed meta-learner. The tested groups include General, Info-Theory, Statistical, Model-Based, Landmarking, and Pipeline Statistics (denoted as Pipeline), as well as a combination of all groups. Models were evaluated using RRMSE (a), RMSE (b), and \( R^2 \) (c).}
    \label{fig:meta-feature-groups-performance}
\end{figure}

Figure \ref{fig:meta-feature-groups-performance} presents the cross-validation performance of different meta-feature groups over 10 repetitions on \( D_{\text{train}} \), using RF as the fixed meta-learner. Across all metrics, Pipeline Statistics achieved the best median performance, followed by Landmarking and Info-Theory. However, the combination of all meta-feature groups resulted in a higher overall median, indicating that integrating diverse meta-feature groups improves the predictive performance of the meta-model.  

\begin{figure}[!htb]
    \centering
    \begin{minipage}{0.48\textwidth}
        \captionsetup{justification=raggedright, singlelinecheck=false}
        \subcaption{}        \includegraphics[width=\textwidth]{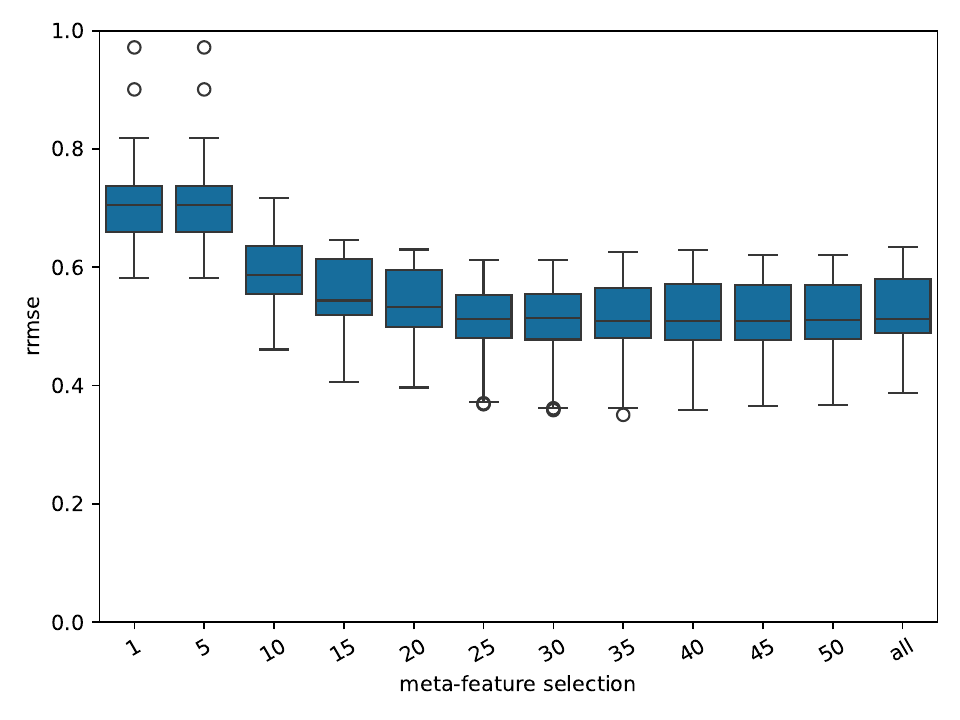}
    \end{minipage}
    \begin{minipage}{0.48\textwidth}
        \captionsetup{justification=raggedright, singlelinecheck=false}
        \subcaption{}        \includegraphics[width=\textwidth]{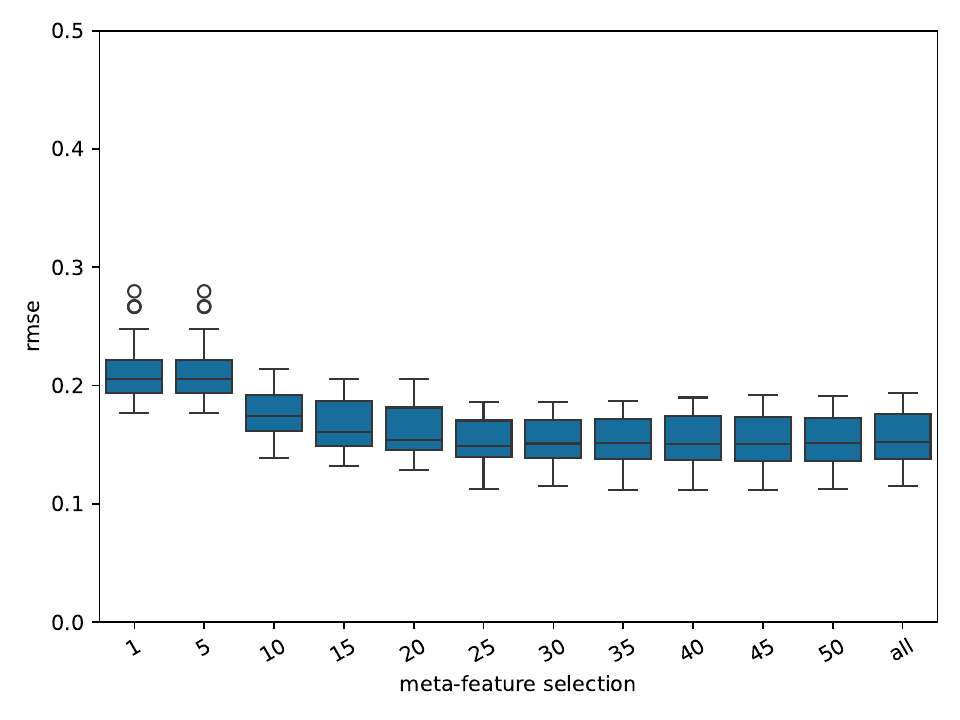}
    \end{minipage}
    \begin{minipage}{0.48\textwidth}
        \captionsetup{justification=raggedright, singlelinecheck=false}
        \subcaption{}        \includegraphics[width=\textwidth]{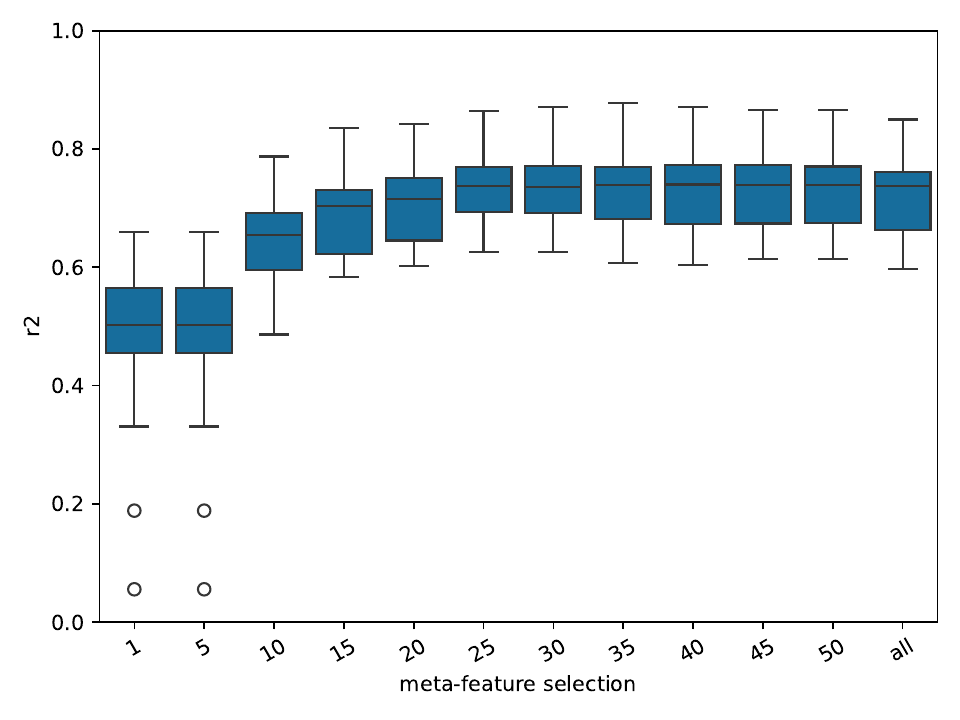}
    \end{minipage}

    \caption{Cross-validation performance with varying numbers of meta-features over 10 repetitions on \( D_{\text{train}} \), using Random Forest as the fixed meta-learner. Meta-features were ranked based on their importance scores from Random Forest. Models were evaluated using RRMSE (a), RMSE (b), and \( R^2 \) (c).}
    \label{fig:meta-feature-selection-performance}
\end{figure}

Figure \ref{fig:meta-feature-selection-performance} shows the cross-validation performance with varying numbers of meta-features over 10 repetitions on \( D_{\text{train}} \), using RF as the fixed meta-learner. Meta-features were ranked by feature importance scores from RF. The results indicate that no individual meta-feature or small subset (e.g., 1 or 5 features) achieves competitive performance. Performance stabilizes across all metrics when using at least 25 meta-features.  

Based on these results, we selected RF regressor as the meta-model, incorporating all meta-feature groups but retaining only the top 25 features ranked by importance.

\subsubsection{Meta-model Explainability}
\label{result:meta_model_explainability}

SHapley Additive exPlanations (SHAP) is a method for interpreting ML 
models by assigning each feature a contribution value based on Shapley values from cooperative game theory \cite{lundberg-2020}. It provides a fair way to explain predictions by quantifying the impact of each feature, with the sum of all feature contributions equal to the difference between a model’s prediction and the average prediction. SHAP was used to address research question \textbf{Q2.3}.

\begin{figure}[!htb]
    \centering
    \includegraphics[width=0.70\textwidth]
    {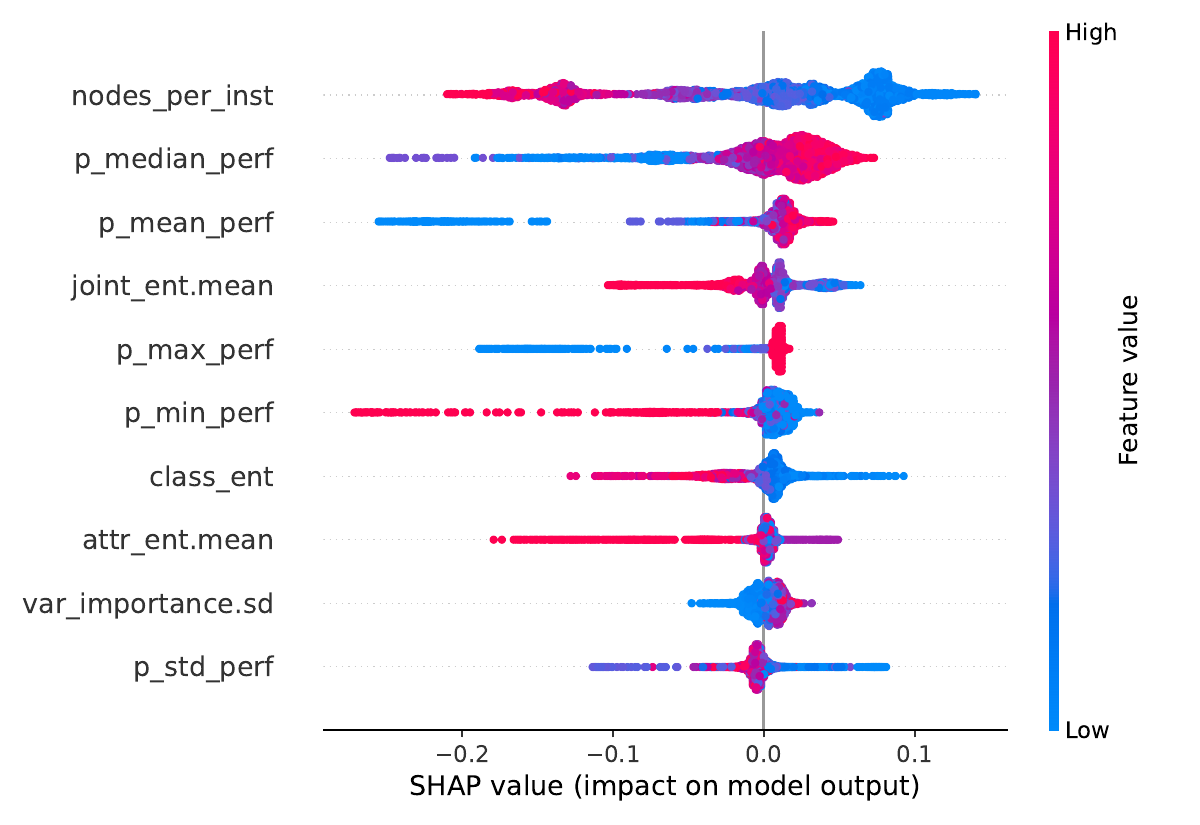}
    \caption{ SHAP summary plot showing the contribution of top 10 meta-feature to the meta-model's predictions. Each point represents a SHAP value for a specific feature and instance, with color indicating the feature’s value. Features are ordered by their average absolute SHAP value, highlighting the most influential ones. Positive SHAP values indicate an increase in the prediction, while negative values indicate a decrease.}
    \label{result:shapp-analysis}
\end{figure}

Figure \ref{result:shapp-analysis} presents a SHAP summary plot illustrating
the contribution of top 10 meta-feature to the model's predictions across all instances in the dataset. Each point on the plot corresponds to a SHAP value for a feature and instance, with color indicating the feature’s value. Features are ordered by the average absolute SHAP value, highlighting those with the 
highest
impact on predictions. Positive SHAP values correspond to an increase in the prediction, while negative values indicate a decrease.

The most influential feature is the ratio of non-leaf nodes per instance in a decision tree model (\textit{nodes\_per\_inst}). Higher values of \textit{nodes\_per\_inst} are associated with a negative contribution to the prediction. Additionally, the meta-features \textit{p\_median\_perf} and \textit{p\_mean\_perf}, which represent the median and mean performance statistics for a pipeline combination, show positive contributions to the target when their values are higher.

Entropy-based meta-features from the Info-Theory group appear three times among the top 10 features: mean joint entropy between each attribute and the class (\textit{joint\_ent.mean}), mean target attribute Shannon entropy (\textit{class\_ent}), and mean Shannon entropy of predictive attributes (\textit{attr\_ent.mean}). Higher entropy values are associated with negative contributions to the target. The full plot with all 25 meta-features is available in Appendix \ref{app:shap-summary-plot-meta-model}.

Therefore, the meta-model integrates three key aspects: (i) the complexity of tree-based algorithms through \textit{nodes\_per\_inst}, (ii) the historical statistical performance of pipeline combinations, and (iii) entropy-based information from the task, where higher entropy indicates higher 
unpredictability.

\subsection{Pipeline Design}
\label{result:pipeline-design}

This section investigates the use of metalearning to dynamically create search spaces for AutoML and assess whether it reduces search space size, thereby saving computational time. Section \ref{result:dynamic-search-space} analyzes dynamic search space creation, while Section \ref{result:time-saving} examines the impact of search space reduction on computational efficiency. Additionally, in Section \ref{result:pipeline-recommendation}, we evaluate the most frequently recommended preprocessor-classifier combinations. 

\subsubsection{Dynamic Search Space Creation}
\label{result:dynamic-search-space}

We investigate the research question \textbf{Q3.1} in this subsection. The meta-model from Section \ref{result:meta_model_design} was used to dynamically generate the search space for unseen datasets. First, we estimated the predictive performance of all possible preprocessor-classifier combinations of a given dataset. Then, we ranked them based on their predicted F1 score, from highest to lowest. A search space for a specific dataset was designed by selecting the top combinations, where a threshold \(\theta\) was applied to include only those with an F1 quantile \(\geq \theta\). After designing the dynamic search space, Random Search was used as the optimization method.

To systematically evaluate search space reduction, RS was used as a fixed reference. RS-mtl-$\theta$ was restricted to generating only pipelines already explored by RS, ensuring that identical hyperparameter settings were used. This setup allows us to assess whether the search space was effectively reduced. RS serves as a strong baseline, representing the upper bound of performance.

\begin{figure}[!htb]
    \centering
     \makebox[\textwidth]{\includegraphics[scale=0.40]{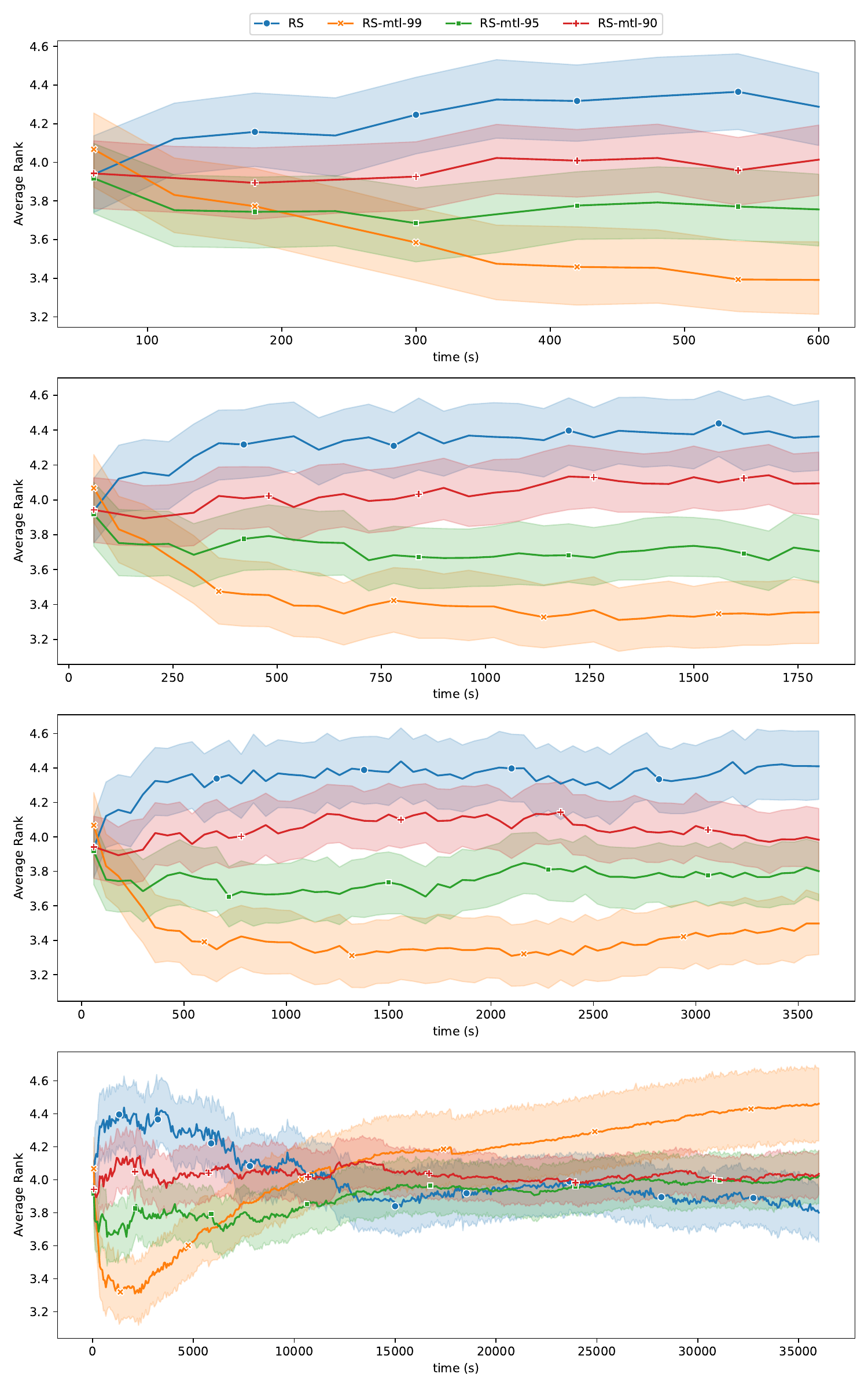}}
    \caption{Comparison of average ranking over time of RS, RS-mtl-99, RS-mtl-95, and RS-mtl-90.}
    \label{fig:ranking_RS_RS-mtl-x}
\end{figure}

Figure \ref{fig:ranking_RS_RS-mtl-x} presents the average rank of F1-weighted scores for RS and RS-mtl-$\theta$  (with \(\theta = 0.99, 0.95, 0.90\)) over the \(D_{\text{test}}\) set with 10 repetitions. Results are shown for time slices of 600s (10 min), 1800s (30 min), 3600s (1h), and 36000s (10h). Initially, RS-mtl-$\theta$  achieves higher rankings, acting as a warm start for optimization, particularly RS-mtl-99. This suggests that high-performing regions are prioritized, whereas RS explores configurations randomly.

Over time, RS is expected to converge towards these high-performing regions. If $\text{RS-mtl-}\theta$ does not excessively reduce the search space, its performance should converge with RS. The final plot in Figure \ref{fig:ranking_RS_RS-mtl-x} confirms this behavior, except for RS-mtl-99, which appears to have overly restricted the search space, limiting its long-term performance.

\begin{figure}[!htb]
    \centering
    \begin{minipage}{0.90\textwidth}
        \centering
        \includegraphics[width=\textwidth]{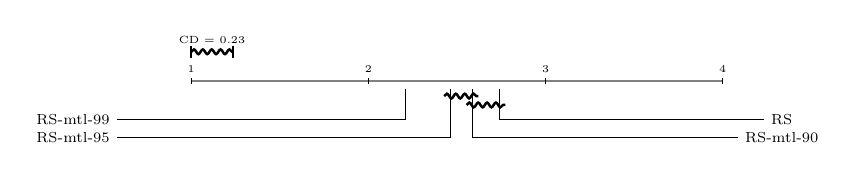}
        \subcaption{600s}
    \end{minipage}
    \begin{minipage}{0.90\textwidth}
        \centering
        \includegraphics[width=\textwidth]{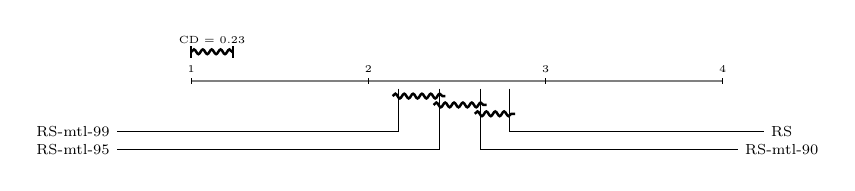}
        \subcaption{1800s}
    \end{minipage}
    \hspace{0.1em}
    \centering
    \begin{minipage}{0.90\textwidth}
        \centering
        \includegraphics[width=\textwidth]{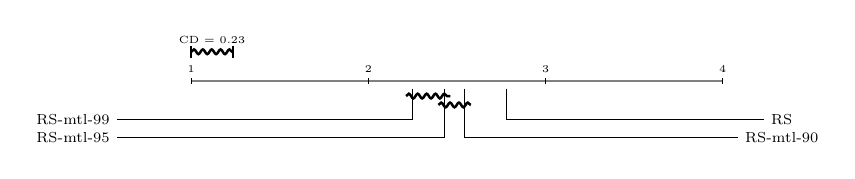}
        \subcaption{3600s}
    \end{minipage}
    \begin{minipage}{0.90\textwidth}
        \centering
        \includegraphics[width=\textwidth]{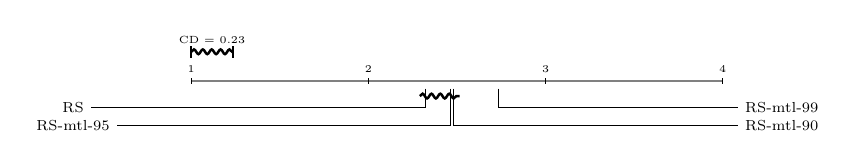}
        \subcaption{36000s}
    \end{minipage}
    \caption{Friedman statistical test with Nemenyi post-hoc analysis ($\alpha = 0.05$) comparing RS and RS-mtl variants (\(\theta = 0.99, 0.95, 0.90\)) across different time slices (600s, 1800s, 3600s, and 36000s). }
    \label{statistical-tests-v2}
\end{figure}

We conducted the Friedman statistical test followed by the Nemenyi post-hoc analysis for each time slice. At each specified time point, the best pipeline was selected based on 10-fold cross-validation of the F1 score and then evaluated on the test set. Figure \ref{statistical-tests-v2} presents the statistical test results. At 600s, RS-mtl-99 showed superior performance. At 1800s and 3600s, RS-mtl-99 and RS-mtl-95 performed similarly, maintaining the superior performance. At 36000s (10h), RS-mtl-90 and RS-mtl-95 tied with RS, while RS-mtl-99 was statistically inferior.

These results provide evidence that: (i) metalearning can dynamically create search spaces while maintaining competitive pipeline compositions, and (ii) metalearning can effectively warm-start optimization, even when reducing the search space.

\begin{figure}[!htb]
    \centering
     \makebox[\textwidth]{\includegraphics[scale=0.40]{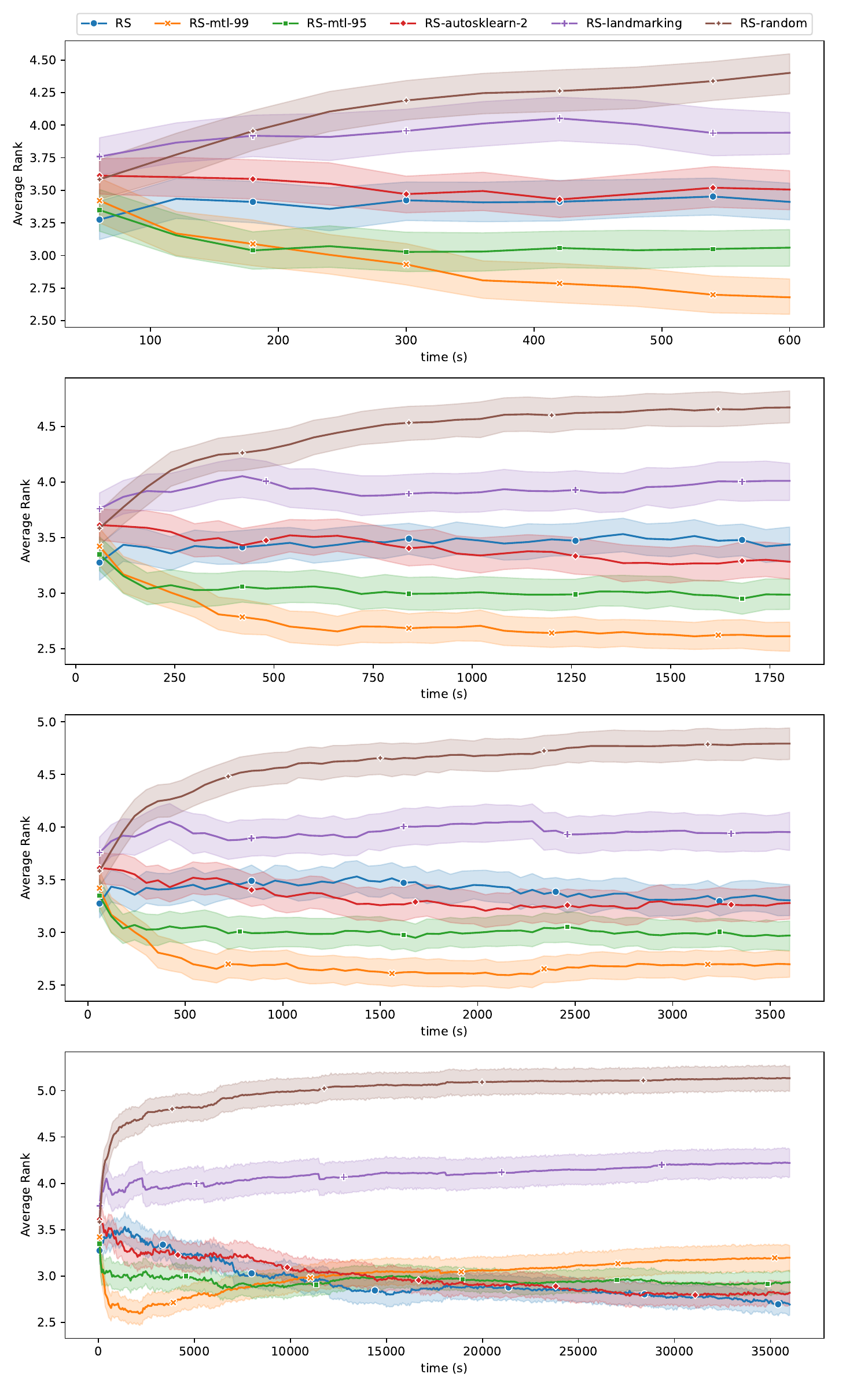}}
    \caption{Comparison of average ranking over time of RS, RS-mtl-99, RS-mtl-95, RS-landmarking, RS-autosklearn-2 and RS-random.}
    \label{fig:rank-benchmarking}
\end{figure}

Similarly to Figure \ref{fig:ranking_RS_RS-mtl-x}, Figure \ref{fig:rank-benchmarking} presents the average rank over time, now including additional baselines from the literature. As a baseline, we introduce \textit{RS-random}, which designs the search space by selecting a single preprocessor and classifier at random. \textit{RS-landmarking} is an adaptation from \citeauthor{kedziora-2024}, using landmarking meta-features and a distance-based algorithm to rank preprocessor-classifier combinations based on past datasets. The best combination from the most similar dataset in \( D_{\text{train}} \) is then used to design the AutoML search space. Additionally, we adapted the metalearning stage of Auto-Sklearn 2.0 \cite{feurer-2022}, referred to as \textit{RS-autosklearn-2}, which designs the search space by selecting pipeline combinations that performed best for each dataset in \( D_{\text{train}} \). All baselines use RS as the optimizer with the same setup as \textit{RS-mtl-99} and \textit{RS-mtl-95}.

Initially, \textit{RS-mtl-99} and \textit{RS-mtl-95} achieved higher rankings, suggesting that high-performing regions are prioritized, whereas \textit{RS-random} explores configurations randomly. All AutoML variations outperformed \textit{RS-random}. \textit{RS-autosklearn-2} and \textit{RS} performed similarly. After 10 hours, \textit{RS}, \textit{RS-mtl-95}, and \textit{RS-autosklearn-2} converged, while \textit{RS-random} and \textit{RS-landmarking} had the lowest performance.

\begin{figure}[!htb]
    \centering
    \begin{minipage}{0.90\textwidth}
        \centering
        \includegraphics[width=\textwidth]{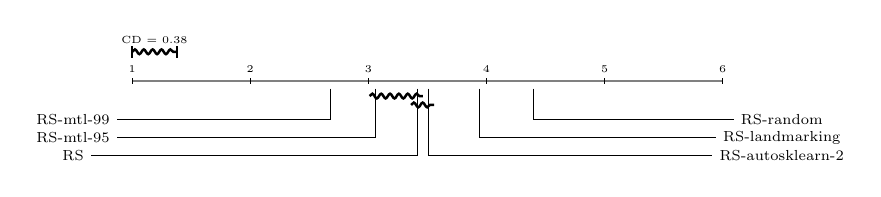}
        \subcaption{600}
    \end{minipage}
    \begin{minipage}{0.90\textwidth}
        \centering
        \includegraphics[width=\textwidth]{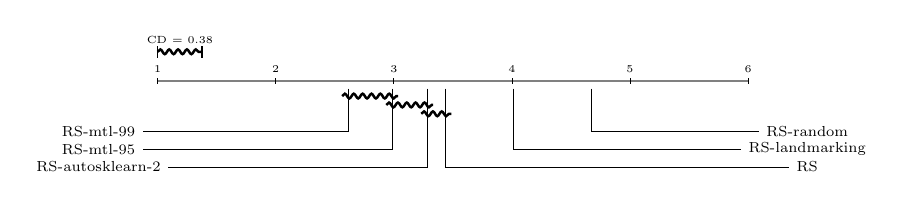}
        \subcaption{1800}
    \end{minipage}
    \hspace{0.1em}
    \centering
    \begin{minipage}{0.90\textwidth}
        \centering
        \includegraphics[width=\textwidth]{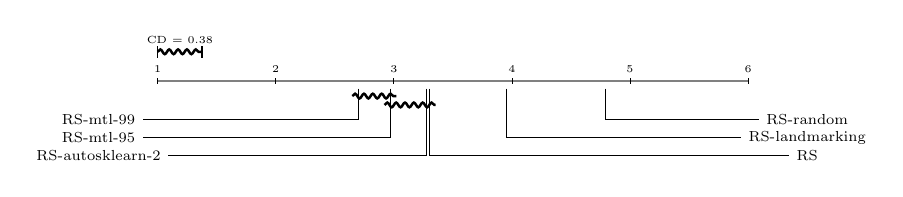}
        \subcaption{3600}
    \end{minipage}
    \begin{minipage}{0.90\textwidth}
        \centering
        \includegraphics[width=\textwidth]{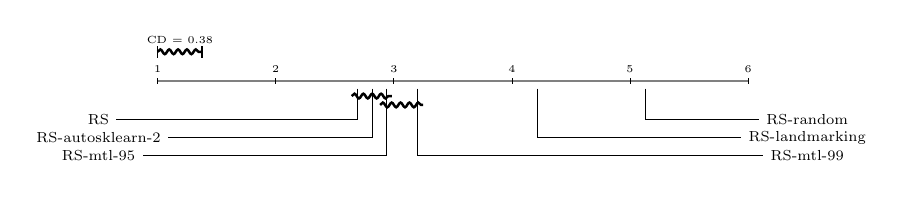}
        \subcaption{36000}
    \end{minipage}
    \caption{Friedman statistical test with Nemenyi post-hoc analysis ($\alpha = 0.05$) comparing RS, RS-mtl-99, RS-mtl-95, RS-landmarking, RS-autosklearn-2 and RS-random, across different time slices (600s, 1800s, 3600s, and 36000s). }
    \label{fig:statistical-tests-v4}
\end{figure}

We also conducted the Friedman statistical test followed by the Nemenyi post-hoc analysis for each time slice. Figure \ref{fig:statistical-tests-v4} presents the results. At 600s, \textit{RS-mtl-99} had the highest rank. At 1800s and 3600s, \textit{RS-mtl-99} and \textit{RS-mtl-95} maintained superior performance. At 36000s (10h), \textit{RS-mtl-90} and \textit{RS-autosklearn-2} tied with \textit{RS}, while \textit{RS-mtl-99}, \textit{RS-landmarking}, and \textit{RS-random} were statistically inferior. 

These results provide evidence that the proposed approach for dynamic search space creation with metalearning is competitive with existing methods in the literature. A table in Appendix \ref{app:performance-table} presents the F1-weighted scores for each dataset across all approaches.

\subsubsection{Time Saving}
\label{result:time-saving}

\begin{figure}[!htb]
    \centering
     \makebox[\textwidth]{\includegraphics[scale=0.50]{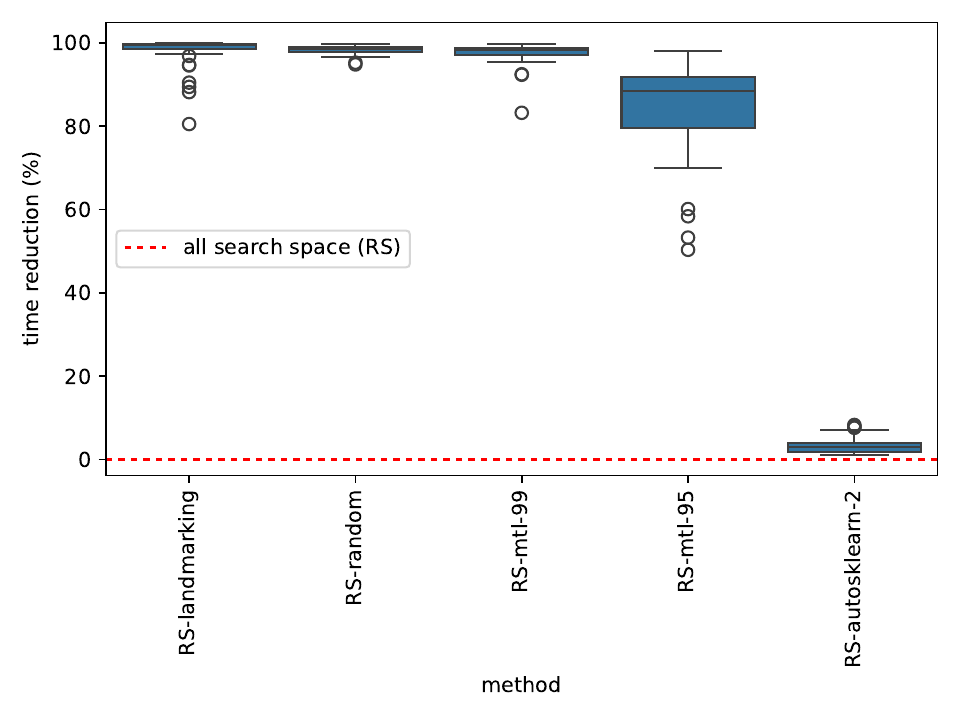}}
    \caption{Percentage of time reduction relative to the RS baseline. The reduction is calculated by dividing the total runtime of each AutoML approach per dataset by the runtime of RS for the same dataset. In red is the search for all space, which means a 0\% reduction.}
    \label{fig:boxplot_time_reduction}
\end{figure}

In this subsection, we explore the research question \textbf{Q2.2}. As the RS search space was fixed, each method could only select pipelines from this predefined space, allowing for a precise evaluation of search space utilization and the corresponding time spent. Figure \ref{fig:boxplot_time_reduction} presents the percentage of time reduction relative to RS, calculated by dividing the total runtime of each AutoML approach per dataset by the runtime of RS for the same dataset.

RS-landmarking, RS-random, and RS-mtl-99 achieved the highest reduction, reducing runtime by approximately 98\%, but they underperformed as shown in Section \ref{result:dynamic-search-space}. RS-autosklearn-2, despite being competitive in performance, did not significantly reduce the search space. RS-mtl-95, however, reduced runtime by approximately 89\% while maintaining competitive performance. In Appendix \ref{app:boxplot_time_reduction-theta}, we show the reduction plot for each variation of $\text{RS-mtl-}\theta$.

\begin{figure}[!htb]
    \centering
    \begin{minipage}{0.45\textwidth}
        \centering
        \subcaption{}
        \includegraphics[width=\textwidth]{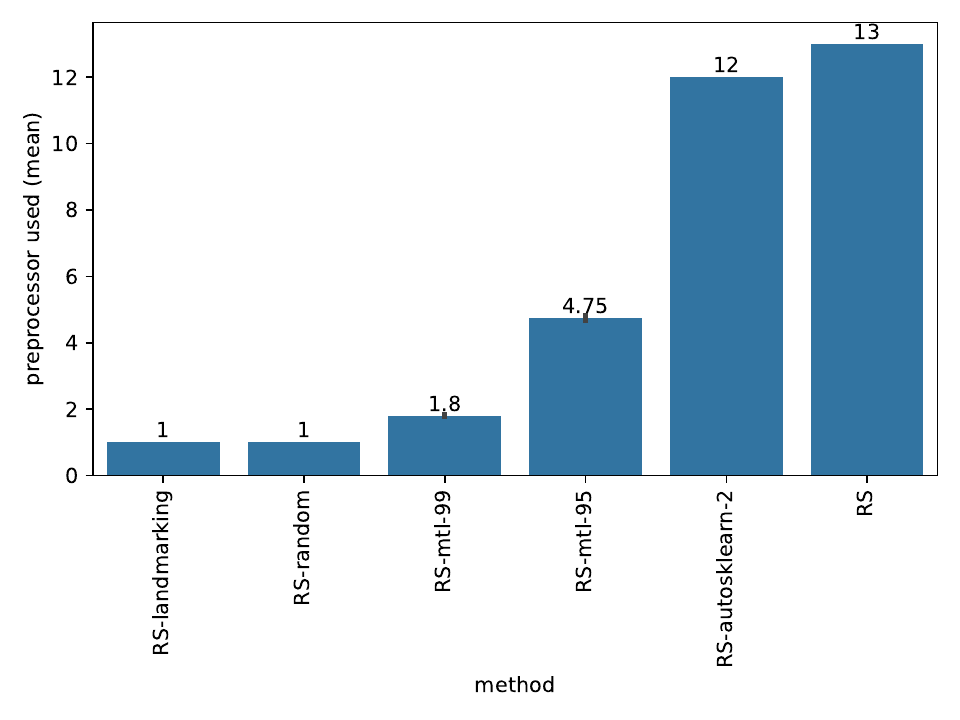}
    \end{minipage}
    \begin{minipage}{0.45\textwidth}
        \centering
        \subcaption{}
        \includegraphics[width=\textwidth]{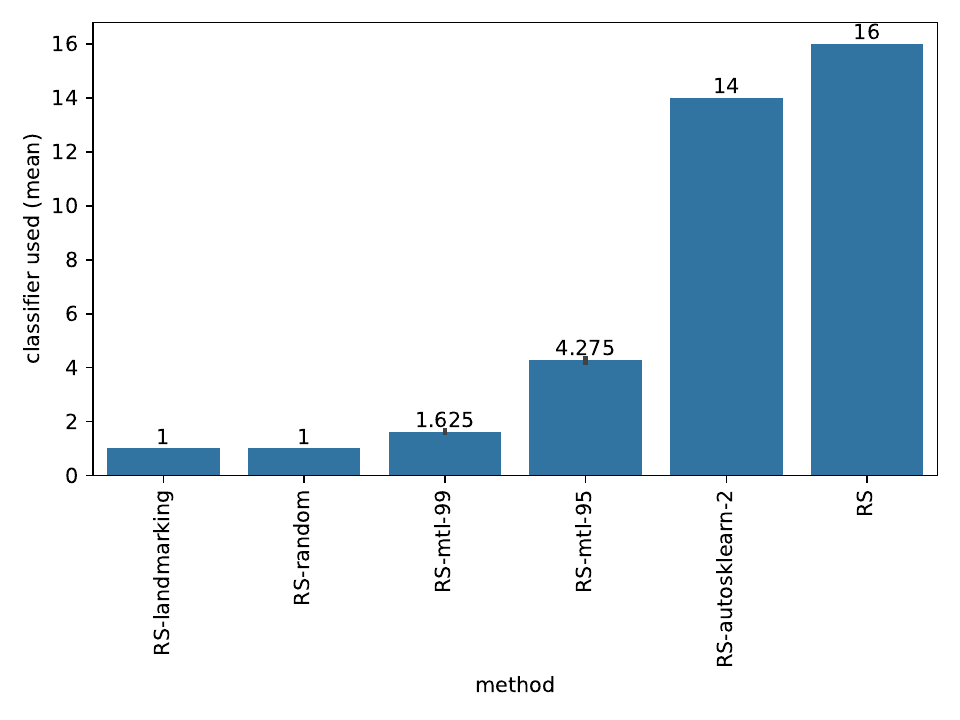}
    \end{minipage}
     \caption{Average number of preprocessors (a) and classifiers (b) used per dataset for each method. The figure illustrates the restrictiveness of the search space across different approaches.}
    \label{fig:barplot_classifier_preprocessor}
\end{figure}

Figure \ref{fig:barplot_classifier_preprocessor} presents the average number of preprocessors (a) and classifiers (b) used by each method per dataset, illustrating the restrictiveness of the search space. RS includes all 13 preprocessors and 16 classifiers in its search space. RS-random and RS-landmarking each select only one preprocessor and one classifier per dataset, resulting in a median of 1. RS-autosklearn-2 designs the search space by selecting the best pipeline for each dataset in $D_{\text{train}}$, leading to the inclusion of almost all preprocessors and classifiers, thus failing to effectively reduce the search space. RS-mtl-99 and RS-mtl-95 generate search spaces with an average of 1.6/16 to 4.3/16 classifiers and 1.8/13 to 4.8/13 preprocessors. It aligns with the observations from Figure \ref{fig:boxplot_time_reduction}.

\subsubsection{Pipelines Recommendations}
\label{result:pipeline-recommendation}

In this subsection, we analyze the recommendations of classifiers and preprocessors within the dynamically designed search space of RS-mtl-95, answering the research question \textbf{Q3.3}. The figures presented show the percentage of selected classifiers, preprocessors, and preprocessor-classifier combinations. These percentages were extracted by aggregating the search space components across all datasets over 10 runs.

\begin{figure}[!htb]
    \centering
    \begin{minipage}{0.52\textwidth}
        \centering
        \subcaption{}
        \includegraphics[width=\textwidth]{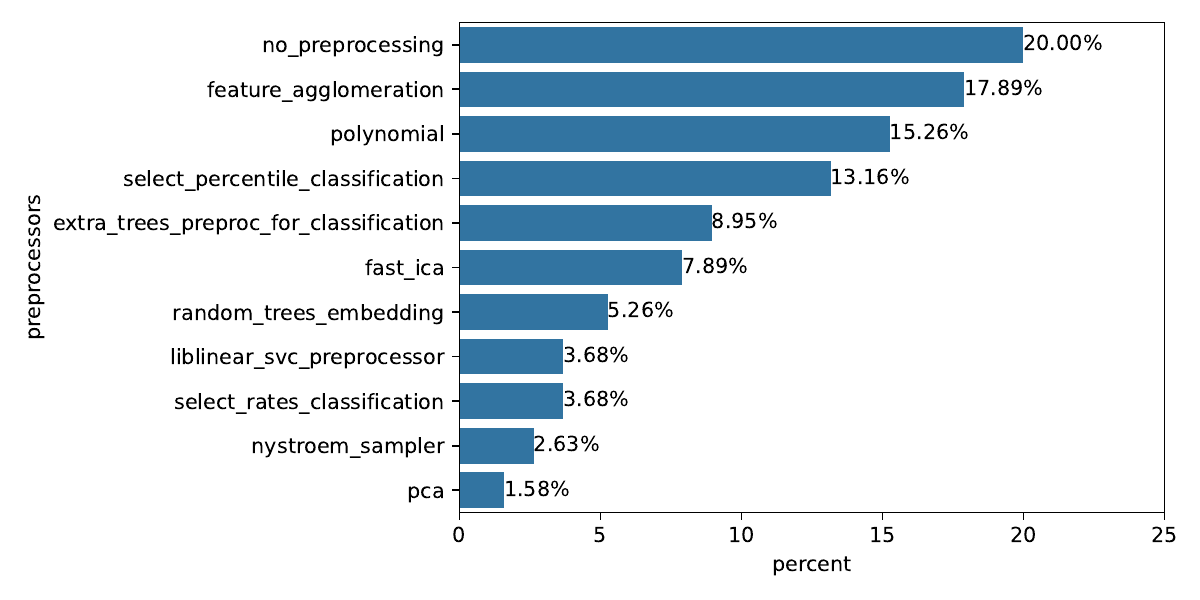}
    \end{minipage}
    \begin{minipage}{0.47\textwidth}
        \centering
        \subcaption{}
        \includegraphics[width=\textwidth]{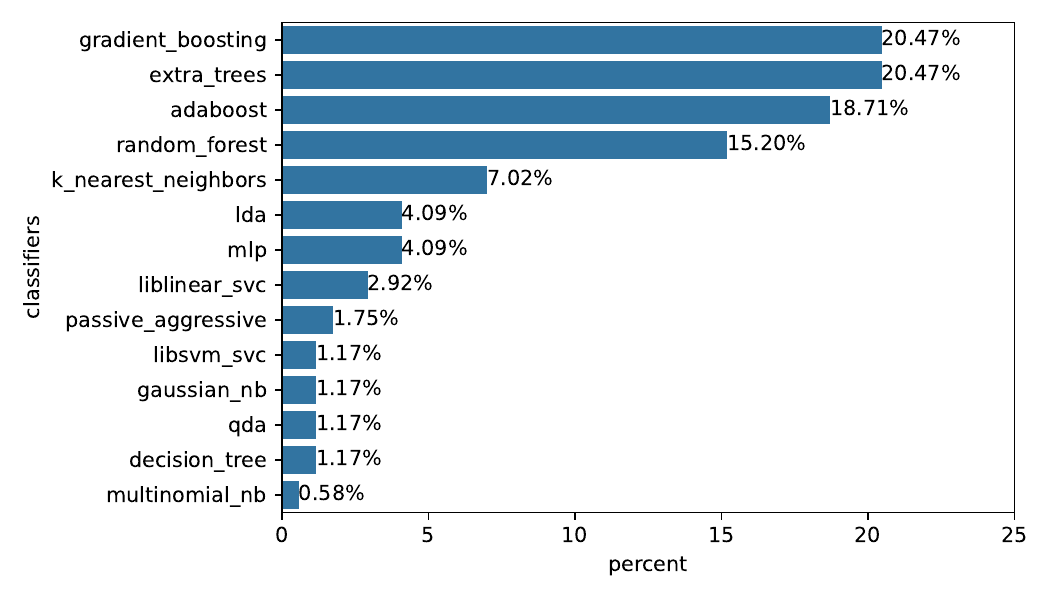}
    \end{minipage}
     \caption{Percentage of times each preprocessor (a) and classifier (b) was included in the dynamically designed search space of RS-mtl-95.}

    \label{fig:perc_classifiers_preprocessors}
\end{figure}

Figure \ref{fig:perc_classifiers_preprocessors} displays the percentage of times each preprocessor (a) and classifier (b) was included in the search space. The most frequently selected preprocessor was no feature preprocesing (20\%), followed by Feature Agglomeration (17.89\%) and Polynomial Features algorithm (15.26\%). The least selected were Principal Component Analysis (1.58\%) and Nystroem kernel transformation (2.63\%). Among classifiers, Gradient Boosting and Extra Trees were the most recommended (20.47\% each), followed by AdaBoost (18.71\%). The least recommended classifiers were Multinomial Naïve Bayes (0.58\%), followed by Decision Tree, Quadratic Discriminant Analysis, Gaussian Naïve Bayes, and Support Vector Machine (1.17\% each). 

\begin{figure}[!htb]
    \centering
     \makebox[\textwidth]{\includegraphics[scale=0.40]{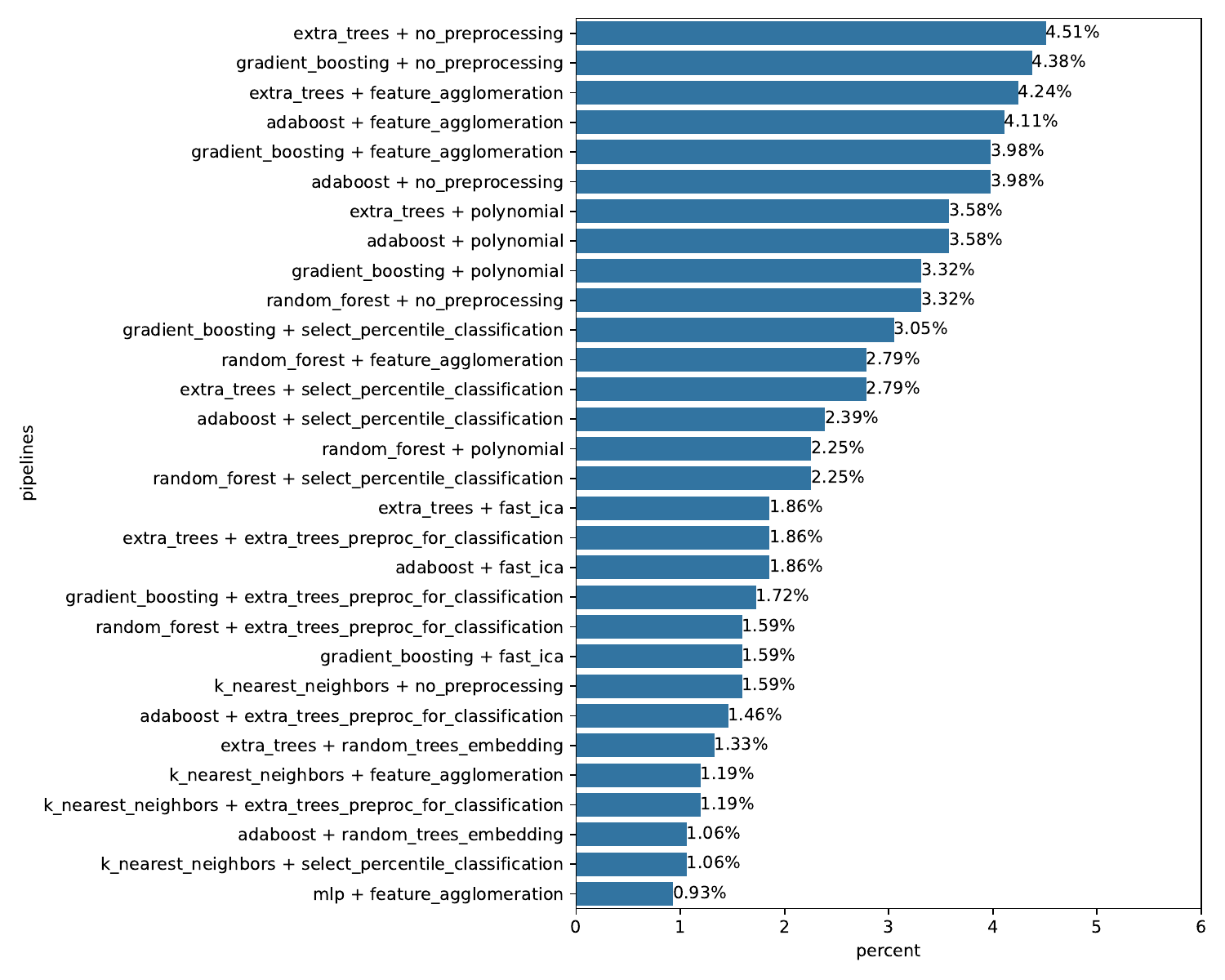}}
    \caption{Percentage of times each preprocessor-classifier combination was included in the dynamically designed search space of RS-mtl-95.}
    \label{fig:perc_pipelines}
\end{figure}

Unlike Figure~\ref{fig:perc_classifiers_preprocessors}, which shows preprocessors and classifiers in isolation, Figure~\ref{fig:perc_pipelines} illustrates the percentage of times each preprocessor-classifier combination was recommended. Only the top 30 combinations, representing approximately 70\% of the total selections, are shown. A full list of all combinations is provided in Appendix~\ref{app:perc_pipelines_all}.

The most frequently used combinations included Extra Trees, Gradient Boosting, and AdaBoost with no preprocessing, Feature Agglomeration, and Polynomial Features. Other notable combinations involved non-ensemble classifiers, such as k-Nearest Neighbors and Multilayer Perceptron. Therefore, these results suggest a preference for ensemble-based classifiers in the dynamically designed search space. Moreover, no preprocessing was the most common choice, as well as Feature Agglomeration, a feature selection algorithm, and Polynomial Features, a feature generation algorithm.

\subsection{Dynamic Search Space on Auto-Sklearn}

In this section, we present an adaptation of Auto-Sklearn that incorporates a dynamic search space strategy for selecting feature preprocessing and modeling algorithms, exploring the research question \textbf{Q.4}. The baseline configuration follows the standard Auto-Sklearn setup, which includes metalearning (with 25 initial configurations) and ensemble construction (combining up to 50 models), using a hold-out validation strategy.

Each experiment was repeated 10 times with different random seeds. For every dataset, we imposed a time budget of 1 hour (3600 seconds), with a maximum per-run limit of 10 minutes (600 seconds), and 10G of RAM, consistent with the configuration used in Section \ref{result:pipeline-design}. Regarding the dynamic search space, we evaluated two thresholds, $\theta = 0.95$ and $\theta = 0.90$, and report results for $\theta = 0.90$, which yielded better performance overall.

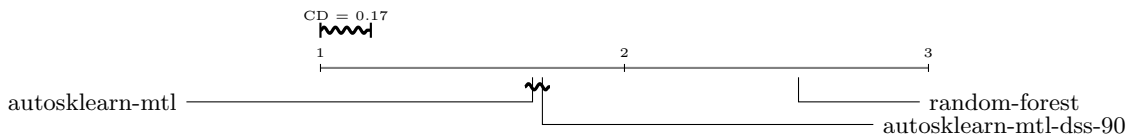
\begin{figure}[!htb]
\centering
\begin{tikzpicture}[xscale=2]
\node (Label) at (2.1657458295101266, 0.7){\tiny{CD = 0.17}}; 
\draw[decorate,decoration={snake,amplitude=.4mm,segment length=1.5mm,post length=0mm},very thick, color = black] (2.0,0.5) -- (2.3314916590202532,0.5);
\foreach \x in {2.0, 2.3314916590202532} \draw[thick,color = black] (\x, 0.4) -- (\x, 0.6);
 
\draw[gray, thick](2.0,0) -- (6.0,0);
\foreach \x in {2.0,4.0,6.0} \draw (\x cm,1.5pt) -- (\x cm, -1.5pt);
\node (Label) at (2.0,0.2){\tiny{1}};
\node (Label) at (4.0,0.2){\tiny{2}};
\node (Label) at (6.0,0.2){\tiny{3}};
\draw[decorate,decoration={snake,amplitude=.4mm,segment length=1.5mm,post length=0mm},very thick, color = black](3.345,-0.25) -- (3.5099999999999993,-0.25);
\node (Point) at (3.395, 0){};\node (Label) at (0.5,-0.45){\scriptsize{autosklearn-mtl}}; \draw (Point) |- (Label);
\node (Point) at (5.145, 0){};\node (Label) at (6.5,-0.45){\scriptsize{random-forest}}; \draw (Point) |- (Label);
\node (Point) at (3.4599999999999995, 0){};\node (Label) at (6.5,-0.75){\scriptsize{autosklearn-mtl-dss-90}}; \draw (Point) |- (Label);
\end{tikzpicture}
\caption{Friedman test with Nemenyi post-hoc analysis ($\alpha = 0.05$) comparing Auto-Sklearn with metalearning and ensemble step (\texttt{autosklearn-mtl}), Auto-Sklearn adapted to dynamic search space (\texttt{autosklearn-mtl-dss-90}), and a Random Forest baseline (\texttt{random-forest}) at 3600 seconds.}
\label{fig:autosklearn-dss-at-theta-90-with}
\end{figure}

Figure~\ref{fig:autosklearn-dss-at-theta-90-with} presents the results of a Friedman statistical test followed by a Nemenyi post-hoc analysis, comparing standard Auto-Sklearn (\texttt{autosklearn-mtl}), its dynamic search space enhanced version (\texttt{autosklearn-mtl-dss-90}), and a Random Forest baseline (\texttt{random-forest})—the latter being a restricted version of Auto-Sklearn using only the Random Forest algorithm and no preprocessing. The analysis indicates no statistically significant difference between \texttt{autosklearn-mtl} and \texttt{autosklearn-mtl-dss-90}, but both significantly outperform the \texttt{random-forest} baseline. Notably, \texttt{autosklearn-mtl-dss-90} achieves comparable performance while operating with a reduced search space, using on average only $7.08$ out of 13 feature preprocessors and $6.50$ out of 15 classifiers.

\begin{figure}[!htb]
\centering
\begin{tikzpicture}[xscale=2]
\node (Label) at (2.1657458295101266, 0.7){\tiny{CD = 0.17}}; 
\draw[decorate,decoration={snake,amplitude=.4mm,segment length=1.5mm,post length=0mm},very thick, color = black] (2.0,0.5) -- (2.3314916590202532,0.5);
\foreach \x in {2.0, 2.3314916590202532} \draw[thick,color = black] (\x, 0.4) -- (\x, 0.6);
 
\draw[gray, thick](2.0,0) -- (6.0,0);
\foreach \x in {2.0,4.0,6.0} \draw (\x cm,1.5pt) -- (\x cm, -1.5pt);
\node (Label) at (2.0,0.2){\tiny{1}};
\node (Label) at (4.0,0.2){\tiny{2}};
\node (Label) at (6.0,0.2){\tiny{3}};
\draw[decorate,decoration={snake,amplitude=.4mm,segment length=1.5mm,post length=0mm},very thick, color = black](3.4225000000000003,-0.25) -- (3.585,-0.25);
\node (Point) at (3.4725, 0){};\node (Label) at (0.5,-0.45){\scriptsize{autosklearn\_dss\_90}}; \draw (Point) |- (Label);
\node (Point) at (4.9925, 0){};\node (Label) at (6.5,-0.45){\scriptsize{random\_forest}}; \draw (Point) |- (Label);
\node (Point) at (3.535, 0){};\node (Label) at (6.5,-0.75){\scriptsize{autosklearn}}; \draw (Point) |- (Label);
\end{tikzpicture}
\caption{Friedman test with Nemenyi post-hoc analysis ($\alpha = 0.05$) comparing Auto-Sklearn without metalearning step (\texttt{autosklearn}), Auto-Sklearn without metalearning adapted to dynamic search space (\texttt{autosklearn-dss-90}), and a Random Forest baseline (\texttt{random-forest}) at 3600 seconds.}
\label{fig:autosklearn-dss-at-theta-90-without}
\end{figure}
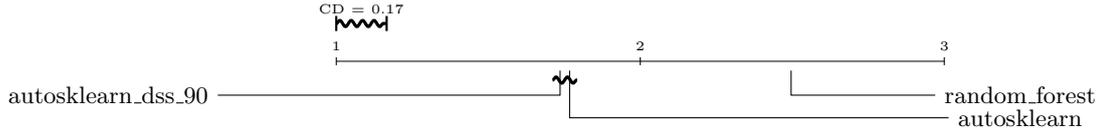

Figure~\ref{fig:autosklearn-dss-at-theta-90-without} presents the same experiment with the metalearning warm start deactivated. The results similarly show that Auto-Sklearn with dynamic search space and without metalearning (\textit{autosklearn-dss-90}) performs comparably to the version without metalearning (\textit{autosklearn}), with no significant difference observed between them. Additional statistical tests for $\theta = 0.95$ are provided in Appendix~\ref{app:autosklearn-dss-at-theta-95} for reference.

Table~\ref{tab:autosklearn-dss-performance-stat} reports the mean and standard deviation of the F1-weighted score across datasets. We applied the Wilcoxon signed-rank test ($\alpha = 0.05$) for pairwise comparisons between \texttt{autosklearn-mtl-dss-90} and each of the two baselines. The \texttt{autosklearn-mtl-dss-90} variant achieved statistically significant wins in 3 out of 40 cases against \texttt{autosklearn-mtl}, and in 21 out of 40 cases against the \texttt{random-forest} baseline. In contrast, \texttt{autosklearn-mtl} recorded 5 wins over \texttt{autosklearn-mtl-dss-90}, while the \texttt{random-forest} baseline had none.

These results suggest that the dynamic search space approach offers a significant improvement over the \texttt{random-forest} baseline, while maintaining performance comparable to the \texttt{autosklearn-mtl} using the whole search space.

\begin{longtable}[c]{rlll}
\caption{The mean and standard deviation of the F1-weighted score for Auto-Sklearn with metalearning and ensemble steps (\textit{autosklearn-mtl}), Auto-Sklearn with a dynamic search space (\textit{autosklearn-mtl-dss-90}), and a tuned Random Forest (\textit{random-forest}) at 3600 seconds. Pairwise comparisons between \textit{autosklearn-mtl-dss-90} and each baseline are performed using the Wilcoxon signed-rank test. Statistically significant differences ($\alpha = 0.05$) are indicated with a square when comparing \textit{autosklearn-mtl-dss-90} versus \textit{autosklearn-mtl}, and with a circle when comparing \textit{autosklearn-mtl-dss-90} versus \textit{random-forest}.}
\label{tab:autosklearn-dss-performance-stat}\\
\hline
\multicolumn{1}{l}{\textbf{Datasets ID}} & \textbf{autosklearn-mtl-dss-90} & \textbf{autosklearn-mtl} & \textbf{\textit{random}-forest} \\ \hline
\endfirsthead
\multicolumn{4}{c}%
{{\bfseries Table \thetable\ continued from previous page}} \\
\hline
\multicolumn{1}{l}{\textbf{Datasets ID}} & \textbf{autosklearn-mtl-dss-90} & \textbf{autosklearn-mtl} & \textbf{random-forest} \\ \hline
\endhead
\hline
\endfoot
\endlastfoot
11 & $0.9792 \pm 0.02$ \markercircle & $0.9980 \pm 0.01$ \markersquare & $0.9069 \pm 0.02$ \\
23 & $0.5471 \pm 0.02$ & $0.5547 \pm 0.02$ & $0.5425 \pm 0.02$ \\
28 & $0.9896 \pm 0.00$ \markersquare~\markercircle & $0.9883 \pm 0.00$ & $0.9807 \pm 0.00$ \\
30 & $0.9724 \pm 0.00$ & $0.9720 \pm 0.01$ & $0.9720 \pm 0.01$ \\
155 & $0.7661 \pm 0.24$ & $0.9989 \pm 0.00$ \markersquare & $0.9926 \pm 0.00$ \\
181 & $0.6018 \pm 0.02$ & $0.6003 \pm 0.02$ & $0.5986 \pm 0.02$ \\
307 & $0.9729 \pm 0.02$ \markercircle & $0.9745 \pm 0.01$ & $0.9313 \pm 0.01$ \\
725 & $0.9409 \pm 0.00$ \markercircle & $0.9432 \pm 0.00$ \markersquare & $0.9370 \pm 0.00$ \\
761 & $0.9372 \pm 0.01$ \markercircle & $0.9375 \pm 0.01$ & $0.9332 \pm 0.00$ \\
770 & $1.0000 \pm 0.00$ \markercircle & $1.0000 \pm 0.00$ & $0.9618 \pm 0.02$ \\
816 & $0.8323 \pm 0.01$ & $0.8322 \pm 0.00$ & $0.8305 \pm 0.01$ \\
823 & $0.9836 \pm 0.00$ & $0.9835 \pm 0.00$ & $0.9828 \pm 0.00$ \\
826 & $0.6592 \pm 0.04$ & $0.6597 \pm 0.04$ & $0.6646 \pm 0.04$ \\
841 & $0.9605 \pm 0.01$ \markercircle & $0.9588 \pm 0.01$ & $0.9470 \pm 0.02$ \\
846 & $0.9051 \pm 0.00$ \markercircle & $0.9048 \pm 0.00$ & $0.8646 \pm 0.00$ \\
901 & $0.9322 \pm 0.00$ \markercircle & $0.9364 \pm 0.00$ \markersquare & $0.9178 \pm 0.00$ \\
934 & $0.9415 \pm 0.01$ & $0.9460 \pm 0.01$ & $0.9422 \pm 0.01$ \\
937 & $0.8953 \pm 0.02$ \markercircle & $0.9027 \pm 0.02$ & $0.8501 \pm 0.03$ \\
940 & $0.9946 \pm 0.01$ & $0.9970 \pm 0.01$ & $0.9954 \pm 0.01$ \\
951 & $1.0000 \pm 0.00$ & $1.0000 \pm 0.00$ & $0.9993 \pm 0.00$ \\
1044 & $0.7513 \pm 0.02$ \markercircle & $0.7492 \pm 0.02$ & $0.6833 \pm 0.01$ \\
1046 & $0.9518 \pm 0.00$ \markercircle & $0.9602 \pm 0.00$ \markersquare & $0.9495 \pm 0.00$ \\
1053 & $0.7888 \pm 0.01$ \markercircle & $0.7874 \pm 0.00$ & $0.7860 \pm 0.01$ \\
1166 & $0.9435 \pm 0.01$ & $0.9430 \pm 0.01$ & $0.9389 \pm 0.01$ \\
1233 & $0.6483 \pm 0.02$ \markercircle & $0.6347 \pm 0.03$ & $0.5208 \pm 0.03$ \\
1494 & $0.8652 \pm 0.03$ & $0.8637 \pm 0.02$ & $0.8509 \pm 0.02$ \\
1501 & $0.9464 \pm 0.01$ & $0.9516 \pm 0.01$ & $0.9405 \pm 0.01$ \\
1515 & $0.9163 \pm 0.03$ \markercircle & $0.9215 \pm 0.01$ & $0.8844 \pm 0.04$ \\
1528 & $0.8977 \pm 0.01$ \markersquare~\markercircle & $0.8943 \pm 0.01$ & $0.8933 \pm 0.01$ \\
1535 & $0.9542 \pm 0.00$ & $0.9552 \pm 0.00$ & $0.9550 \pm 0.00$ \\
1541 & $0.9188 \pm 0.00$ & $0.9182 \pm 0.00$ & $0.9182 \pm 0.00$ \\
1553 & $0.4883 \pm 0.03$ & $0.4800 \pm 0.04$ & $0.4854 \pm 0.03$ \\
4134 & $0.7960 \pm 0.01$ & $0.7966 \pm 0.01$ & $0.7959 \pm 0.01$ \\
4538 & $0.6788 \pm 0.01$ \markercircle & $0.6762 \pm 0.02$ & $0.6535 \pm 0.01$ \\
40499 & $0.9982 \pm 0.00$ \markercircle & $0.9981 \pm 0.00$ & $0.9850 \pm 0.00$ \\
40649 & $0.7190 \pm 0.02$ \markercircle & $0.7236 \pm 0.02$ & $0.6525 \pm 0.02$ \\
40704 & $0.7650 \pm 0.01$ & $0.7625 \pm 0.01$ & $0.7631 \pm 0.01$ \\
40985 & $0.0561 \pm 0.00$ & $0.0561 \pm 0.00$ & $0.0548 \pm 0.00$ \\
41145 & $0.8072 \pm 0.01$ \markercircle & $0.8092 \pm 0.01$ & $0.7631 \pm 0.01$ \\
41990 & $0.7966 \pm 0.00$ \markersquare~\markercircle & $0.7910 \pm 0.00$ & $0.5487 \pm 0.01$ \\ \hline
\end{longtable}

Finally, we investigated the potential impact of the dynamic search space on overfitting reduction for Auto-Sklearn, following a methodology similar to that proposed by \citeauthor{fabris-2019}. We analyzed the differences in F1-weighted scores between training, validation, and test sets—specifically train-test, train-validation, and validation-test gaps—as summarized in Table~\ref{tab:overfiting-train-test-val-autosklearn-dss}. 

Using the Wilcoxon signed-rank test ($\alpha = 0.05$), we found no statistically significant differences between the standard and dynamic search space variants. Thus, there is no empirical evidence to suggest that the dynamic search space approach reduces overfitting in Auto-Sklearn.

\begin{longtable}[!htb]{llll}
\caption{The mean and standard deviation of performance differences train $-$ test, train 
$-$ validation, and validation $-$ test. }
\label{tab:overfiting-train-test-val-autosklearn-dss}\\
\hline
\textbf{method} & \textbf{train - test} & \textbf{train - val} & \textbf{val - test} \\ \hline
\endfirsthead
\multicolumn{4}{c}%
{{\bfseries Table \thetable\ continued from previous page}} \\
\textbf{method} & \textbf{train - test} & \textbf{train - validation} & \textbf{validation - test} \\ \hline
\endhead
autosklearn-mtl-dss-90 & $0.0844 \pm 0.11$ & $0.0595 \pm 0.09$ & $0.0249 \pm 0.04$ \\
autosklearn-mtl & $0.0790 \pm 0.10$ & $0.0549 \pm 0.09$ & $0.0241 \pm 0.04$\\
\hline
\end{longtable}

\section{Final Remarks}
\label{sec:final-remarks}
This section presents the key insights from our study and discusses its limitations. We first summarize the main findings, followed by an analysis of the approach’s constraints and potential areas for improvement.

\subsection{Key Insights}
The proposed meta-model successfully predicts the performance of preprocessor-classifier combinations by learning three key aspects: (i) the complexity of tree-based algorithms through \textit{nodes\_per\_inst}, (ii) the historical performance of pipeline combinations, and (iii) entropy-based information from the task, where higher entropy indicates higher 
unpredictability. When analyzing the computational complexity of meta-feature extraction, we identified Landmarking and Statistical meta-features as the most expensive, with their computation time correlated to the number of examples (0.40) and the number of features (0.95), respectively.

The dynamic search space generation approach effectively prioritized high-performing configurations, resulting in a reduction in computational cost without compromising performance. In particular, RS-mtl-95 achieved an approximate 89\% reduction in runtime while maintaining competitive results. Compared to baselines and competing methods, it demonstrated superior performance within the first hour and converged with RS after ten hours. These findings align with recent research indicating that restricting the search space can yield substantial computational efficiency while preserving competitive performance~\cite{yakovlev-2020, xue-2022, borboudakis-2023, kedziora-2024}.

Furthermore, the meta-model provided a warm-start advantage, particularly in the early stages of optimization, by directing the search toward promising regions of the search space rather than relying on purely random exploration. The ability of metalearning to enhance warm-start optimization has been well documented in the literature~\cite{gomes-2012, feurer-2015a, feurer-2015b, mantovani-2019}. Notably, this advantage has been leveraged in various contexts, such as suggesting initial points for particle swarm optimization and Tabu Search for tuning SVM, as well as in sequential model-based Bayesian optimization approaches~\cite{feurer-2015a, gomes-2012}. 

Our analysis also revealed a preference for ensemble-based classifiers, with Gradient Boosting, Extra Trees, and AdaBoost being the most frequently recommended models. Among preprocessing techniques, no preprocessing, Feature Agglomeration, and Polynomial Features were the most commonly selected. In particular, Polynomial Features plays a key role in TPOT~\cite{olson-2016}, where it serves as a critical feature engineering step. Interestingly, while some empirical studies suggest that feature preprocessing does not always enhance predictive performance~\cite{schoenfeld-2018}, it can significantly reduce computational training time, which is particularly beneficial when handling large datasets~\cite{bolon-2015, cai-2018}.

Moreover, ensemble methods consistently outperform individual models by combining the strengths of diverse learners, thereby reducing variance and bias to enhance predictive performance. This phenomenon has been widely documented in the AutoML literature~\cite{thornton-2013, feurer-2015a, olson-2016, wistuba-2017}. Our findings reinforce this notion, demonstrating that ensemble-based strategies remain a cornerstone of automated ML pipelines.

Finally, the dynamic search space strategy in Auto-Sklearn maintains comparable performance to the standard version while reducing the number of preprocessing and modeling components by 7.08/13 and 6.50/13, respectively. Statistical tests suggest no significant difference between the standard and dynamic variants; however, both outperform the Random Forest baseline. Moreover, no evidence was found that dynamic search reduces overfitting of Auto-Sklearn.

\subsection{Limitations}
Despite its advantages, the proposed approach has some limitations. The search space reduction strategy, particularly in RS-mtl-99, sometimes resulted in overly restrictive configurations, limiting long-term optimization performance. The impact of search space size depends on the choice of threshold \(\theta\), which should be carefully selected to balance search efficiency and solution diversity.

The generalization ability of the meta-model depends on the diversity of the meta-knowledge base (\textit{D\_train}). If the meta-training set lacks datasets with certain characteristics, performance may degrade when encountering unseen tasks with those properties.

\section{Conclusion}
\label{sec:conclusion-dp-dp}

This study presented a novel metalearning approach for dynamically designing search spaces in AutoML, enhancing computational efficiency while maintaining competitive performance. The experimental results for RS demonstrated that dynamically generated search spaces can reduce computational costs, while the meta-model offers a warm-start advantage by prioritizing promising regions of the search space. Moreover, the dynamic search space strategy in Auto-Sklearn maintained comparable performance while reducing the search space. However, no evidence was found that dynamic search can reduce overfitting of Auto-Sklearn.

Future research will explore the inclusion of new meta-feature groups, as well as introduce strategies to reduce the computational cost, such as early stopping. Ultimately, we intend to extend this approach to other ML tasks, including regression and clustering, to assess its broader applicability and impact.

\bmhead{Acknowledgements}
Research carried out using the computational resources of the Center for Mathematical Sciences Applied to Industry (CeMEAI) funded by FAPESP (grant 2013/07375-0) and FAPESP (grant 2018/14819-5). This study was financed in part by the Coordenação de Aperfeiçoamento de Pessoal de Nível Superior - Brasil (CAPES) - Finance Code 001.

\bibliography{sn-bibliography}

\begin{appendices}
\newpage

\section{Datasets}
\label{sec:appx-datasets-split-train-test}

Table \ref{tab:sup-dataset-md-train} presents the $D_\text{train}$ used for building the meta-model, while Table \ref{tab:sup-dataset-md-test} presents the $D_\text{test}$ used for assessing testing performance in Auto-Sklearn and Random Search experiments.

\begin{center}
\tiny
\begin{longtable}[c]{lllllll}
\caption{Datasets meta data train}
\label{tab:sup-dataset-md-train}\\
\hline
\multicolumn{1}{c}{\textbf{\begin{tabular}[c]{@{}c@{}}OpenML \\ ID\end{tabular}}} & \multicolumn{1}{c}{\textbf{\# Examples}} & \multicolumn{1}{c}{\textbf{\# Features}} & \multicolumn{1}{c}{\textbf{\begin{tabular}[c]{@{}c@{}}\# Categorical \\ Features\end{tabular}}} & \multicolumn{1}{c}{\textbf{\# Class}} & \multicolumn{1}{c}{\textbf{\begin{tabular}[c]{@{}c@{}}Majority\\ Class \%\end{tabular}}} & \multicolumn{1}{c}{\textbf{\begin{tabular}[c]{@{}c@{}}Minority\\ Class \%\end{tabular}}} \\ \hline
\endfirsthead
\multicolumn{7}{c}%
{{\bfseries Table \thetable\ continued from previous page}} \\
\hline
\multicolumn{1}{c}{\textbf{\begin{tabular}[c]{@{}c@{}}OpenML \\ ID\end{tabular}}} & \multicolumn{1}{c}{\textbf{\# Examples}} & \multicolumn{1}{c}{\textbf{\# Features}} & \multicolumn{1}{c}{\textbf{\begin{tabular}[c]{@{}c@{}}\# Categorical \\ Features\end{tabular}}} & \multicolumn{1}{c}{\textbf{\# Class}} & \multicolumn{1}{c}{\textbf{\begin{tabular}[c]{@{}c@{}}Majority\\ Class \%\end{tabular}}} & \multicolumn{1}{c}{\textbf{\begin{tabular}[c]{@{}c@{}}Minority\\ Class \%\end{tabular}}} \\ \hline
\endhead
\hline
\endfoot
\endlastfoot
2 & 898 & 38 & 32 & 5 & 76.17 & 0.0* \\
6 & 20000 & 16 & 0 & 26 & 4.07 & 3.67 \\
15 & 699 & 9 & 0 & 2 & 65.52 & 34.48 \\
24 & 8124 & 22 & 22 & 2 & 51.8 & 48.2 \\
26 & 12960 & 8 & 8 & 5 & 33.33 & 0.02 \\
32 & 10992 & 16 & 0 & 10 & 10.41 & 9.6 \\
37 & 768 & 8 & 0 & 2 & 65.1 & 34.9 \\
42 & 683 & 35 & 35 & 19 & 13.47 & 1.17 \\
44 & 4601 & 57 & 0 & 2 & 60.6 & 39.4 \\
46 & 3190 & 60 & 60 & 3 & 51.88 & 24.04 \\
50 & 958 & 9 & 9 & 2 & 65.34 & 34.66 \\
57 & 3772 & 29 & 22 & 4 & 92.29 & 0.05 \\
60 & 5000 & 40 & 0 & 3 & 33.84 & 33.06 \\
151 & 45312 & 8 & 1 & 2 & 57.55 & 42.45 \\
182 & 6430 & 36 & 0 & 6 & 23.81 & 9.72 \\
184 & 28056 & 6 & 6 & 18 & 16.23 & 0.1 \\
185 & 1340 & 16 & 1 & 3 & 90.67 & 4.25 \\
188 & 736 & 19 & 5 & 5 & 29.08 & 14.27 \\
279 & 45164 & 74 & 0 & 11 & 50.97 & 0.0* \\
300 & 7797 & 617 & 0 & 26 & 3.85 & 3.82 \\
310 & 11183 & 6 & 0 & 2 & 97.68 & 2.32 \\
311 & 937 & 49 & 0 & 2 & 95.62 & 4.38 \\
333 & 556 & 6 & 6 & 2 & 50.0 & 50.0 \\
334 & 601 & 6 & 6 & 2 & 65.72 & 34.28 \\
335 & 554 & 6 & 6 & 2 & 51.99 & 48.01 \\
375 & 9961 & 14 & 0 & 9 & 16.2 & 7.85 \\
377 & 600 & 60 & 0 & 6 & 16.67 & 16.67 \\
451 & 500 & 5 & 3 & 2 & 55.6 & 44.4 \\
458 & 841 & 70 & 0 & 4 & 37.69 & 6.54 \\
469 & 797 & 4 & 4 & 6 & 19.45 & 15.43 \\
470 & 672 & 9 & 4 & 2 & 66.67 & 33.33 \\
715 & 1000 & 25 & 0 & 2 & 55.7 & 44.3 \\
717 & 508 & 10 & 0 & 2 & 56.3 & 43.7 \\
722 & 15000 & 48 & 0 & 2 & 66.39 & 33.61 \\
727 & 40768 & 10 & 0 & 2 & 50.09 & 49.91 \\
728 & 4052 & 7 & 0 & 2 & 76.04 & 23.96 \\
734 & 13750 & 40 & 0 & 2 & 57.61 & 42.39 \\
735 & 8192 & 12 & 0 & 2 & 69.76 & 30.24 \\
737 & 3107 & 6 & 0 & 2 & 50.4 & 49.6 \\
740 & 1000 & 10 & 0 & 2 & 56.0 & 44.0 \\
742 & 500 & 100 & 0 & 2 & 56.6 & 43.4 \\
750 & 500 & 7 & 0 & 2 & 50.8 & 49.2 \\
752 & 8192 & 32 & 0 & 2 & 50.39 & 49.61 \\
757 & 528 & 21 & 2 & 2 & 89.77 & 10.23 \\
772 & 2178 & 3 & 0 & 2 & 55.51 & 44.49 \\
799 & 1000 & 5 & 0 & 2 & 50.3 & 49.7 \\
802 & 1945 & 18 & 6 & 2 & 50.03 & 49.97 \\
803 & 7129 & 5 & 0 & 2 & 53.06 & 46.94 \\
807 & 8192 & 8 & 0 & 2 & 50.88 & 49.12 \\
819 & 9517 & 6 & 0 & 2 & 50.28 & 49.72 \\
821 & 22784 & 16 & 0 & 2 & 70.4 & 29.6 \\
825 & 506 & 20 & 3 & 2 & 55.93 & 44.07 \\
833 & 8192 & 32 & 0 & 2 & 68.96 & 31.04 \\
837 & 1000 & 50 & 0 & 2 & 54.7 & 45.3 \\
839 & 782 & 8 & 2 & 2 & 64.96 & 35.04 \\
847 & 6574 & 14 & 0 & 2 & 53.26 & 46.74 \\
871 & 3848 & 5 & 0 & 2 & 50.0 & 50.0 \\
881 & 40768 & 10 & 3 & 2 & 59.66 & 40.34 \\
884 & 500 & 5 & 0 & 2 & 50.2 & 49.8 \\
886 & 500 & 7 & 0 & 2 & 50.2 & 49.8 \\
897 & 1161 & 15 & 2 & 2 & 70.03 & 29.97 \\
903 & 1000 & 25 & 0 & 2 & 56.3 & 43.7 \\
920 & 500 & 50 & 0 & 2 & 59.0 & 41.0 \\
923 & 8641 & 4 & 1 & 2 & 55.01 & 44.99 \\
930 & 1302 & 33 & 1 & 2 & 52.84 & 47.16 \\
936 & 500 & 10 & 0 & 2 & 54.4 & 45.6 \\
947 & 559 & 4 & 1 & 2 & 95.71 & 4.29 \\
949 & 559 & 4 & 1 & 2 & 85.69 & 14.31 \\
950 & 559 & 4 & 1 & 2 & 96.6 & 3.4 \\
981 & 10108 & 68 & 68 & 2 & 73.14 & 26.86 \\
1039 & 4229 & 1617 & 0 & 2 & 96.48 & 3.52 \\
1049 & 1458 & 37 & 0 & 2 & 87.79 & 12.21 \\
1050 & 1563 & 37 & 0 & 2 & 89.76 & 10.24 \\
1056 & 9466 & 38 & 0 & 2 & 99.28 & 0.72 \\
1063 & 522 & 21 & 0 & 2 & 79.5 & 20.5 \\
1068 & 1109 & 21 & 0 & 2 & 93.06 & 6.94 \\
1069 & 5589 & 36 & 0 & 2 & 99.59 & 0.41 \\
1116 & 6598 & 167 & 1 & 2 & 84.59 & 15.41 \\
1120 & 19020 & 10 & 0 & 2 & 64.84 & 35.16 \\
1128 & 1545 & 10935 & 0 & 2 & 77.73 & 22.27 \\
1130 & 1545 & 10935 & 0 & 2 & 91.84 & 8.16 \\
1134 & 1545 & 10935 & 0 & 2 & 83.17 & 16.83 \\
1142 & 1545 & 10935 & 0 & 2 & 96.05 & 3.95 \\
1146 & 1545 & 10935 & 0 & 2 & 95.53 & 4.47 \\
1161 & 1545 & 10935 & 0 & 2 & 81.49 & 18.51 \\
1457 & 1500 & 10000 & 0 & 50 & 2.0 & 2.0 \\
1459 & 10218 & 7 & 0 & 10 & 13.86 & 5.87 \\
1462 & 1372 & 4 & 0 & 2 & 55.54 & 44.46 \\
1466 & 2126 & 35 & 0 & 10 & 27.23 & 2.49 \\
1471 & 14980 & 14 & 0 & 2 & 55.12 & 44.88 \\
1475 & 6118 & 51 & 0 & 6 & 41.75 & 7.94 \\
1478 & 10299 & 561 & 0 & 6 & 18.88 & 13.65 \\
1479 & 1212 & 100 & 0 & 2 & 50.0 & 50.0 \\
1480 & 583 & 10 & 1 & 2 & 71.36 & 28.64 \\
1481 & 28056 & 6 & 3 & 18 & 16.23 & 0.1 \\
1483 & 164860 & 7 & 2 & 11 & 33.05 & 0.84 \\
1485 & 2600 & 500 & 0 & 2 & 50.0 & 50.0 \\
1487 & 2534 & 72 & 0 & 2 & 93.69 & 6.31 \\
1491 & 1600 & 64 & 0 & 100 & 1.0 & 1.0 \\
1496 & 7400 & 20 & 0 & 2 & 50.49 & 49.51 \\
1497 & 5456 & 24 & 0 & 4 & 40.41 & 6.01 \\
1502 & 245057 & 3 & 0 & 2 & 79.25 & 20.75 \\
1503 & 263256 & 14 & 0 & 10 & 10.06 & 9.92 \\
1507 & 7400 & 20 & 0 & 2 & 50.04 & 49.96 \\
1509 & 149332 & 4 & 0 & 22 & 14.73 & 0.61 \\
1510 & 569 & 30 & 0 & 2 & 62.74 & 37.26 \\
1529 & 1521 & 3 & 0 & 5 & 90.01 & 1.91 \\
1530 & 1515 & 3 & 0 & 5 & 90.1 & 1.91 \\
1531 & 10176 & 3 & 0 & 5 & 96.22 & 0.26 \\
1532 & 10668 & 3 & 0 & 5 & 96.41 & 0.24 \\
1536 & 10130 & 3 & 0 & 5 & 96.21 & 0.26 \\
1538 & 8753 & 3 & 0 & 5 & 94.42 & 0.64 \\
1542 & 1183 & 3 & 0 & 5 & 91.55 & 0.76 \\
1547 & 1000 & 20 & 0 & 2 & 74.1 & 25.9 \\
1549 & 750 & 40 & 3 & 8 & 22.0 & 7.6 \\
1552 & 1100 & 12 & 4 & 5 & 27.73 & 13.91 \\
1590 & 48842 & 14 & 8 & 2 & 76.07 & 23.93 \\
4534 & 11055 & 30 & 30 & 2 & 55.69 & 44.31 \\
4541 & 101766 & 49 & 36 & 3 & 53.91 & 11.16 \\
6332 & 540 & 37 & 19 & 2 & 57.78 & 42.22 \\
23380 & 2796 & 33 & 2 & 6 & 24.32 & 9.8 \\
23381 & 500 & 12 & 11 & 2 & 58.0 & 42.0 \\
40496 & 500 & 7 & 0 & 10 & 11.4 & 7.4 \\
40498 & 4898 & 11 & 0 & 7 & 44.88 & 0.1 \\
40536 & 8378 & 120 & 61 & 2 & 83.53 & 16.47 \\
40646 & 1600 & 20 & 20 & 2 & 50.0 & 50.0 \\
40647 & 1600 & 20 & 20 & 2 & 50.0 & 50.0 \\
40648 & 1600 & 20 & 20 & 2 & 50.0 & 50.0 \\
40650 & 1600 & 20 & 20 & 2 & 50.0 & 50.0 \\
40668 & 67557 & 42 & 42 & 3 & 65.83 & 9.55 \\
40670 & 3186 & 180 & 180 & 3 & 51.91 & 24.01 \\
40672 & 100968 & 29 & 15 & 8 & 41.71 & 0.01 \\
40677 & 3200 & 24 & 24 & 10 & 10.53 & 9.25 \\
40680 & 1324 & 10 & 10 & 2 & 77.95 & 22.05 \\
40691 & 1599 & 11 & 0 & 6 & 42.59 & 0.63 \\
40693 & 973 & 9 & 9 & 2 & 66.91 & 33.09 \\
40701 & 5000 & 20 & 4 & 2 & 85.86 & 14.14 \\
40705 & 959 & 44 & 2 & 2 & 63.92 & 36.08 \\
40706 & 1124 & 10 & 10 & 2 & 50.44 & 49.56 \\
40900 & 5100 & 36 & 0 & 2 & 98.53 & 1.47 \\
40922 & 88588 & 6 & 0 & 2 & 50.08 & 49.92 \\
40966 & 1080 & 77 & 0 & 8 & 13.89 & 9.72 \\
40971 & 1000 & 19 & 0 & 30 & 8.0 & 0.6 \\
40982 & 1941 & 27 & 0 & 7 & 34.67 & 2.83 \\
40983 & 4839 & 5 & 0 & 2 & 94.61 & 5.39 \\
40994 & 540 & 18 & 0 & 2 & 91.48 & 8.52 \\
41082 & 9298 & 256 & 0 & 10 & 16.7 & 7.61 \\
41084 & 575 & 10304 & 0 & 20 & 8.35 & 3.3 \\
41144 & 3140 & 259 & 0 & 2 & 50.29 & 49.71 \\
41146 & 5124 & 20 & 0 & 2 & 50.0 & 50.0 \\
41150 & 130064 & 50 & 0 & 2 & 71.94 & 28.06 \\
41162 & 72983 & 32 & 18 & 2 & 87.7 & 12.3 \\
41163 & 10000 & 2000 & 0 & 5 & 20.49 & 19.13 \\
41671 & 20000 & 20 & 0 & 5 & 55.81 & 3.72 \\
41972 & 9144 & 220 & 0 & 8 & 44.29 & 0.22 \\
41989 & 51839 & 2916 & 0 & 43 & 5.79 & 0.52 \\
42193 & 5278 & 13 & 6 & 2 & 52.96 & 47.04 \\ \hline
\end{longtable}
\end{center}

\begin{center}
\tiny
\begin{longtable}[c]{lllllll}
\caption{Datasets meta data test.}
\label{tab:sup-dataset-md-test}\\
\hline
\multicolumn{1}{c}{\textbf{\begin{tabular}[c]{@{}c@{}}OpenML \\ ID\end{tabular}}} & \multicolumn{1}{c}{\textbf{\# Examples}} & \multicolumn{1}{c}{\textbf{\# Features}} & \multicolumn{1}{c}{\textbf{\begin{tabular}[c]{@{}c@{}}\# Categorical \\ Features\end{tabular}}} & \multicolumn{1}{c}{\textbf{\# Class}} & \multicolumn{1}{c}{\textbf{\begin{tabular}[c]{@{}c@{}}Majority\\ Class \%\end{tabular}}} & \multicolumn{1}{c}{\textbf{\begin{tabular}[c]{@{}c@{}}Minority\\ Class \%\end{tabular}}} \\ \hline
\endfirsthead
\multicolumn{7}{c}%
{{\bfseries Table \thetable\ continued from previous page}} \\
\hline
\multicolumn{1}{c}{\textbf{\begin{tabular}[c]{@{}c@{}}OpenML \\ ID\end{tabular}}} & \multicolumn{1}{c}{\textbf{\# Examples}} & \multicolumn{1}{c}{\textbf{\# Features}} & \multicolumn{1}{c}{\textbf{\begin{tabular}[c]{@{}c@{}}\# Categorical \\ Features\end{tabular}}} & \multicolumn{1}{c}{\textbf{\# Class}} & \multicolumn{1}{c}{\textbf{\begin{tabular}[c]{@{}c@{}}Majority\\ Class \%\end{tabular}}} & \multicolumn{1}{c}{\textbf{\begin{tabular}[c]{@{}c@{}}Minority\\ Class \%\end{tabular}}} \\ \hline
\endhead
\hline
\endfoot
\endlastfoot
11 & 625 & 4 & 0 & 3 & 46.08 & 7.84 \\
23 & 1473 & 9 & 7 & 3 & 42.7 & 22.61 \\
28 & 5620 & 64 & 0 & 10 & 10.18 & 9.86 \\
30 & 5473 & 10 & 0 & 5 & 89.77 & 0.51 \\
155 & 829201 & 10 & 5 & 10 & 50.11 & 0.0* \\
181 & 1484 & 8 & 0 & 10 & 31.2 & 0.34 \\
307 & 990 & 12 & 2 & 11 & 9.09 & 9.09 \\
725 & 8192 & 8 & 0 & 2 & 59.63 & 40.37 \\
761 & 8192 & 21 & 0 & 2 & 69.76 & 30.24 \\
770 & 625 & 6 & 0 & 2 & 50.4 & 49.6 \\
816 & 8192 & 8 & 0 & 2 & 50.22 & 49.78 \\
823 & 20640 & 8 & 0 & 2 & 56.81 & 43.19 \\
826 & 576 & 11 & 11 & 2 & 58.51 & 41.49 \\
841 & 950 & 9 & 0 & 2 & 51.37 & 48.63 \\
846 & 16599 & 18 & 0 & 2 & 69.09 & 30.91 \\
901 & 40768 & 10 & 0 & 2 & 50.11 & 49.89 \\
934 & 1156 & 5 & 4 & 2 & 77.85 & 22.15 \\
937 & 500 & 50 & 0 & 2 & 56.4 & 43.6 \\
940 & 527 & 36 & 15 & 2 & 84.82 & 15.18 \\
951 & 559 & 4 & 1 & 2 & 97.67 & 2.33 \\
1044 & 10936 & 27 & 3 & 3 & 38.97 & 26.24 \\
1046 & 15545 & 5 & 0 & 2 & 67.14 & 32.86 \\
1053 & 10885 & 21 & 0 & 2 & 80.65 & 19.35 \\
1166 & 1545 & 10935 & 0 & 2 & 87.18 & 12.82 \\
1233 & 945 & 6373 & 0 & 7 & 14.81 & 12.59 \\
1494 & 1055 & 41 & 0 & 2 & 66.26 & 33.74 \\
1501 & 1593 & 256 & 0 & 10 & 10.17 & 9.73 \\
1515 & 571 & 1300 & 0 & 20 & 10.51 & 1.93 \\
1528 & 1623 & 3 & 0 & 5 & 90.63 & 1.79 \\
1535 & 9989 & 3 & 0 & 5 & 96.1 & 0.26 \\
1541 & 8654 & 3 & 0 & 5 & 94.33 & 0.65 \\
1553 & 700 & 12 & 4 & 3 & 35.0 & 30.57 \\
4134 & 3751 & 1776 & 0 & 2 & 54.23 & 45.77 \\
4538 & 9873 & 32 & 0 & 5 & 29.88 & 10.11 \\
40499 & 5500 & 40 & 0 & 11 & 9.09 & 9.09 \\
40649 & 1600 & 20 & 20 & 2 & 50.0 & 50.0 \\
40704 & 2201 & 3 & 0 & 2 & 67.7 & 32.3 \\
40985 & 45781 & 2 & 0 & 20 & 6.35 & 3.05 \\
41145 & 5832 & 308 & 0 & 2 & 50.0 & 50.0 \\
41990 & 51839 & 256 & 0 & 43 & 5.79 & 0.52 \\ \hline
\end{longtable}
\end{center}

\section{Machine Learning Components}
\label{app:feature-data-class-used}

Table \ref{fig:feature-data-class-used} shows the machine learning components used during experiments for dynamic search space composition.

\begin{longtable}[c]{lll}
\caption{Algorithms used in each pipeline step.}
\label{fig:feature-data-class-used}\\
\hline
\textbf{Data Preprocessing} & \textbf{Feature Preprocessing} & \textbf{Modeling} \\ \hline
\endfirsthead
\multicolumn{3}{c}%
{{\bfseries Table \thetable\ continued from previous page}} \\
\hline
\textbf{Data Preprocessing} & \textbf{Feature Preprocessing} & \textbf{Modeling} \\ \hline
\endhead
\hline
\endfoot
\endlastfoot
imputation & extra\_trees\_preproc & adaboost \\
rescaling & fast\_ica & bernoulli\_nb \\
categorical\_encoding & feature\_agglomeration & decision\_tree \\
 & kernel\_pca & extra\_trees \\
 & kitchen\_sinks & gaussian\_nb \\
 & liblinear\_svc\_preprocessor & gradient\_boosting \\
 & no\_preprocessing & k\_nearest\_neighbors \\
 & nystroem\_sampler & lda \\
 & pca & liblinear\_svc \\
 & polynomial & libsvm\_svc \\
 & random\_trees\_embedding & mlp \\
 & select\_percentile & multinomial\_nb \\
 & select\_rates & passive\_aggressive \\
 &  & qda \\
 &  & random\_forest \\
 &  & sgd \\ \hline
\end{longtable}

\section{Pairwise Plots Summarizing Dataset Characteristics with Contour Plots.}
\label{app:pairwise-plots-contour}

\begin{figure}[H]
    \centering
    \includegraphics[width=0.70\textwidth]
    {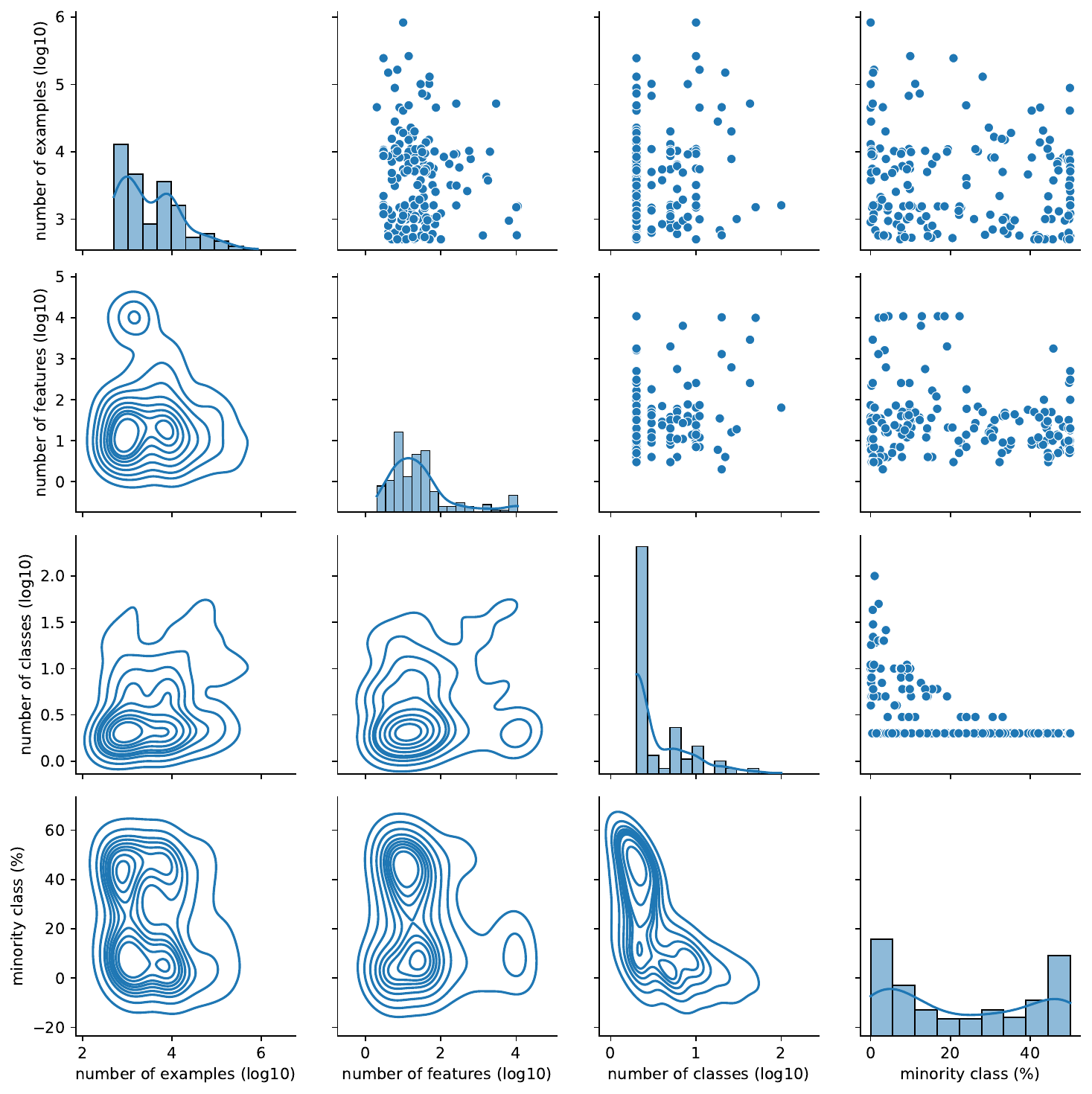}
    \caption{Pairwise plots summarizing dataset characteristics in Dataset set  \(D = D_{\text{train}} \cup D_{\text{test}}\). The diagonal plots show the distribution of each characteristic: the number of examples, number of features, number of classes (all on a \(\log_{10}\) scale), and the percentage of the minority class. The upper triangle contains scatter plots, where each point represents a dataset, while the lower triangle displays contour plots.}    \label{app:datasets_pair_grid_all}
\end{figure}

\begin{figure}[H]
    \centering
    \includegraphics[width=0.70\textwidth]
    {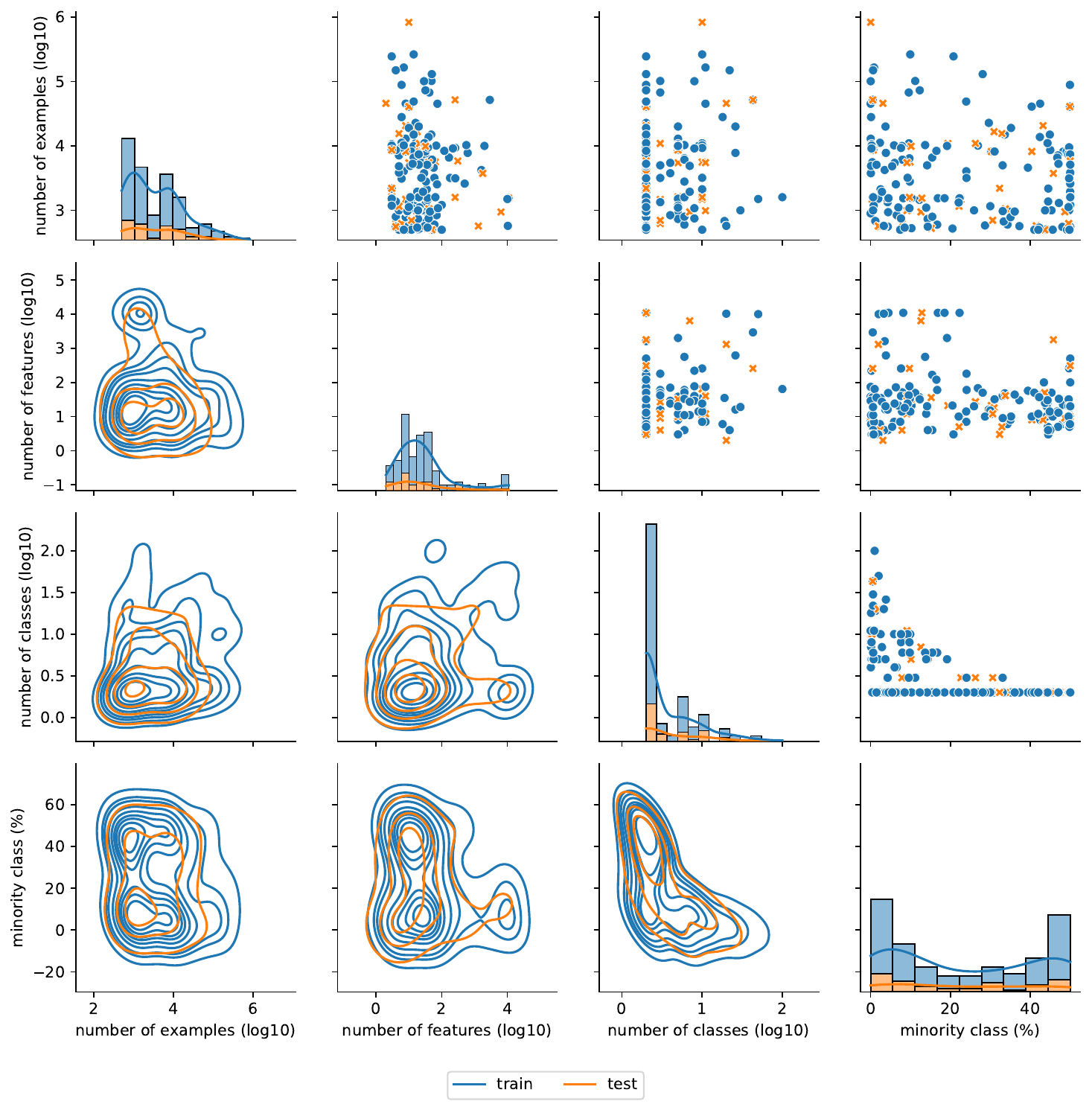}
    \caption{Pairwise plots summarizing dataset characteristics in \(D_{\text{train}}\) (blue) and  \(D_{\text{test}}\) (orange). The diagonal plots show the distribution of each characteristic: the number of examples, number of features, number of classes (all on a \(\log_{10}\) scale), and the percentage of the minority class. The upper triangle contains scatter plots, where each point represents a dataset, while the lower triangle displays contour plots.} \label{app:datasets_pair_grid_group}
\end{figure}

Figure \ref{app:datasets_pair_grid_all} shows the pairwise plots summarizing dataset characteristics with contour plots. Figure \ref{app:datasets_pair_grid_group} shows the same but dividing in $D_\text{train}$ and $D_\text{test}$

\section{Time Spent Analysis for All Meta-features}
\label{app:time-spent-analysis-all-metafeatures}

\begin{figure}[H]
    \centering
    \includegraphics[width=0.69\textwidth]
    {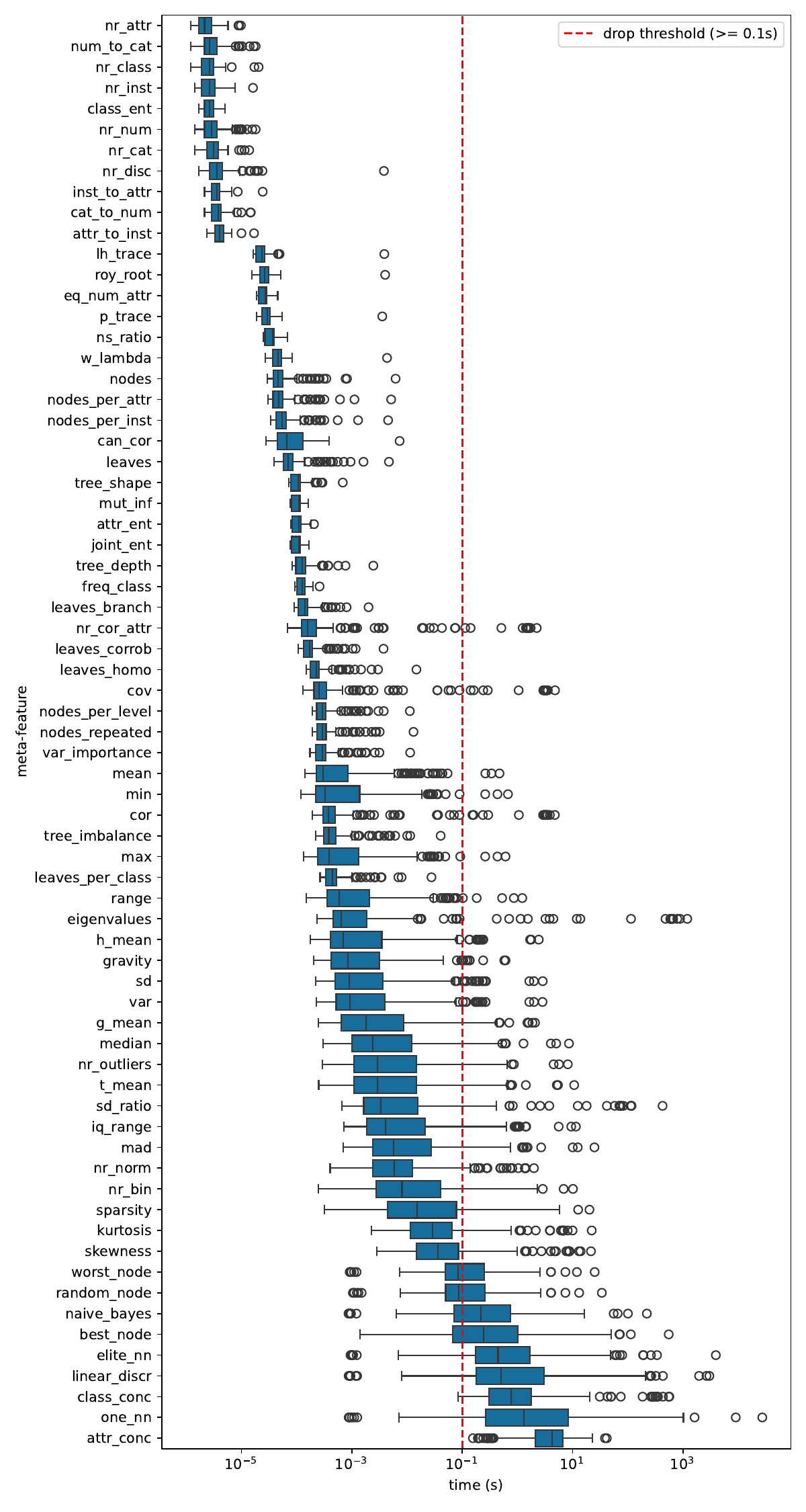}
    \caption{Computation time distribution of meta-features. Distribution of each meta-feature group, with dashed lines representing the first quartile, median, and third quartile. In red a vertical line indicates the selection threshold. Meta-feature with median $>= 0.1s$ were dropped.}
    \label{app:meta-feature-time-spent-all}
\end{figure}

Figure \ref{app:meta-feature-time-spent-all} shows the time spent box plot ordered by median for all meta-features used. The red line indicates the threshold used for drop out meta-features with median time spent $\ge 0.1s$

\section{SHAP Summary Plot of Meta-model}
\label{app:shap-summary-plot-meta-model}

\begin{figure}[H]
    \centering
    \includegraphics[width=0.70\textwidth]
    {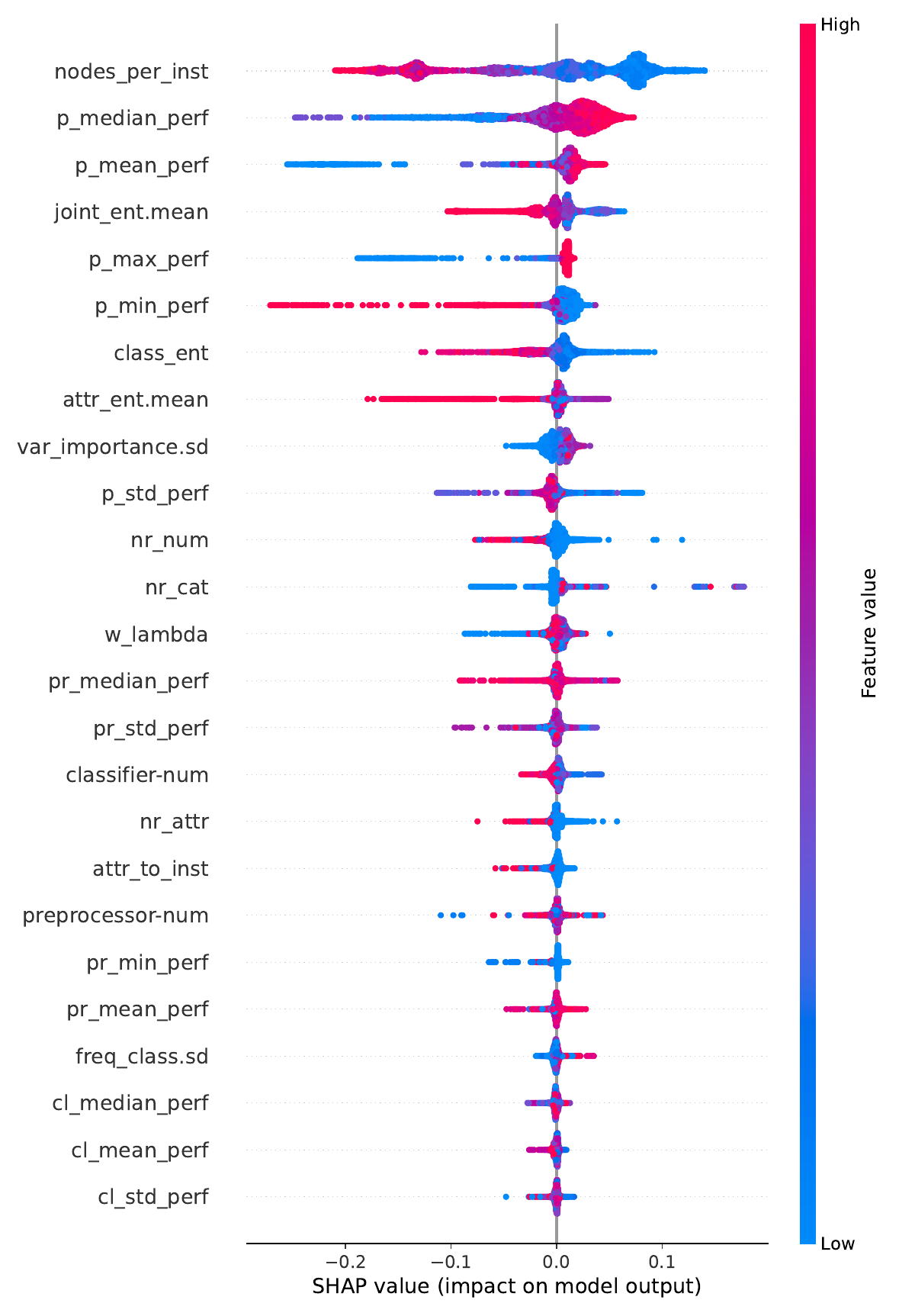}
    \caption{SHAP summary plot showing the contribution of all meta-feature to the meta-model's predictions. Each point represents a SHAP value for a specific feature and instance, with color indicating the feature’s value. Features are ordered by their average absolute SHAP value, highlighting the most influential ones. Positive SHAP values indicate an increase in the prediction, while negative values indicate a decrease.}
    \label{result:shapp-analysis-all}
\end{figure}

Figure \ref{result:shapp-analysis-all} shows the complete SHAP values of the meta-model using the top 25 meta-features. 

\section{Ranking of All Variations Over Time}

\begin{figure}[H]
    \centering
     \makebox[\textwidth]{\includegraphics[scale=0.38]{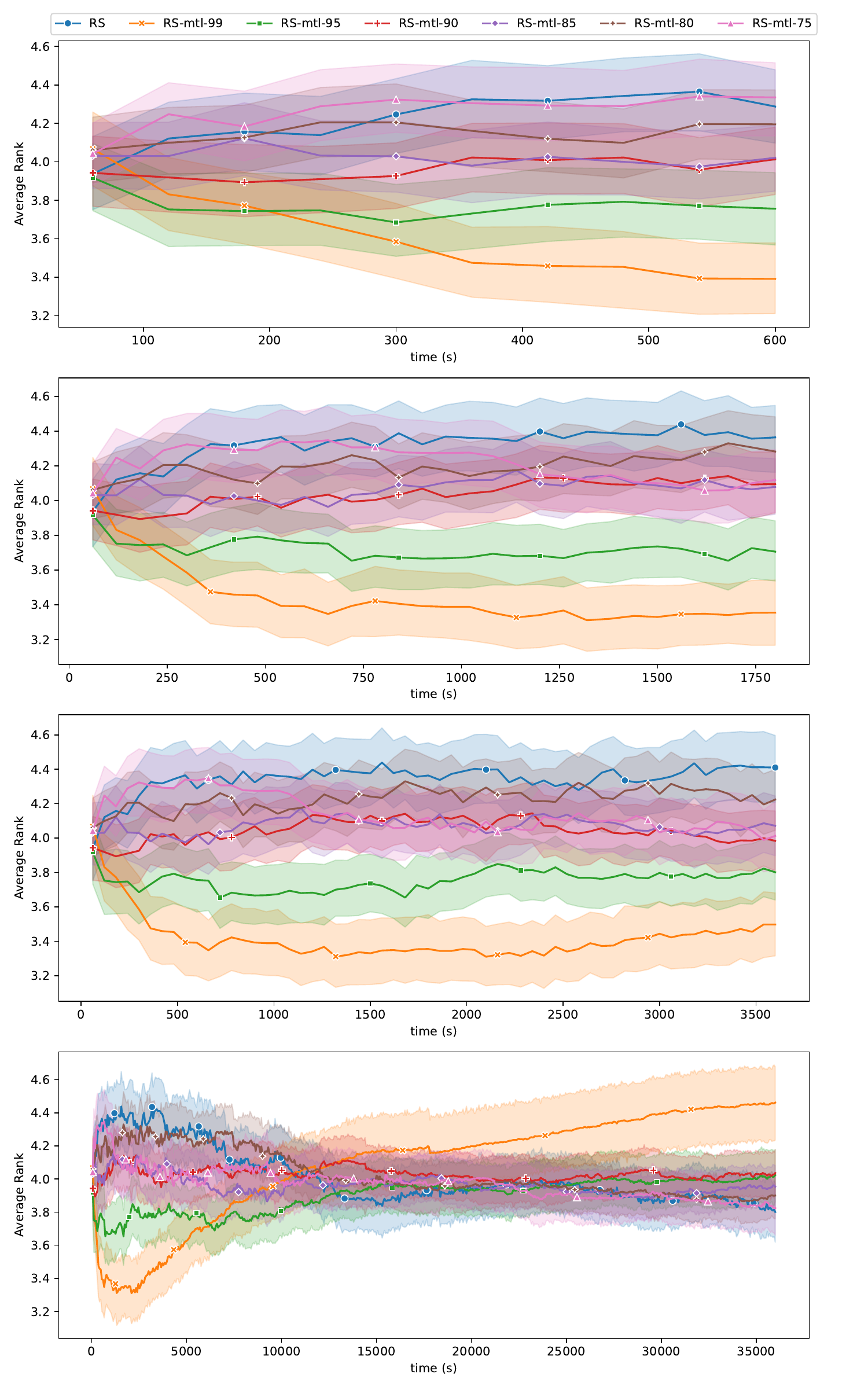}}
    \caption{Ranking RS, RS-mtl-99, RS-mtl-95, RS-mtl-90, RS-mtl-85, RS-mtl-80 and RS-mtl-75.}
    \label{fig:ranking_RS_RS-mtl-99_RS-mtl-95_RS-mtl-90_RS-mtl-85_RS-mtl-80_RS-mtl-75}
\end{figure}

Figure \ref{fig:ranking_RS_RS-mtl-99_RS-mtl-95_RS-mtl-90_RS-mtl-85_RS-mtl-80_RS-mtl-75} shows the ranking performance over time of all RS-mtl-$\theta$ variants tested against the RS complete search space.

\section{Performance Table}
\label{app:performance-table}

Tables \ref{tab:performance-table-at-600s} and \ref{tab:performance-table-at-3600s} depict the F1-weighted mean and standard deviation scores for each dataset across all approaches.

\begin{center}
\tiny
\begin{longtable}[c]{llllllll}
\caption{Mean and standard deviation of F1-weighted scores over 10 experimental repetitions for each dataset across all approaches at 600s.}
\label{tab:performance-table-at-600s}\\
\hline
 & \multicolumn{1}{c}{\textbf{\begin{tabular}[c]{@{}c@{}}OpenML \\ ID\end{tabular}}} & \multicolumn{1}{c}{\textbf{Short-Name}} & \multicolumn{1}{c}{\textbf{RS-mtl-99}} & \multicolumn{1}{c}{\textbf{RS-mtl-95}} & \multicolumn{1}{c}{\textbf{RS-landmarking}} & \multicolumn{1}{c}{\textbf{RS-autosklearn-2}} & \multicolumn{1}{c}{\textbf{RS}} \\ \hline
\endfirsthead
\multicolumn{8}{c}%
{{\bfseries Table \thetable\ continued from previous page}} \\
\hline
 & \multicolumn{1}{c}{\textbf{\begin{tabular}[c]{@{}c@{}}OpenML \\ ID\end{tabular}}} & \multicolumn{1}{c}{\textbf{Short-Name}} & \multicolumn{1}{c}{\textbf{RS-mtl-99}} & \multicolumn{1}{c}{\textbf{RS-mtl-95}} & \multicolumn{1}{c}{\textbf{RS-landmarking}} & \multicolumn{1}{c}{\textbf{RS-autosklearn-2}} & \multicolumn{1}{c}{\textbf{RS}} \\ \hline
\endhead
\hline
\endfoot
\endlastfoot
0 & 11 & balance-... & $ 0.8962 \pm 0.00 $ & $ 0.9630 \pm 0.01 $ & $ 0.8416 \pm 0.00 $ & $ 0.8608 \pm 0.06 $ & $ 0.9167 \pm 0.07 $ \\
1 & 23 & cmc & $ 0.5469 \pm 0.00 $ & $ 0.5043 \pm 0.03 $ & $ 0.5671 \pm 0.00 $ & $ 0.4634 \pm 0.03 $ & $ 0.4740 \pm 0.06 $ \\
2 & 28 & optdigit... & $ 0.9818 \pm 0.00 $ & $ 0.9784 \pm 0.00 $ & $ 0.9836 \pm 0.00 $ & $ 0.9395 \pm 0.05 $ & $ 0.9766 \pm 0.01 $ \\
3 & 30 & page-blo... & $ 0.9664 \pm 0.01 $ & $ 0.8740 \pm 0.31 $ & $ 0.9075 \pm 0.00 $ & $ 0.9666 \pm 0.01 $ & $ 0.9606 \pm 0.01 $ \\
4 & 155 & pokerhan... & $ 0.0000 \pm 0.00 $ & $ 0.0433 \pm 0.14 $ & $ 0.0000 \pm 0.00 $ & $ 0.1591 \pm 0.26 $ & $ 0.1388 \pm 0.32 $ \\
5 & 181 & yeast & $ 0.5532 \pm 0.00 $ & $ 0.5428 \pm 0.01 $ & $ 0.5493 \pm 0.00 $ & $ 0.5437 \pm 0.03 $ & $ 0.5458 \pm 0.03 $ \\
6 & 307 & vowel & $ 0.9478 \pm 0.00 $ & $ 0.9077 \pm 0.05 $ & $ 0.6338 \pm 0.00 $ & $ 0.9157 \pm 0.04 $ & $ 0.8968 \pm 0.10 $ \\
7 & 725 & bank8fm & $ 0.9413 \pm 0.00 $ & $ 0.9387 \pm 0.00 $ & $ 0.9340 \pm 0.00 $ & $ 0.9305 \pm 0.01 $ & $ 0.9326 \pm 0.01 $ \\
8 & 761 & cpu\_act & $ 0.9354 \pm 0.00 $ & $ 0.8820 \pm 0.11 $ & $ 0.1870 \pm 0.39 $ & $ 0.9057 \pm 0.02 $ & $ 0.9185 \pm 0.03 $ \\
9 & 770 & strikes & $ 0.9758 \pm 0.01 $ & $ 0.9834 \pm 0.01 $ & $ 0.8514 \pm 0.00 $ & $ 0.9841 \pm 0.02 $ & $ 0.9751 \pm 0.06 $ \\
10 & 816 & puma8nh & $ 0.8193 \pm 0.00 $ & $ 0.7368 \pm 0.26 $ & $ 0.8340 \pm 0.00 $ & $ 0.7918 \pm 0.04 $ & $ 0.8035 \pm 0.02 $ \\
11 & 823 & houses & $ 0.4826 \pm 0.51 $ & $ 0.9796 \pm 0.00 $ & $ 0.9801 \pm 0.00 $ & $ 0.9675 \pm 0.01 $ & $ 0.9739 \pm 0.01 $ \\
12 & 826 & sensory & $ 0.6601 \pm 0.00 $ & $ 0.6554 \pm 0.01 $ & $ 0.5587 \pm 0.00 $ & $ 0.6408 \pm 0.02 $ & $ 0.6408 \pm 0.02 $ \\
13 & 841 & stock & $ 0.9412 \pm 0.00 $ & $ 0.9609 \pm 0.01 $ & $ 0.9285 \pm 0.00 $ & $ 0.9576 \pm 0.01 $ & $ 0.9580 \pm 0.01 $ \\
14 & 846 & elevator... & $ 0.6120 \pm 0.42 $ & $ 0.7752 \pm 0.27 $ & $ 0.4460 \pm 0.47 $ & $ 0.7437 \pm 0.27 $ & $ 0.8176 \pm 0.08 $ \\
15 & 901 & fried & $ 0.9109 \pm 0.01 $ & $ 0.7144 \pm 0.38 $ & $ 0.9223 \pm 0.00 $ & $ 0.8580 \pm 0.03 $ & $ 0.7880 \pm 0.28 $ \\
16 & 934 & socmob & $ 0.9446 \pm 0.00 $ & $ 0.9418 \pm 0.00 $ & $ 0.9046 \pm 0.00 $ & $ 0.9291 \pm 0.02 $ & $ 0.9280 \pm 0.02 $ \\
17 & 937 & fri\_c3\_5... & $ 0.8783 \pm 0.00 $ & $ 0.8197 \pm 0.07 $ & $ 0.5200 \pm 0.36 $ & $ 0.7621 \pm 0.11 $ & $ 0.7489 \pm 0.07 $ \\
18 & 940 & water-tr... & $ 0.9609 \pm 0.02 $ & $ 0.9567 \pm 0.03 $ & $ 0.8079 \pm 0.00 $ & $ 0.9523 \pm 0.04 $ & $ 0.9265 \pm 0.05 $ \\
19 & 951 & arsenic-... & $ 0.9979 \pm 0.00 $ & $ 0.9941 \pm 0.00 $ & $ 1.0000 \pm 0.00 $ & $ 0.9976 \pm 0.00 $ & $ 0.9984 \pm 0.00 $ \\
20 & 1044 & eye\_move... & $ 0.4872 \pm 0.26 $ & $ 0.4269 \pm 0.30 $ & $ 0.1838 \pm 0.30 $ & $ 0.5322 \pm 0.07 $ & $ 0.4927 \pm 0.12 $ \\
21 & 1046 & mozilla4... & $ 0.9442 \pm 0.00 $ & $ 0.9371 \pm 0.01 $ & $ 0.9370 \pm 0.00 $ & $ 0.9253 \pm 0.03 $ & $ 0.9310 \pm 0.02 $ \\
22 & 1053 & jm1 & $ 0.7746 \pm 0.01 $ & $ 0.7736 \pm 0.01 $ & $ 0.0000 \pm 0.00 $ & $ 0.7608 \pm 0.01 $ & $ 0.7563 \pm 0.02 $ \\
23 & 1166 & ova\_ovar... & $ 0.6385 \pm 0.44 $ & $ 0.3497 \pm 0.45 $ & $ 0.8240 \pm 0.00 $ & $ 0.4464 \pm 0.47 $ & $ 0.8749 \pm 0.04 $ \\
24 & 1233 & eating & $ 0.6128 \pm 0.08 $ & $ 0.2743 \pm 0.28 $ & $ 0.4903 \pm 0.01 $ & $ 0.4563 \pm 0.26 $ & $ 0.4407 \pm 0.16 $ \\
25 & 1494 & qsar-bio... & $ 0.6695 \pm 0.35 $ & $ 0.7303 \pm 0.26 $ & $ 0.4220 \pm 0.44 $ & $ 0.8367 \pm 0.02 $ & $ 0.8295 \pm 0.02 $ \\
26 & 1501 & semeion & $ 0.6408 \pm 0.44 $ & $ 0.8088 \pm 0.29 $ & $ 0.9423 \pm 0.00 $ & $ 0.8073 \pm 0.29 $ & $ 0.8336 \pm 0.29 $ \\
27 & 1515 & micro-ma... & $ 0.8354 \pm 0.01 $ & $ 0.7674 \pm 0.11 $ & $ 0.4273 \pm 0.00 $ & $ 0.8362 \pm 0.12 $ & $ 0.6559 \pm 0.35 $ \\
28 & 1528 & volcanoe... & $ 0.8892 \pm 0.00 $ & $ 0.8939 \pm 0.00 $ & $ 0.8801 \pm 0.00 $ & $ 0.8871 \pm 0.01 $ & $ 0.8778 \pm 0.01 $ \\
29 & 1535 & volcanoe... & $ 0.9499 \pm 0.00 $ & $ 0.9518 \pm 0.00 $ & $ 0.9517 \pm 0.00 $ & $ 0.9452 \pm 0.01 $ & $ 0.9461 \pm 0.01 $ \\
30 & 1541 & volcanoe... & $ 0.9190 \pm 0.00 $ & $ 0.9174 \pm 0.00 $ & $ 0.9132 \pm 0.00 $ & $ 0.9117 \pm 0.01 $ & $ 0.9088 \pm 0.01 $ \\
31 & 1553 & autouniv... & $ 0.4784 \pm 0.00 $ & $ 0.4707 \pm 0.04 $ & $ 0.5074 \pm 0.00 $ & $ 0.4657 \pm 0.05 $ & $ 0.4746 \pm 0.04 $ \\
32 & 4134 & biorespo... & $ 0.8009 \pm 0.01 $ & $ 0.7802 \pm 0.04 $ & $ 0.5068 \pm 0.35 $ & $ 0.5327 \pm 0.37 $ & $ 0.7587 \pm 0.05 $ \\
33 & 4538 & gesturep... & $ 0.1656 \pm 0.27 $ & $ 0.5080 \pm 0.19 $ & $ 0.1727 \pm 0.28 $ & $ 0.4862 \pm 0.10 $ & $ 0.5239 \pm 0.11 $ \\
34 & 40499 & texture & $ 0.9737 \pm 0.00 $ & $ 0.9785 \pm 0.01 $ & $ 0.9913 \pm 0.00 $ & $ 0.8763 \pm 0.31 $ & $ 0.9824 \pm 0.02 $ \\
35 & 40649 & gametes\_... & $ 0.6949 \pm 0.00 $ & $ 0.5712 \pm 0.21 $ & $ 0.6033 \pm 0.00 $ & $ 0.6176 \pm 0.03 $ & $ 0.6340 \pm 0.05 $ \\
36 & 40704 & titanic & $ 0.7745 \pm 0.00 $ & $ 0.7745 \pm 0.00 $ & $ 0.7621 \pm 0.00 $ & $ 0.7729 \pm 0.00 $ & $ 0.7732 \pm 0.00 $ \\
37 & 40985 & tamilnad... & $ 0.0488 \pm 0.01 $ & $ 0.0535 \pm 0.00 $ & $ 0.0539 \pm 0.00 $ & $ 0.0431 \pm 0.01 $ & $ 0.0321 \pm 0.02 $ \\
38 & 41145 & philippi... & $ 0.8014 \pm 0.02 $ & $ 0.6707 \pm 0.24 $ & $ 0.0000 \pm 0.00 $ & $ 0.6253 \pm 0.22 $ & $ 0.5601 \pm 0.30 $ \\
39 & 41990 & gtsrb-hu... & $ 0.1200 \pm 0.17 $ & $ 0.1385 \pm 0.17 $ & $ 0.0000 \pm 0.00 $ & $ 0.0804 \pm 0.10 $ & $ 0.0414 \pm 0.09 $ \\ \hline
\end{longtable}
\end{center}

\begin{center}
\centering    
\tiny
\begin{longtable}[c]{llllllll}
\caption{Mean and standard deviation of F1-weighted scores over 10 experimental repetitions for each dataset across all approaches at 3600s.}
\label{tab:performance-table-at-3600s}\\
\hline
\multicolumn{1}{c}{\textbf{}} & \multicolumn{1}{c}{\textbf{\begin{tabular}[c]{@{}c@{}}OpenML\\ ID\end{tabular}}} & \multicolumn{1}{c}{\textbf{Short-Name}} & \multicolumn{1}{c}{\textbf{RS-mtl-99}} & \multicolumn{1}{c}{\textbf{RS-mtl-95}} & \multicolumn{1}{c}{\textbf{RS-landmarking}} & \multicolumn{1}{c}{\textbf{RS-autosklearn-2}} & \multicolumn{1}{c}{\textbf{RS}} \\ \hline
\endfirsthead
\multicolumn{8}{c}%
{{\bfseries Table \thetable\ continued from previous page}} \\
\hline
\multicolumn{1}{c}{\textbf{}} & \multicolumn{1}{c}{\textbf{\begin{tabular}[c]{@{}c@{}}OpenML\\ ID\end{tabular}}} & \multicolumn{1}{c}{\textbf{Short-Name}} & \multicolumn{1}{c}{\textbf{RS-mtl-99}} & \multicolumn{1}{c}{\textbf{RS-mtl-95}} & \multicolumn{1}{c}{\textbf{RS-landmarking}} & \multicolumn{1}{c}{\textbf{RS-autosklearn-2}} & \multicolumn{1}{c}{\textbf{RS}} \\ \hline
\endhead
\hline
\endfoot
\endlastfoot
0 & 11 & balance-... & $ 0.8962 \pm 0.00 $ & $ 0.9754 \pm 0.00 $ & $ 0.8416 \pm 0.00 $ & $ 0.9289 \pm 0.05 $ & $ 0.9391 \pm 0.05 $ \\
1 & 23 & cmc & $ 0.5469 \pm 0.00 $ & $ 0.4993 \pm 0.00 $ & $ 0.5671 \pm 0.00 $ & $ 0.4890 \pm 0.03 $ & $ 0.4986 \pm 0.02 $ \\
2 & 28 & optdigit... & $ 0.9801 \pm 0.00 $ & $ 0.9829 \pm 0.00 $ & $ 0.9836 \pm 0.00 $ & $ 0.9664 \pm 0.03 $ & $ 0.9813 \pm 0.01 $ \\
3 & 30 & page-blo... & $ 0.9784 \pm 0.00 $ & $ 0.9728 \pm 0.00 $ & $ 0.9075 \pm 0.00 $ & $ 0.9726 \pm 0.00 $ & $ 0.9667 \pm 0.01 $ \\
4 & 155 & pokerhan... & $ 0.0000 \pm 0.00 $ & $ 0.3764 \pm 0.41 $ & $ 0.0000 \pm 0.00 $ & $ 0.4752 \pm 0.26 $ & $ 0.3694 \pm 0.36 $ \\
5 & 181 & yeast & $ 0.5532 \pm 0.00 $ & $ 0.5428 \pm 0.00 $ & $ 0.5493 \pm 0.00 $ & $ 0.5731 \pm 0.02 $ & $ 0.5628 \pm 0.02 $ \\
6 & 307 & vowel & $ 0.9478 \pm 0.00 $ & $ 0.9556 \pm 0.00 $ & $ 0.6338 \pm 0.00 $ & $ 0.9512 \pm 0.00 $ & $ 0.9438 \pm 0.01 $ \\
7 & 725 & bank8fm & $ 0.9413 \pm 0.00 $ & $ 0.9394 \pm 0.00 $ & $ 0.9340 \pm 0.00 $ & $ 0.9380 \pm 0.00 $ & $ 0.9374 \pm 0.00 $ \\
8 & 761 & cpu\_act & $ 0.9354 \pm 0.00 $ & $ 0.9348 \pm 0.00 $ & $ 0.9376 \pm 0.00 $ & $ 0.9221 \pm 0.02 $ & $ 0.9295 \pm 0.01 $ \\
9 & 770 & strikes & $ 0.9758 \pm 0.01 $ & $ 0.9885 \pm 0.01 $ & $ 0.8514 \pm 0.00 $ & $ 0.9936 \pm 0.01 $ & $ 0.9681 \pm 0.06 $ \\
10 & 816 & puma8nh & $ 0.8193 \pm 0.00 $ & $ 0.8192 \pm 0.00 $ & $ 0.8340 \pm 0.00 $ & $ 0.8184 \pm 0.00 $ & $ 0.8184 \pm 0.00 $ \\
11 & 823 & houses & $ 0.9799 \pm 0.00 $ & $ 0.9800 \pm 0.00 $ & $ 0.9801 \pm 0.00 $ & $ 0.9797 \pm 0.00 $ & $ 0.9761 \pm 0.01 $ \\
12 & 826 & sensory & $ 0.6601 \pm 0.00 $ & $ 0.6662 \pm 0.01 $ & $ 0.5587 \pm 0.00 $ & $ 0.6665 \pm 0.03 $ & $ 0.6631 \pm 0.02 $ \\
13 & 841 & stock & $ 0.9412 \pm 0.00 $ & $ 0.9664 \pm 0.00 $ & $ 0.9285 \pm 0.00 $ & $ 0.9685 \pm 0.00 $ & $ 0.9639 \pm 0.01 $ \\
14 & 846 & elevator... & $ 0.8786 \pm 0.01 $ & $ 0.7922 \pm 0.28 $ & $ 0.8798 \pm 0.00 $ & $ 0.8472 \pm 0.05 $ & $ 0.8590 \pm 0.04 $ \\
15 & 901 & fried & $ 0.9121 \pm 0.00 $ & $ 0.8291 \pm 0.29 $ & $ 0.9223 \pm 0.00 $ & $ 0.9031 \pm 0.04 $ & $ 0.8963 \pm 0.03 $ \\
16 & 934 & socmob & $ 0.9446 \pm 0.00 $ & $ 0.9417 \pm 0.00 $ & $ 0.9046 \pm 0.00 $ & $ 0.9368 \pm 0.01 $ & $ 0.9358 \pm 0.02 $ \\
17 & 937 & fri\_c3\_5... & $ 0.8783 \pm 0.00 $ & $ 0.8783 \pm 0.00 $ & $ 0.8792 \pm 0.00 $ & $ 0.8146 \pm 0.06 $ & $ 0.8421 \pm 0.06 $ \\
18 & 940 & water-tr... & $ 0.9609 \pm 0.02 $ & $ 0.9838 \pm 0.02 $ & $ 0.8079 \pm 0.00 $ & $ 0.9565 \pm 0.02 $ & $ 0.9469 \pm 0.04 $ \\
19 & 951 & arsenic-... & $ 0.9979 \pm 0.00 $ & $ 0.9932 \pm 0.00 $ & $ 1.0000 \pm 0.00 $ & $ 0.9962 \pm 0.00 $ & $ 0.9978 \pm 0.00 $ \\
20 & 1044 & eye\_move... & $ 0.6648 \pm 0.00 $ & $ 0.5217 \pm 0.28 $ & $ 0.3721 \pm 0.32 $ & $ 0.6229 \pm 0.08 $ & $ 0.5536 \pm 0.14 $ \\
21 & 1046 & mozilla4... & $ 0.9442 \pm 0.00 $ & $ 0.9442 \pm 0.00 $ & $ 0.9370 \pm 0.00 $ & $ 0.9418 \pm 0.00 $ & $ 0.9366 \pm 0.02 $ \\
22 & 1053 & jm1 & $ 0.7834 \pm 0.00 $ & $ 0.7661 \pm 0.01 $ & $ 0.7781 \pm 0.00 $ & $ 0.7661 \pm 0.01 $ & $ 0.7622 \pm 0.01 $ \\
23 & 1166 & ova\_ovar... & $ 0.9365 \pm 0.01 $ & $ 0.9320 \pm 0.01 $ & $ 0.8240 \pm 0.00 $ & $ 0.8244 \pm 0.29 $ & $ 0.9008 \pm 0.05 $ \\
24 & 1233 & eating & $ 0.6646 \pm 0.00 $ & $ 0.5091 \pm 0.27 $ & $ 0.4806 \pm 0.00 $ & $ 0.5387 \pm 0.20 $ & $ 0.5213 \pm 0.18 $ \\
25 & 1494 & qsar-bio... & $ 0.8236 \pm 0.00 $ & $ 0.8362 \pm 0.02 $ & $ 0.8236 \pm 0.00 $ & $ 0.8318 \pm 0.02 $ & $ 0.8377 \pm 0.01 $ \\
26 & 1501 & semeion & $ 0.9272 \pm 0.00 $ & $ 0.9221 \pm 0.02 $ & $ 0.9423 \pm 0.00 $ & $ 0.9276 \pm 0.02 $ & $ 0.9368 \pm 0.01 $ \\
27 & 1515 & micro-ma... & $ 0.8330 \pm 0.00 $ & $ 0.8142 \pm 0.00 $ & $ 0.4273 \pm 0.00 $ & $ 0.9034 \pm 0.05 $ & $ 0.8805 \pm 0.06 $ \\
28 & 1528 & volcanoe... & $ 0.8892 \pm 0.00 $ & $ 0.8939 \pm 0.00 $ & $ 0.8801 \pm 0.00 $ & $ 0.8926 \pm 0.01 $ & $ 0.8880 \pm 0.01 $ \\
29 & 1535 & volcanoe... & $ 0.9546 \pm 0.00 $ & $ 0.9520 \pm 0.00 $ & $ 0.9517 \pm 0.00 $ & $ 0.9477 \pm 0.01 $ & $ 0.9495 \pm 0.00 $ \\
30 & 1541 & volcanoe... & $ 0.9190 \pm 0.00 $ & $ 0.9174 \pm 0.00 $ & $ 0.9132 \pm 0.00 $ & $ 0.9144 \pm 0.00 $ & $ 0.9155 \pm 0.00 $ \\
31 & 1553 & autouniv... & $ 0.4784 \pm 0.00 $ & $ 0.4596 \pm 0.00 $ & $ 0.5074 \pm 0.00 $ & $ 0.4746 \pm 0.04 $ & $ 0.4793 \pm 0.04 $ \\
32 & 4134 & biorespo... & $ 0.8068 \pm 0.00 $ & $ 0.8047 \pm 0.00 $ & $ 0.7700 \pm 0.00 $ & $ 0.6050 \pm 0.32 $ & $ 0.7708 \pm 0.02 $ \\
33 & 4538 & gesturep... & $ 0.6392 \pm 0.00 $ & $ 0.5417 \pm 0.20 $ & $ 0.1727 \pm 0.28 $ & $ 0.5917 \pm 0.07 $ & $ 0.5481 \pm 0.13 $ \\
34 & 40499 & texture & $ 0.9738 \pm 0.00 $ & $ 0.9887 \pm 0.00 $ & $ 0.9913 \pm 0.00 $ & $ 0.9839 \pm 0.03 $ & $ 0.9881 \pm 0.01 $ \\
35 & 40649 & gametes\_... & $ 0.6949 \pm 0.00 $ & $ 0.6675 \pm 0.00 $ & $ 0.6033 \pm 0.00 $ & $ 0.6274 \pm 0.04 $ & $ 0.6632 \pm 0.01 $ \\
36 & 40704 & titanic & $ 0.7745 \pm 0.00 $ & $ 0.7745 \pm 0.00 $ & $ 0.7621 \pm 0.00 $ & $ 0.7745 \pm 0.00 $ & $ 0.7745 \pm 0.00 $ \\
37 & 40985 & tamilnad... & $ 0.0551 \pm 0.00 $ & $ 0.0546 \pm 0.00 $ & $ 0.0551 \pm 0.00 $ & $ 0.0480 \pm 0.01 $ & $ 0.0374 \pm 0.02 $ \\
38 & 41145 & philippi... & $ 0.8107 \pm 0.00 $ & $ 0.7831 \pm 0.03 $ & $ 0.0000 \pm 0.00 $ & $ 0.6343 \pm 0.23 $ & $ 0.6724 \pm 0.24 $ \\
39 & 41990 & gtsrb-hu... & $ 0.4061 \pm 0.13 $ & $ 0.3199 \pm 0.26 $ & $ 0.0000 \pm 0.00 $ & $ 0.2851 \pm 0.25 $ & $ 0.2310 \pm 0.23 $ \\ \hline
\end{longtable}
\end{center}

\section{Percentage of Time Reduction for $\text{RS-mtl-}\theta$}
\label{app:boxplot_time_reduction-theta}

\begin{figure}[H]
    \centering
     \makebox[\textwidth]{\includegraphics[scale=0.50]{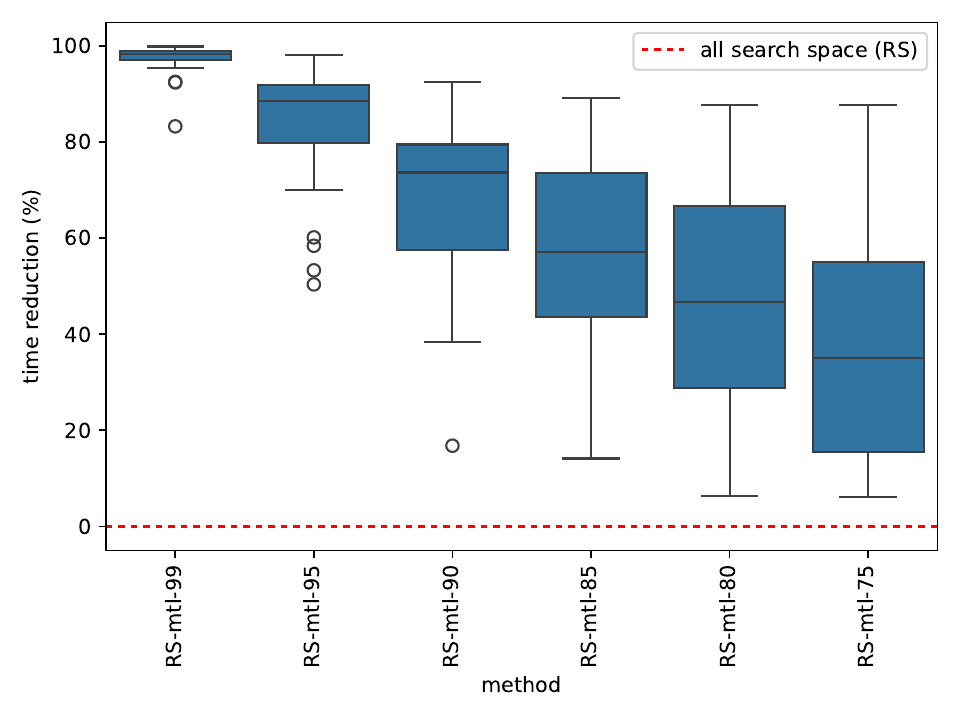}}
    \caption{Percentage of time reduction relative to the RS baseline. The reduction is calculated by dividing the total runtime of each $\text{RS-mtl-}\theta$ approach per dataset by the runtime of RS for the same dataset. In red is the search for all space, which means a 0\% reduction.}
    \label{fig:boxplot_time_reduction_theta}
\end{figure}

Figure \ref{fig:boxplot_time_reduction_theta} shows the percentage of time reduction for all variations studied (0.99, 0.95, 0.90, 0.85, 0.80, 0.75).

\section{Percentage of Preprocessor-Classifier Combination was Recommended (All)}
\label{app:perc_pipelines_all}

\begin{figure}[H]
    \centering
     \makebox[\textwidth]{\includegraphics[scale=0.35]{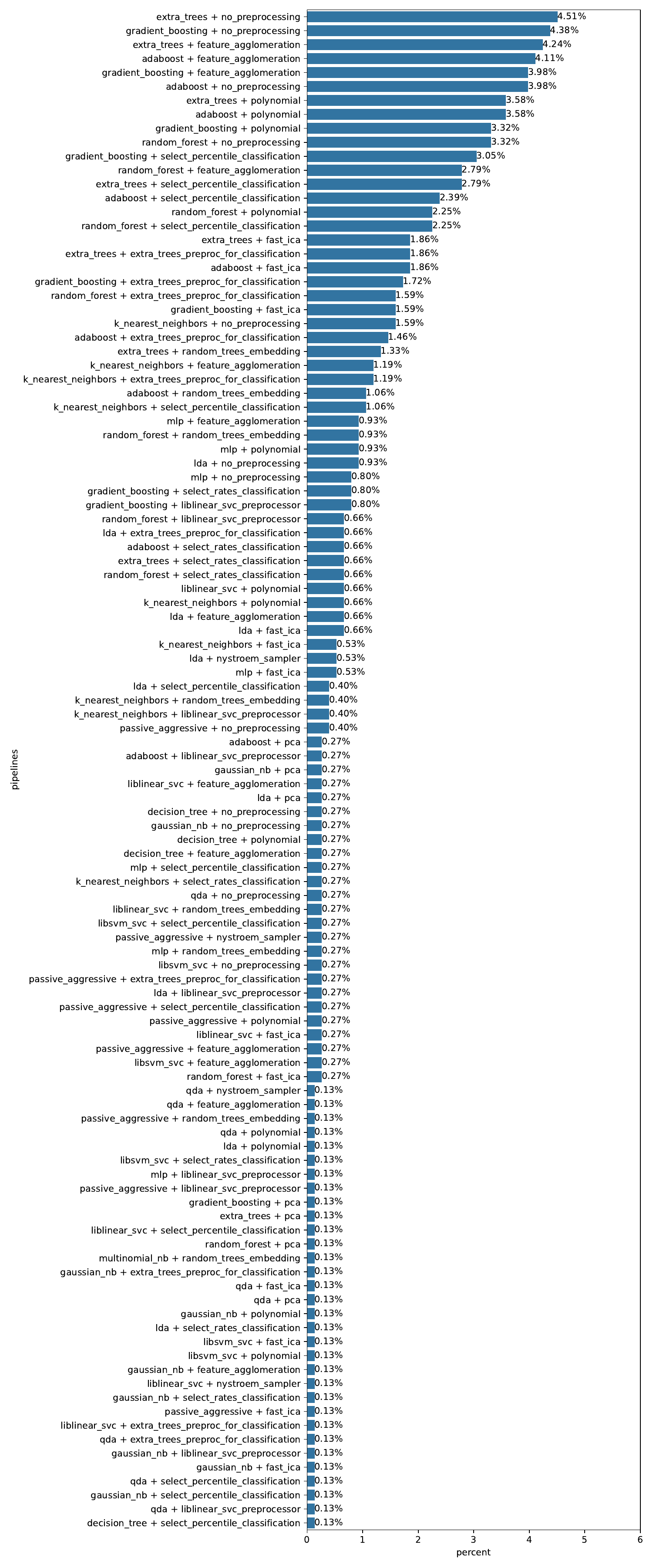}}
    \caption{Percentage of times each preprocessor-classifier combination was included in the dynamically designed search space of RS-mtl-95 (all combinations).}
    \label{fig:perc_pipelines_all}
\end{figure}

Figure \ref{fig:perc_pipelines_all} shows the percentage of the preprocessor-classifier combination that was mostly recommended by the meta-model.

\section{Auto-Sklearn with Dynamic Search Space $\theta=0.95$}
\label{app:autosklearn-dss-at-theta-95}

\begin{figure}[H]
\centering
\begin{tikzpicture}[xscale=2]
\node (Label) at (2.1657458295101266, 0.7){\tiny{CD = 0.17}}; 
\draw[decorate,decoration={snake,amplitude=.4mm,segment length=1.5mm,post length=0mm},very thick, color = black] (2.0,0.5) -- (2.3314916590202532,0.5);
\foreach \x in {2.0, 2.3314916590202532} \draw[thick,color = black] (\x, 0.4) -- (\x, 0.6);
 
\draw[gray, thick](2.0,0) -- (6.0,0);
\foreach \x in {2.0,4.0,6.0} \draw (\x cm,1.5pt) -- (\x cm, -1.5pt);
\node (Label) at (2.0,0.2){\tiny{1}};
\node (Label) at (4.0,0.2){\tiny{2}};
\node (Label) at (6.0,0.2){\tiny{3}};
\draw[decorate,decoration={snake,amplitude=.4mm,segment length=1.5mm,post length=0mm},very thick, color = black](3.3925000000000005,-0.25) -- (3.605,-0.25);
\node (Point) at (3.4425000000000003, 0){};\node (Label) at (0.5,-0.45){\scriptsize{autosklearn}}; \draw (Point) |- (Label);
\node (Point) at (5.0025, 0){};\node (Label) at (6.5,-0.45){\scriptsize{random\_forest}}; \draw (Point) |- (Label);
\node (Point) at (3.555, 0){};\node (Label) at (6.5,-0.75){\scriptsize{autosklearn\_dss\_95}}; \draw (Point) |- (Label);
\end{tikzpicture}
\caption{Friedman test with Nemenyi post-hoc analysis ($\alpha = 0.05$) comparing Auto-Sklearn without metalearning step (\texttt{autosklearn}), Auto-Sklearn without metalearning adapted to dynamic search space $\theta=0.95$ (\texttt{autosklearn-dss-95}), and a Random Forest baseline (\texttt{random-forest}) at 3600 seconds.}
\label{fig:autosklearn-dss-at-theta-95-without}
\end{figure}
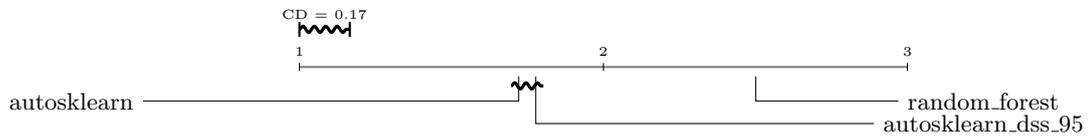

Figure \ref{fig:autosklearn-dss-at-theta-95-without} presents the Friedman test with Nemenyi post-hoc analysis comparing Auto-Sklearn, Auto-Sklearn with dynamic search space ($\theta=0.95$), and the Random Forest.

\end{appendices}
\end{document}